\pdfoutput=1
\documentclass[11pt]{article}
\usepackage[final]{acl}
\usepackage{changepage}

\usepackage{times}
\usepackage{latexsym}
\usepackage[T1]{fontenc}
\usepackage[utf8]{inputenc}

\usepackage{microtype}
\usepackage{inconsolata}
\usepackage{graphicx}
\usepackage{tcolorbox}
\usepackage{amsmath}
\usepackage{subcaption}
\usepackage{url} 
\usepackage{booktabs}
\usepackage{amssymb}
\usepackage{todonotes}
\usepackage{multirow}
\usepackage{colortbl}
\usepackage{xcolor}

\usepackage{tabularx}
\usepackage{array}
\usepackage{caption}
\usepackage{makecell} 

\usepackage{marvosym}

\usepackage{siunitx}

\definecolor{BestOne}{RGB}{198,239,206}   
\definecolor{CompOne}{RGB}{226,245,231}   

\definecolor{BestTwo}{RGB}{189,215,238}   
\definecolor{CompTwo}{RGB}{221,235,247}   

\definecolor{BestThree}{RGB}{255,230,153} 
\definecolor{CompThree}{RGB}{255,242,204} 

\newcommand{\bestone}[1]{\cellcolor{BestOne}\textbf{#1}}
\newcommand{\compone}[1]{\cellcolor{CompOne}#1}

\newcommand{\besttwo}[1]{\cellcolor{BestTwo}\textbf{#1}}
\newcommand{\comptwo}[1]{\cellcolor{CompTwo}#1}

\newcommand{\bestthree}[1]{\cellcolor{BestThree}\textbf{#1}}
\newcommand{\compthree}[1]{\cellcolor{CompThree}#1}

\definecolor{mediumred}{RGB}{255,100,100}

\newcolumntype{Y}{>{\raggedright\arraybackslash}X}

\definecolor{bestcolor}{HTML}{FFF2CC} 

\newcommand{\affilsup}[1]{\rlap{\textsuperscript{\normalfont#1}}}

\title{Multilingual Steering by Design: Multilingual Sparse Autoencoders and Principled Layer Selection}
\author{
  Yusser Al Ghussin\affilsup{$^{1,2}$} \qquad Daniil Gurgurov\affilsup{$^{1,2}$} \qquad Tanja Bäumel\affilsup{$^{1,2,5}$}  \\ 
  \textbf{Josef van Genabith}\affilsup{$^{1,2}$} \qquad \textbf{Patrick Schramowski}\affilsup{$^{2, 3,4,5}$} \qquad \textbf{Simon Ostermann}\affilsup{$^{1,2,5}$} \\
  \\
  \small $^1$ Saarland University \enspace $^2$ German Research Center for Artificial Intelligence (DFKI) \\ 
   \enspace \small $^3$ TU Darmstadt \enspace \small $^4$ hessian.AI \enspace \small $^5$ Centre for European Research in Trusted AI (CERTAIN)  \\
  \texttt{\small yusser.al\_ghussin@dfki.de}
}

\begin{document}
\maketitle

\begin{abstract}
Sparse autoencoders (SAEs) enable feature-level mechanistic interpretability and activation steering in large language models (LLMs), but SAE-based language control remains unreliable in multilingual settings: most SAEs are trained on English-only data, and steering layers are chosen heuristically. We address these limitations by advancing a principled, mechanistic account of multilingual language steering with SAEs.
First, we show that training SAEs on multilingual data consistently strengthens cross-lingual representations and yields more reliable, quality-preserving language control across layers and model families. Second, we introduce an \emph{a priori} steering layer-selection rule based on the intersection of multilingual alignment and language separability, which predicts effective intervention depths without exhaustive layerwise search. 
We evaluate our approach on LLaMA-3.1-8B and Gemma-2-9B across machine translation and cross-lingual summarization (CrossSumm), using SpBLEU, ROUGE-L, COMET, and LaSE. Our results show that multilingual SAEs combined with intersection-selected layers stabilize the trade-off between language identification accuracy and generation quality, providing a principled, predictive, representation-level account of multilingual SAE steering. We release all code and models for reproducibility.
\footnote{\url{https://github.com/Yusser96/Multilingual-Steering-by-Design/}}
\footnote{\url{https://huggingface.co/collections/Yusser/multilingual-steering-by-design}}
\end{abstract}

\section{Introduction}

\begin{figure}[h]
    \centering
    \includegraphics[width=1\linewidth]{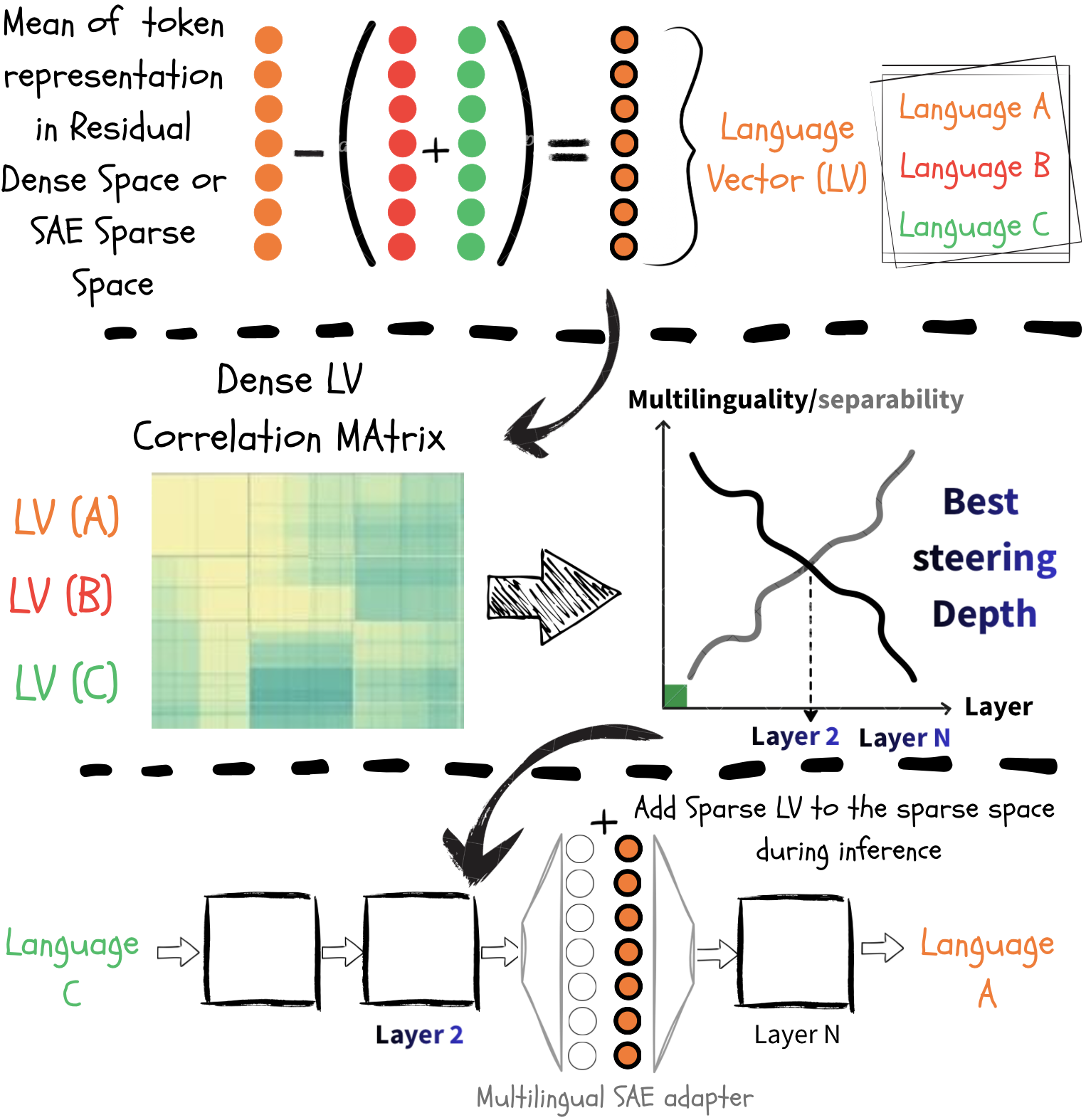}
    \caption{Overview of our language-control pipeline. A language-specific vector is constructed and used for layer selection and generation steering.}
    \label{fig:teaser}
\end{figure}

Large language models (LLMs) can generate text in many languages, yet reliably \emph{controlling} the output language remains challenging. While sparse autoencoders (SAEs) have emerged as a promising tool for interpreting internal activations and constructing steering vectors that causally influence model behavior \cite{cunningham2023sparse, templeton2024scaling}, SAE-based language steering in multilingual settings remains brittle and difficult to reproduce, with steering success varying unpredictably across models and layers: intervention depths are typically chosen heuristically (e.g., “mid-to-late” layers), requiring expensive layer sweeps and yielding inconsistent outcomes \cite{bayat2025steeringlargelanguagemodel, chou2025causal}. As a result, although SAE steering can work, it lacks a predictive, mechanistic account
of \emph{where} and \emph{why} language control should be applied inside the model \cite{tang2024language, deng2025unveilinglanguagespecificfeatureslarge}.

We argue that this haphazardness stems from the lack of a mechanistic perspective on how multilingual information is organized across model depth.
We show that effective language steering requires access to two complementary signals: shared cross-lingual structure that supports fluent generation across languages, and language-specific information that distinguishes one language from another. Prior work has shown that multilingual pretrained models learn shared latent representations across languages, facilitating cross-lingual transfer even in the absence of shared vocabularies or parallel data \cite{conneau2020emerging}. At the same time, language identity and language-specific features are differentially encoded across layers and can transition toward shared abstractions over depth in multilingual models \cite{riemenschneider2025cross,zhang2024same}. If an intervention targets layers dominated by shared structure, steering lacks specificity; if it targets layers dominated by language-specific signals, the model often fails to recover generation quality.
Our hypothesis reframes language steering as a problem of identifying representational balance points, rather than amplifying language-specific features in isolation, as is common in prior work \cite{tang2024language,deng2025unveilinglanguagespecificfeatureslarge,gurgurov2025languagearithmeticssystematiclanguage}.

In this work, we operationalize this mechanistic hypothesis through two complementary contributions. First, we train SAEs directly on multilingual data for LLaMA-3.1-8B \cite{grattafiori2024llama} and Gemma-2-9B \cite{team2024gemma}, showing that multilingual training preserves the shared cross-lingual structure and language-specific distinctions required for predictable and interpretable steering in the sparse representation space. Compared to open-source SAEs \cite{he2024llama, lieberum2024gemma}, these multilingual SAEs yield more stable and quality-preserving language steering across layers and model families. Second, we introduce a principled, \emph{a priori} rule for selecting steering layers based on the intersection of multilingual alignment and language separability, which predicts effective intervention depths without exhaustive layerwise search. Figure~\ref{fig:teaser} provides an overview of the proposed language steering framework.


We validate this mechanistic framework across machine translation and cross-lingual summarization on LLaMA-3.1-8B and Gemma-2-9B, explicitly testing the prediction that balanced layers yield optimal language identification accuracy and generation quality trade-offs. Across both benchmarks, we find that multilingual SAEs combined with intersection-selected layers consistently stabilize language control and improve interpretability, supporting the view that effective steering depth is a property of the model’s internal multilingual organization rather than a heuristic tuning choice.

Our contributions are threefold:
\begin{itemize}
\setlength\itemsep{0.001em}
\item \textbf{Mechanistic characterization of language across depth.}
We show that effective language steering arises at layers where cross-lingual alignment and language separability coexist.

\item \textbf{Principled, \emph{a priori} layer selection.}
We introduce an intersection-based criterion that predicts effective steering depths without layer sweeps.

\item \textbf{Multilingual SAEs as an interpretability enabler for language steering.}
We show that multilingual SAE training preserves the representational structure required for reliable, interpretable language control.
\end{itemize}

\begin{figure}[t]
    \centering
    \includegraphics[width=1.0\linewidth]{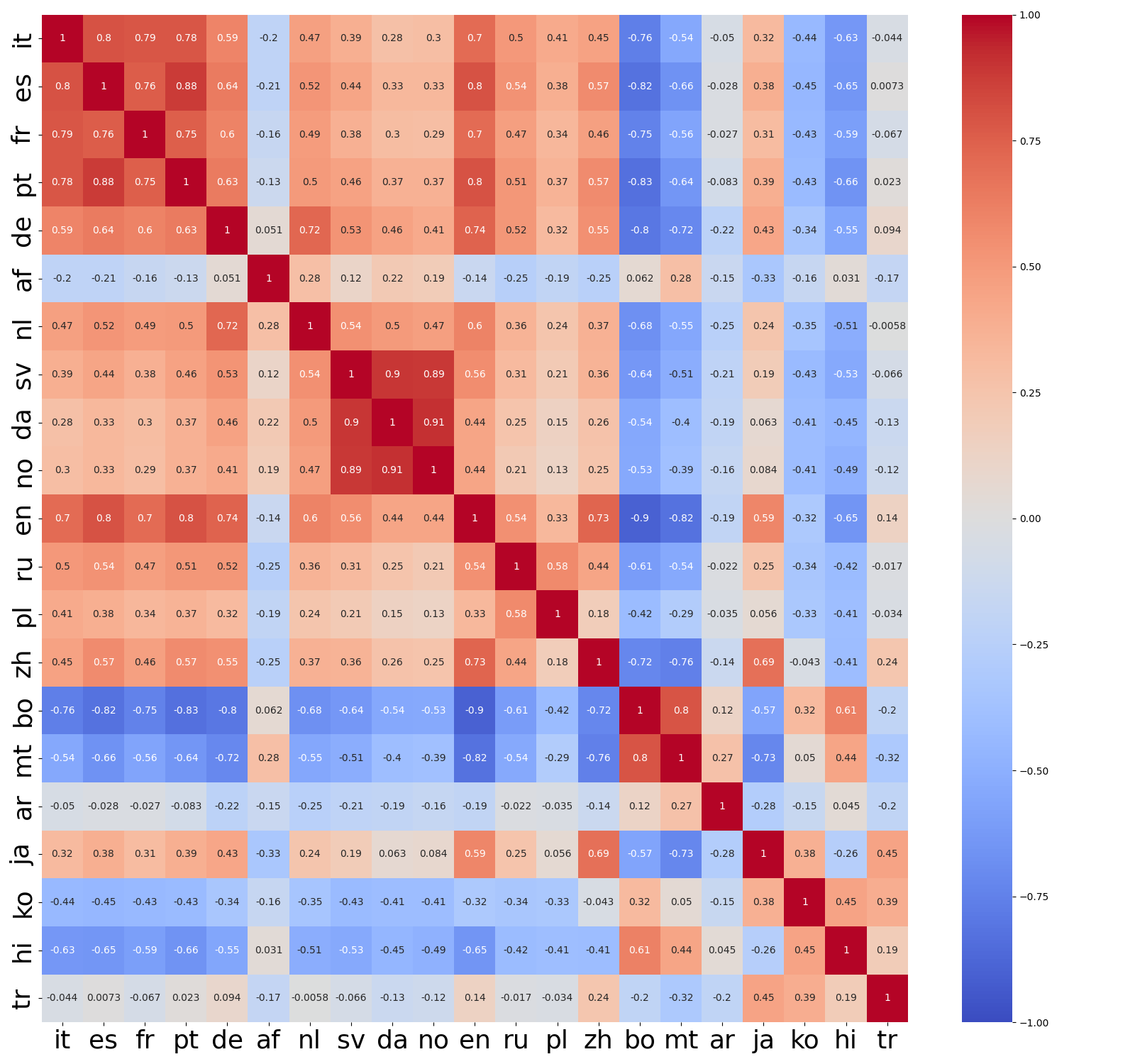}
    \caption{Correlation matrices of per-language contrast (DiffMean) vectors for Gemma-2-9B (Layer 23).}
    \label{fig:langfam_corr}
\end{figure}

\section{Related Work}

 \paragraph{SAE-Based Activation and Language Steering.}
Sparse autoencoders (SAEs) have been widely used to interpret and steer internal activations in large language models \citep{templeton2024scaling,zhao2024steering,o2024steering,wang2025enhancing,zhao2025denoising}. 
Methods such as Sparse Activation Steering (SAS) \citep{bayat2025steeringlargelanguagemodel}, Feature-Guided Activation Addition (FGAA) \citep{soo2025interpretable}, and SAE-Targeted Steering (SAE-TS) \citep{chalnev2024improving} demonstrate that manipulating small sets of sparse features can causally influence model behavior. 
Applied to language control, prior work shows that editing individual SAE features can flip output language in models such as Gemma-2-9B and LLaMA-3.1-8B \citep{chou2025causal,deng2025unveilinglanguagespecificfeatureslarge, gurgurov2026clasbenchcrosslingualalignmentsteering}. 
However, effective steering depths are typically identified through manual exploration or fixed heuristics (e.g., mid-to-late layers), and many existing approaches rely on SAEs trained predominantly on English data. As a result, these methods do not provide a predictive, mechanistic account of where language steering should be applied across depth, nor how multilingual structure is preserved in sparse representations.

\paragraph{Evaluating and Training SAEs.}
Recent benchmarks such as SAE-Bench \citep{karvonen2025saebench} and AxBench \citep{wu2025axbench} evaluate SAE fidelity, interpretability, and intervention quality, reporting mixed results for SAE-based steering compared to simpler baselines. 
Other work emphasizes reconstruction fidelity as critical for causal interventions: Gemma-Scope \citep{lieberum2024gemma} and LLaMA-Scope \citep{he2024llama} report that high reconstruction error degrades steering effectiveness, while JumpReLU SAEs \citep{rajamanoharan2407jumping} improve the fidelity–sparsity trade-off via straight-through training. 
These findings suggest that insufficient SAE fidelity may disproportionately affect low-frequency or multilingual features, motivating our use of high-fidelity JumpReLU SAEs for multilingual language steering.

\paragraph{Language Features inside Models.}
Beyond SAEs, prior analyses point to strong layer-dependent language signals in multilingual models. 
\citet{tang2024language} identify language-specific neurons in BLOOM and LLaMA-2 and show that toggling them can switch the output language. 
\citet{chang2022geometry} study multilingual geometry in XLM-R, finding that languages occupy approximately parallel subspaces separated by linear ``language vectors'' particularly in middle layers; shifting hidden states along these directions flips predictions. 
Our findings echo these trends in the depth-wise distribution of multilingual structure and support treating language as a steerable direction in representation space \cite{gurgurov2026clasbenchcrosslingualalignmentsteering}, while further revealing correlations among language families \citep{gurgurov2025languagearithmeticssystematiclanguage}.

Together, these lines of work motivate the need for a representation-level account of multilingual language steering that explains both how language information is organized across depth and how this organization can be exploited to guide interventions predictively.

\section{Language Representations and Principled Steering}

Our goal is not merely to improve language control, but to explain \emph{where} and \emph{why} language steering is possible inside multilingual LLMs, and to use this explanation to guide interventions \emph{a priori}.

We define \textbf{language vectors} as directions in representation space that capture both the presence of individual languages and the directions along which they can be causally steered, building on prior evidence that language identity is linearly encoded as a direction or low-dimensional subspace within model representations \citep{park2023linearrepresentation,deng2025unveilinglanguagespecificfeatureslarge}. Our layer-selection criterion is motivated by the observation that reliable language control requires access to two complementary signals: (i) \emph{alignment}, corresponding to shared cross-lingual structure that supports generation across languages, and (ii) \emph{separability}, corresponding to language-specific information that distinguishes one language from another. Only at depths where these signals are balanced can a small intervention reliably steer the output language.

\subsection{Language Vectors}

At each layer, we represent languages using contrastive \emph{language vectors} constructed from model activations, either in the dense residual stream or in the sparse space induced by an SAE. Given activations from a target language and a set of other languages, we 
construct language steering vectors using the DiffMean method \citep{wu2025axbench}. For a given target language at layer $\ell$, let $\mathcal{Z}^+$ denote the set of sparse codes corresponding to examples in the target language, and $\mathcal{Z}^-$ the set corresponding to all other languages. 
We compute the mean sparse representations 
by averaging SAE codes over all non-special tokens from all examples in that language
\[
\bar z^+_{\ell} = \frac{1}{|\mathcal Z^+|} \sum_{z \in \mathcal Z^+} z,
\qquad
\bar z^-_{\ell} = \frac{1}{|\mathcal Z^-|} \sum_{z \in \mathcal Z^-} z,
\]
and define the steering vector as
\[
w_{\mathrm{DiffMean}}(\ell) = \bar z^+_{\ell} - \bar z^-_{\ell}.
\]

These vectors are then used additively in the SAE space to influence model outputs. Full mathematical definitions of the SAE representations, DiffMean steering vectors, and the inference-time steering procedure are provided in Appendix~\ref{app:language_vectors}.

Beyond serving as steering directions, these language vectors exhibit meaningful linguistic structure. In particular, at the layers selected by our intersection-based criterion, pairwise correlations between per-language vectors reveal clear language-family groupings. As shown in Figure~\ref{fig:langfam_corr}, languages from the same family (e.g., Romance or Germanic) exhibit high mutual similarity, while cross-family correlations remain lower. At the same time, a shared multilingual component persists across families, reflecting common cross-lingual structure. This coexistence of shared alignment and family-specific separation aligns with the intuition behind our layer-selection criterion and helps explain why these depths yield strong trade-offs between language identification accuracy and generation quality.

\subsection{Multilingual SAEs for Language Steering}
\label{sec:why_multilingual_sae}
A central design choice in our framework is to train sparse autoencoders on multilingual data rather than English-only corpora. This choice is not merely pragmatic, but mechanistically important for reliable and interpretable language steering.

English-only SAEs preferentially encode monolingual structure: features that are frequent and salient in English dominate the sparse representation, while cross-lingual correlations and low-frequency language-specific features are weakly represented or collapse entirely. As a result, steering directions constructed from such representations are brittle. Language vectors may activate English-correlated features without cleanly isolating the intended target language, and the relationship between steering depth and downstream behavior becomes unstable. In contrast, multilingual SAE training exposes the autoencoder to systematic variation across languages, encouraging the sparse feature space to preserve both shared cross-lingual structure and language-specific distinctions. 

From this perspective, multilingual SAEs can act as an \emph{interpretability enabler} for representation-level language steering. They maintain the representational structure required to construct steering vectors whose effects can be predicted from representation-level statistics. The experimental comparisons in later sections empirically validate this claim, but the motivation for multilingual training arises directly from the mechanistic requirements of language steering.

\subsection{Principled Layer Selection}


A common assumption in prior work is that effective language control primarily relies on manipulating strongly language-specific features, which are believed to emerge in later layers, where the model is less able to recover from an intervention \citep{tang2024language,gurgurov2025languagearithmeticssystematiclanguage}. We find that effective steering emerges at depths where language-specific signals coexist with sufficient shared cross-lingual structure, motivating an \emph{a priori} layer-selection strategy based on the depthwise evolution of language representations.
At each layer, we quantify \emph{multilinguality} as the degree to which language vectors share a dominant common direction, and \emph{separability} as the extent to which languages remain distinct in representation space.

Let $\{\lambda_j\}_{j=1}^N$ be the eigenvalues of the language vectors pairwise Pearson correlation matrix $C_{\ell}$; $N$ is the number of languages. We define the \emph{multilinguality} score as the explained-variance ratio of the first principal component,
\[
f_{\ell} = \frac{\max_j \lambda_j}{\sum_{k=1}^N \lambda_k},
\]
which measures the degree of shared alignment across languages. We define \emph{separability} as the complementary quantity
\[
s_{\ell} = 1 - f_{\ell},
\]
which reflects how distinct the language representations remain. We select steering layers at intersection points where these two signals are balanced, corresponding to depths that jointly preserve shared semantic structure while exposing discriminative language information.

We empirically validate this criterion across models, SAE variants, and tasks. 
Our contribution is to replace such heuristic choices with a principled, data-driven criterion that predicts these depths \emph{before} training SAEs and steering experiments are run. Full definitions of the language correlation matrices, multilinguality and separability metrics, and the intersection-based layer-selection procedure are given in Appendix~\ref{app:layer_selection}.

\section{Experiments}
\subsection{Models and Data}
We evaluate on \emph{LLaMA‑3.1‑8B} \cite{grattafiori2024llama} and \emph{Gemma‑2‑9B} \cite{team2024gemma} using 21 FLORES–200 languages (see Appendix~\ref{app:lang_labels}) \cite{costa2022no}. 
For each model, we train parallel English-only and multilingual JumpReLU SAE suites \citep{rajamanoharan2407jumping} on 2.1B Wikipedia \cite{wikimedia_wikipedia_dump_20231101} tokens with identical architectures and optimization settings, isolating the effect of multilingual training data. Full details are provided in Appendix~\ref{app:SAEtraining}.

\subsection{Evaluation}

\paragraph{FLORES–200 Machine Translation.}
We evaluate language steering on machine translation using FLORES–200 \cite{costa2022no}. We use the \texttt{dev} split to construct steering vectors and the \texttt{devtest} split for evaluation. Each \texttt{devtest} set contains approximately 1{,}000 sentences per language, providing a substantially large and clean evaluation set while ensuring strict separation between steering construction and evaluation. 

For each non-English target language $i$ in our language set ($|i|=20$), we define an English$\rightarrow$\texttt{tgt\_$i$} translation task, where English (\texttt{eng\_Latn}) is always the source language. We construct a per-language steering vector using \texttt{dev} sentences, and apply this vector to steer generation into the intended output language, which we denote as \texttt{steer\_$i$}.\footnote{We use \texttt{tgt\_$i$} for the prompt language and \texttt{steer\_$i$} for the intended output language after steering. In our setup, \texttt{steer\_$i$} is the language for which we construct the steering vector. We use the term ``steer language'' to emphasize that the output language is controlled via a steering intervention.}

Prompts are written \emph{in the target language} using natural translation instructions (e.g., German: ``\textit{Übersetze diesen Satz:}''), followed by a target-language answer cue (e.g., ``\textit{Übersetzung:}''). We provide prompt examples in the Appendix~\ref{app:flores_prompts}.
\[
\underbrace{\texttt{Translate this sentence:}}_{\text{instruction in target language}} 
\; \underbrace{``\texttt{<source text>}"}_{\text{Always English}}  .
\]
\[
\underbrace{\texttt{Translation:}}_{\text{answer cue in target language}}
\]
This setup biases the model toward both the translation task and the prompt language, so that any deviation toward a steering language can be attributed to the steering intervention rather than prompt ambiguity. We decode using greedy search with temperature $0$, yielding a conservative and interpretable baseline that isolates the effect of steering from prompt engineering or decoding strategies.
We report relative differences across SAE variants relative to open-source SAE baselines, which directly measure the effectiveness of multilingual training and layer selection, 
for three metrics. (1) \textbf{LangID}, computed by applying a fastText language identification classifier \cite{joulin2016fasttext} to the generated outputs, measures how reliably steering enforces the intended output language. (2) \textbf{SpBLEU} \cite{post-2018-call}, computed against the reference translation in the intended \emph{steer language}, provides a script-agnostic measure of surface-level translation quality. (3) \textbf{COMET} \cite{rei2020comet}, a neural evaluation metric that leverages cross-lingual pretrained encoders and both the source and reference sentences, estimates semantic translation quality and correlates strongly with human judgments.

We report results averaged across all 20 non-English prompt languages, where for each \texttt{tgt\_$i$} we evaluate steering into every other target language \texttt{steer\_$j$} with $j \neq i$. Concretely, the model is prompted in language \texttt{tgt\_$i$} and steered toward \texttt{steer\_$j$}, and results are averaged over the full cross-product of $(i,j)$ pairs. This aggregation directly measures how reliably different SAE variants enable control over the output language, independent of any single prompt–language pairing. As our primary focus is the \emph{relative} (delta) performance differences between SAE variants rather than absolute per-language scores, we present these averages in the main text, while detailed per-language and per-pair results are provided in Appendix~\ref{app:per_lang_diff_res}.

In addition, we report a restricted setting where steering is applied only when $\texttt{steer\_$j$}=\texttt{tgt\_$i$}$, allowing us to analyze the behavior of steering when the prompt language and intended output language coincide (Appendix~\ref{app:per_lang_same_res}).

\begin{figure*}[t]
    \centering

    \includegraphics[width=0.3\linewidth]{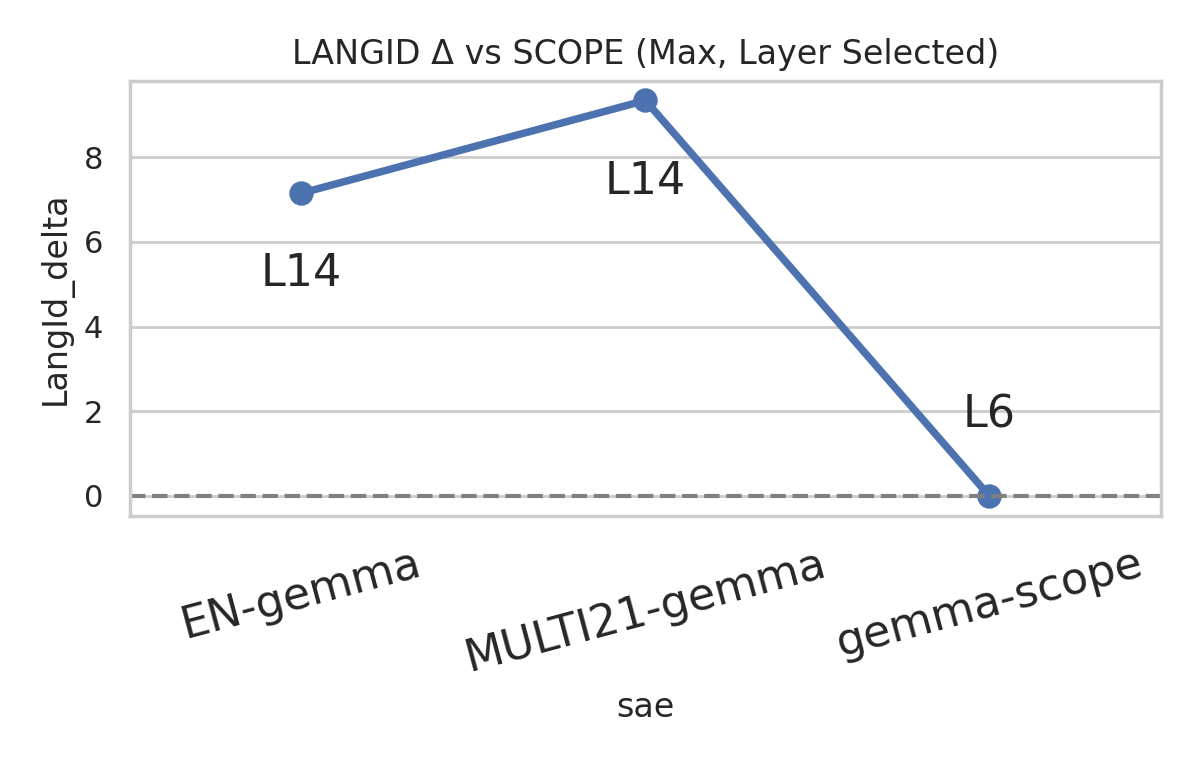}
    \includegraphics[width=0.3\linewidth]{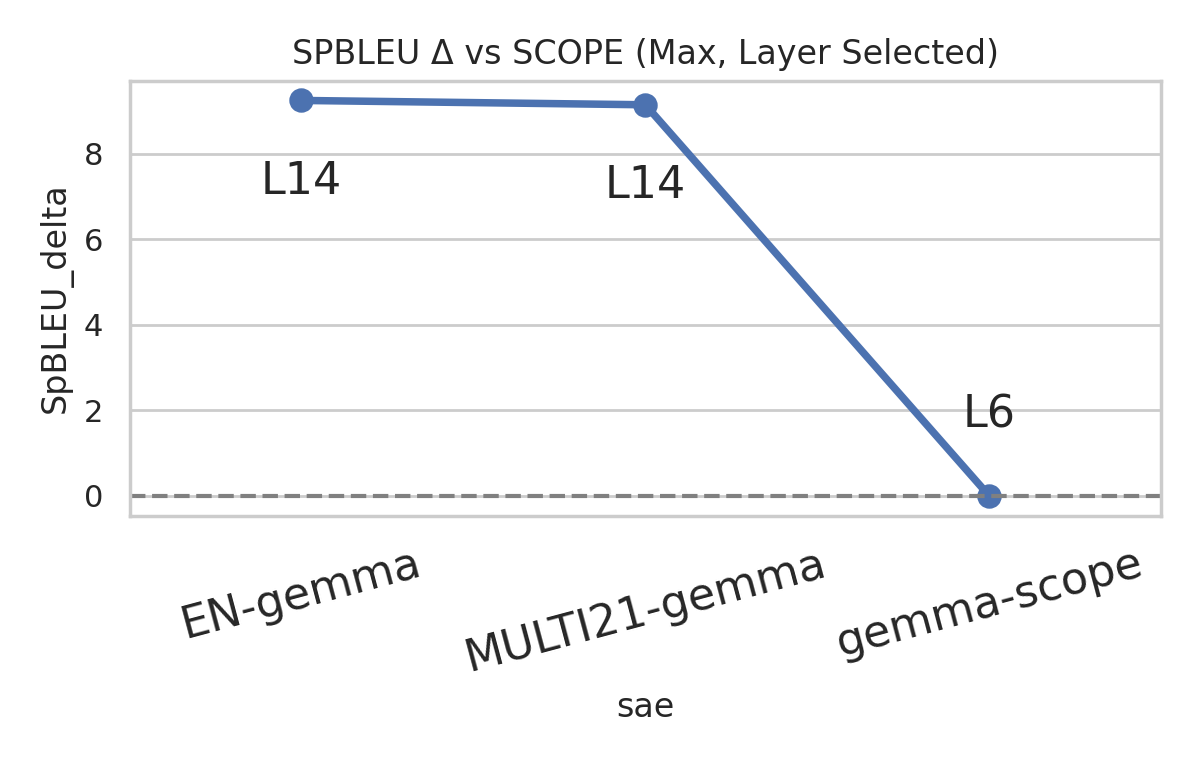}
    \includegraphics[width=0.3\linewidth]{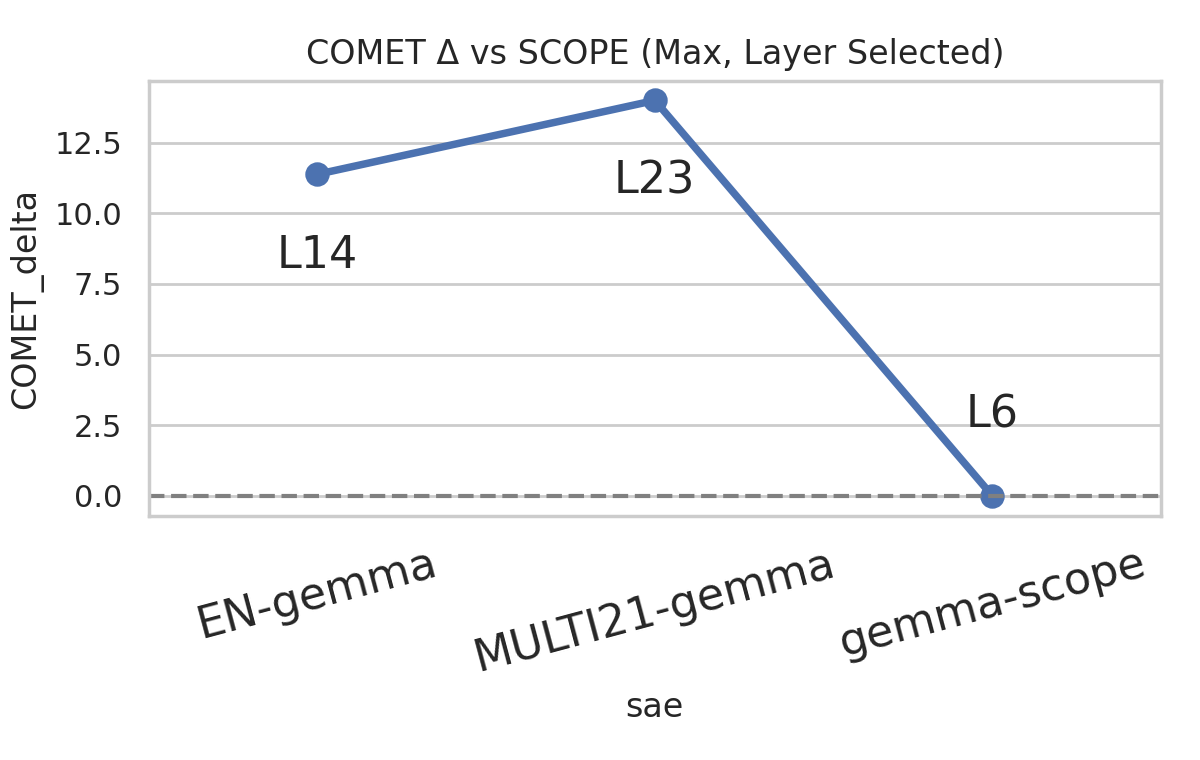}  \\
    
    \includegraphics[width=0.3\linewidth]{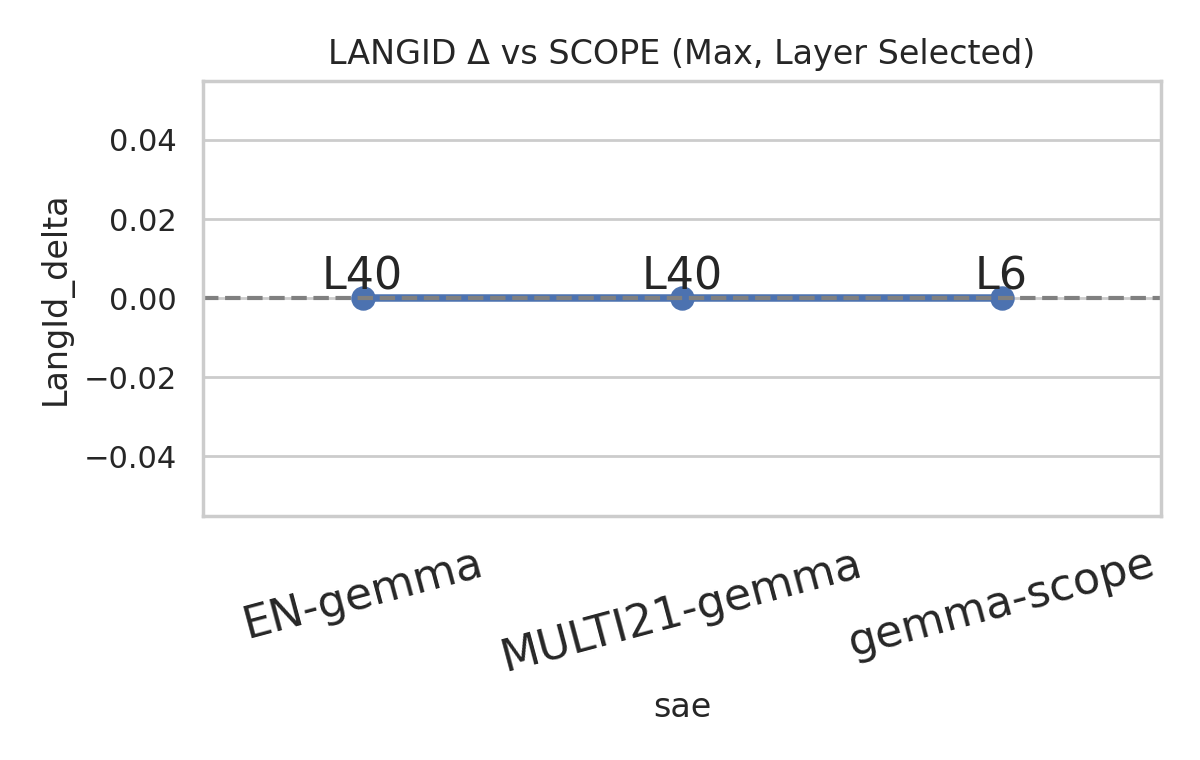}
    \includegraphics[width=0.3\linewidth]{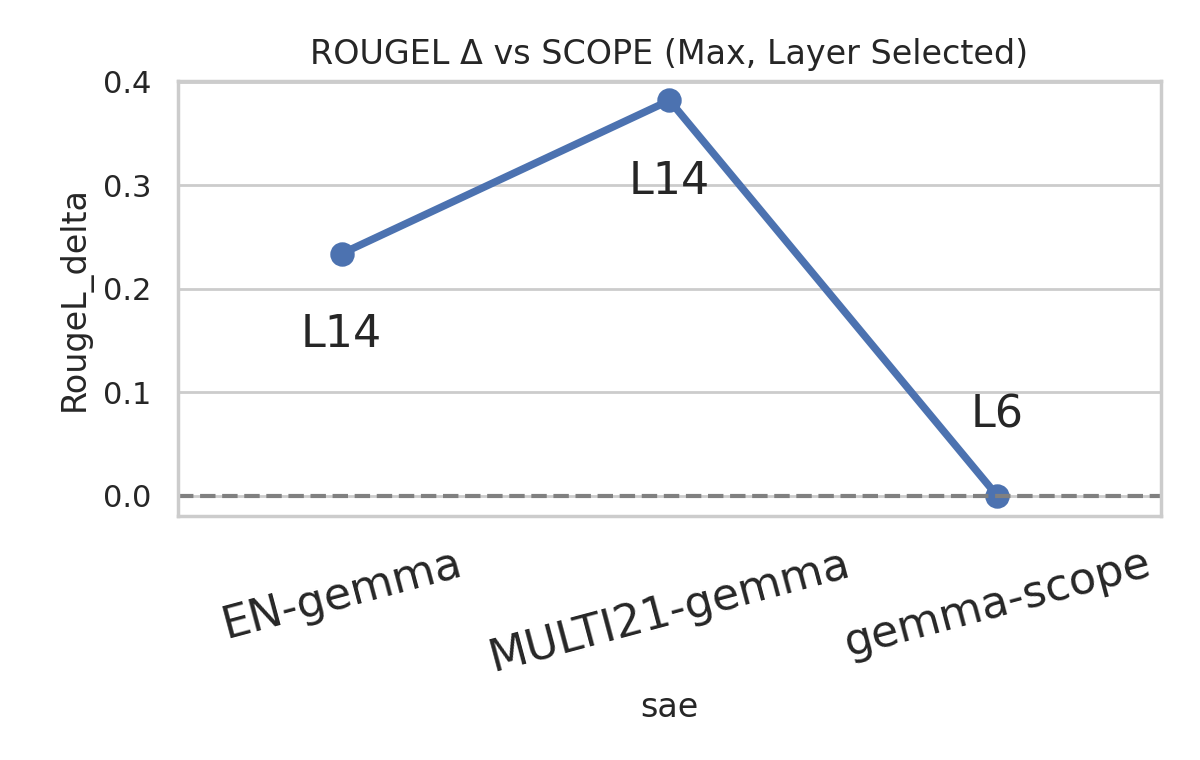}
    \includegraphics[width=0.3\linewidth]{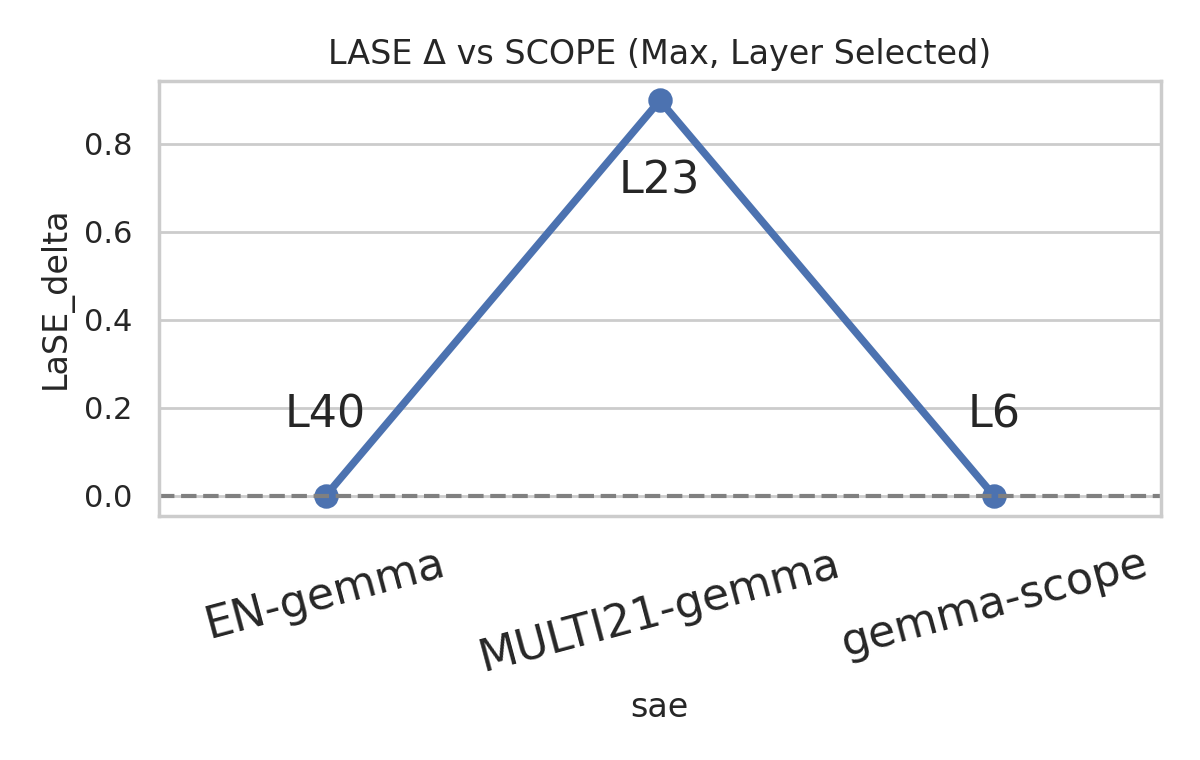}

    \caption{Performance deltas relative to Scope baselines for \textbf{Gemma-2-9B} at the best-performing steering layer. \textbf{Top:} FLORES machine translation (LangID, SpBLEU, COMET). \textbf{Bottom:} Cross-lingual summarization (LangID, ROUGE-L, LaSE).}
    \label{fig:gemma_deltas}
\end{figure*}

\paragraph{Cross-Lingual Summarization (CrossSumm).}
To evaluate whether our findings generalize beyond translation, we use the cross-lingual summarization dataset CrossSum \citep{app15147800}. We select document–summary pairs whose target languages intersect with our translation language set. The resulting dataset consists of 108 fully parallel English source documents paired with reference summaries in one of five target languages: Spanish (es), Russian (ru), Arabic (ar), Hindi (hi), and Turkish (tr).

We follow the same experimental design as in machine translation, reusing the same per-language steering vectors. The only change is the prompt, which is written in the target language and phrased as a natural summarization instruction in the target-language (e.g., “Summarize the following article”) with a target-language answer cue ("summary:"), thereby biasing the model toward both the summarization task and the target language. We provide prompt examples in the Appendix~\ref{app:cross_sum_prompts}.

We evaluate generated summaries using three metrics: \textbf{LangID}, \textbf{ROUGE-L} \citep{lin2004rouge}, and \textbf{LaSE} \cite{app15147800}, following the evaluation protocol of the original dataset. ROUGE-L measures content overlap with the reference summary, while LaSE evaluates cross-lingual semantic similarity between the generated and reference summaries. This setup allows us to test whether steering preserves semantic content while controlling the output language in a non-translation generative task.

\section{Results}
\label{sec:results}
We assess steering performance and layer sensitivity, focusing on: (i) benefits from multilingual SAE training, (ii) optimal intervention layers, and (iii) comparisons with open‑source SAEs.

\subsection{Benefits of Multilingual Training for SAEs}

We study the effect of training data on SAE steering, comparing monolingual (English‑only) SAEs with multilingual SAEs across transformer layers. We evaluate multilingual steering across two generation tasks: (i) machine translation and (ii) cross-lingual summarization, focusing on language identification accuracy, surface quality, and semantic preservation.

\paragraph{Multilingual training improves steering.}
Figures~\ref{fig:gemma_deltas} in the paper and~\ref{fig:llama_deltas} in the appendix summarize performance deltas relative to open-source Scope baselines at the best-performing steering layers (i.e., the layers with the overall highest performance), across both machine translation (FLORES) and cross-lingual summarization (CrossSumm). For Gemma-2-9B (Figure~\ref{fig:gemma_deltas}), multilingual SAEs outperform English-only SAEs across all reported metrics, yielding substantial gains in generation quality for both tasks. In FLORES, multilingual training improves LangID and COMET while maintaining strong SpBLEU; in CrossSumm, it yields higher ROUGE-L and LaSE, indicating better content preservation and semantic alignment. For LLaMA-3.1-8B (Figure~\ref{fig:llama_deltas}), the improvements are smaller in magnitude but remain directionally consistent across tasks and SpBLEU, COMET and LaSE metrics while maintaining competitive LangID. Overall, these results demonstrate that multilingual SAEs induce more effective and semantically aligned steering directions, with consistent benefits across models and task families.



\begin{figure}[t]
    \centering
    \includegraphics[width=0.49\linewidth]{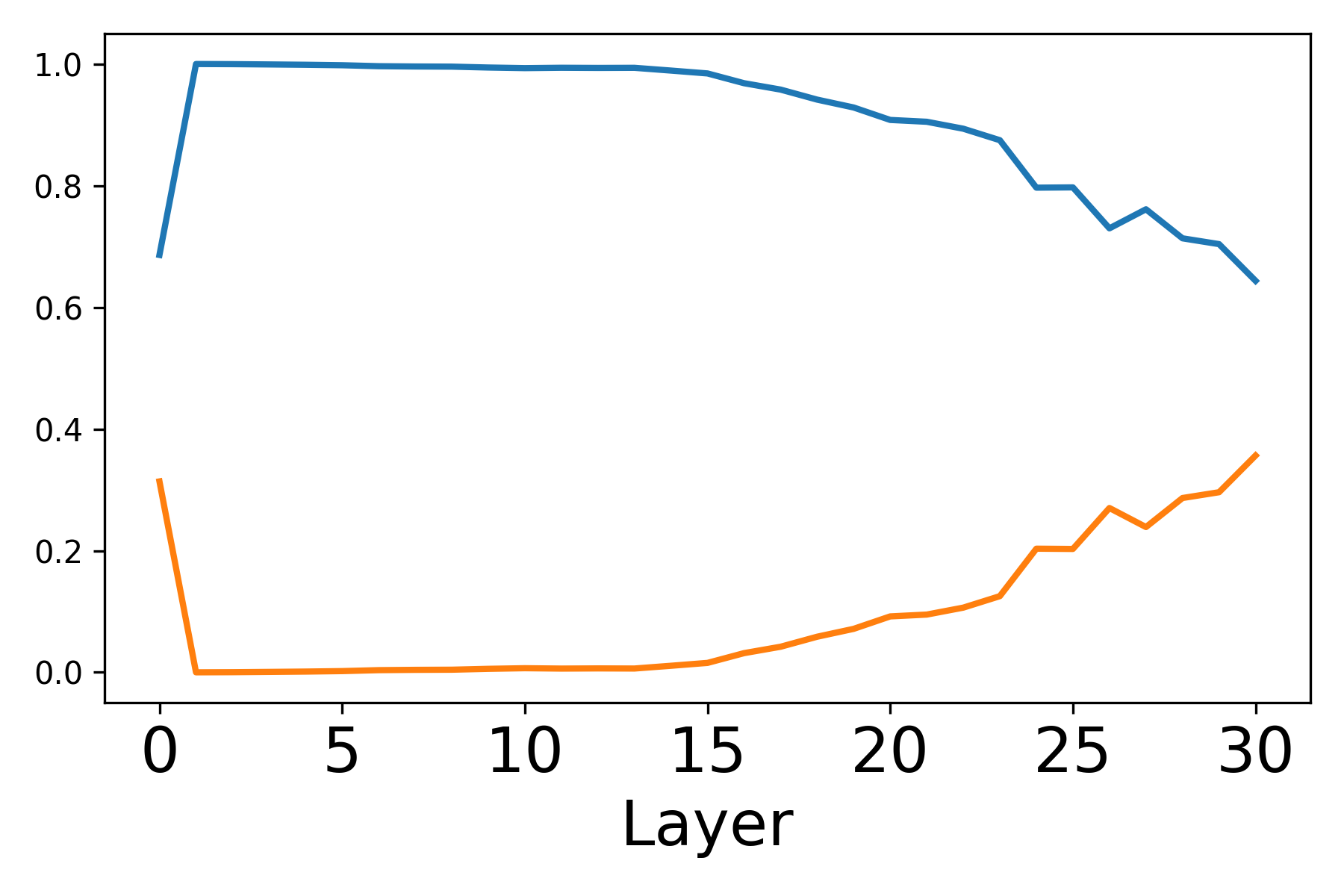}
    \includegraphics[width=0.49\linewidth]{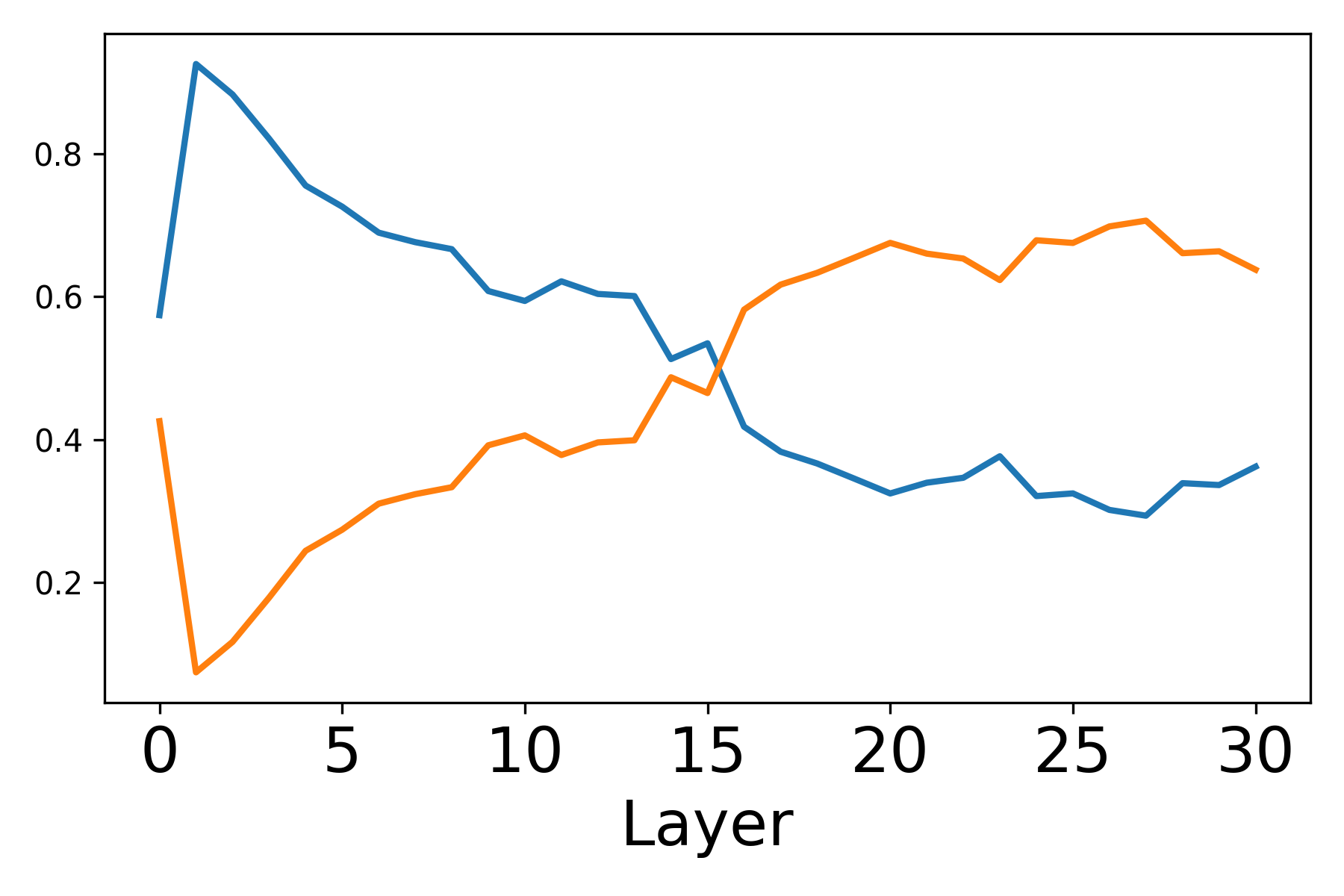} \\[2pt]
    \includegraphics[width=0.49\linewidth]{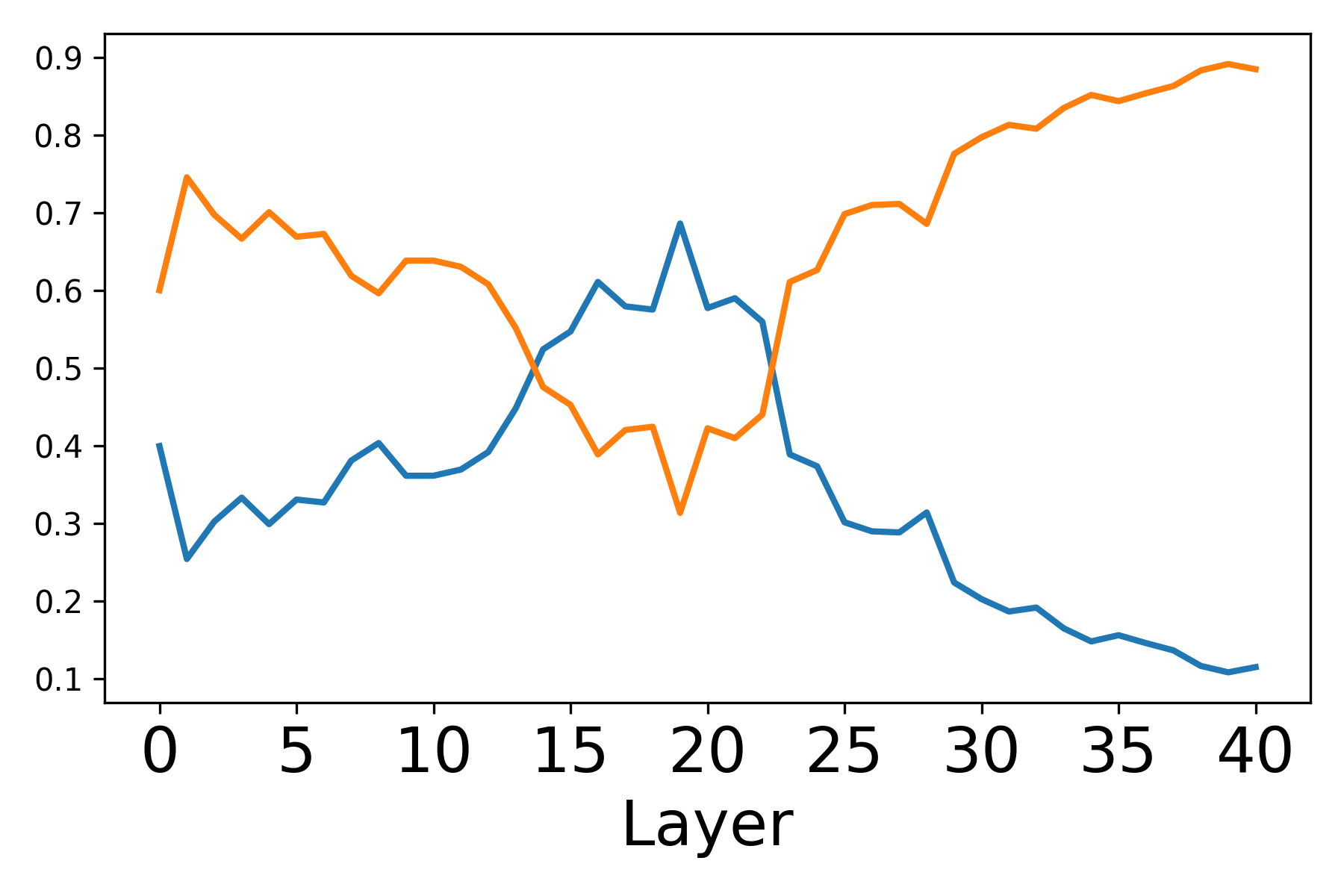}
    \includegraphics[width=0.49\linewidth]{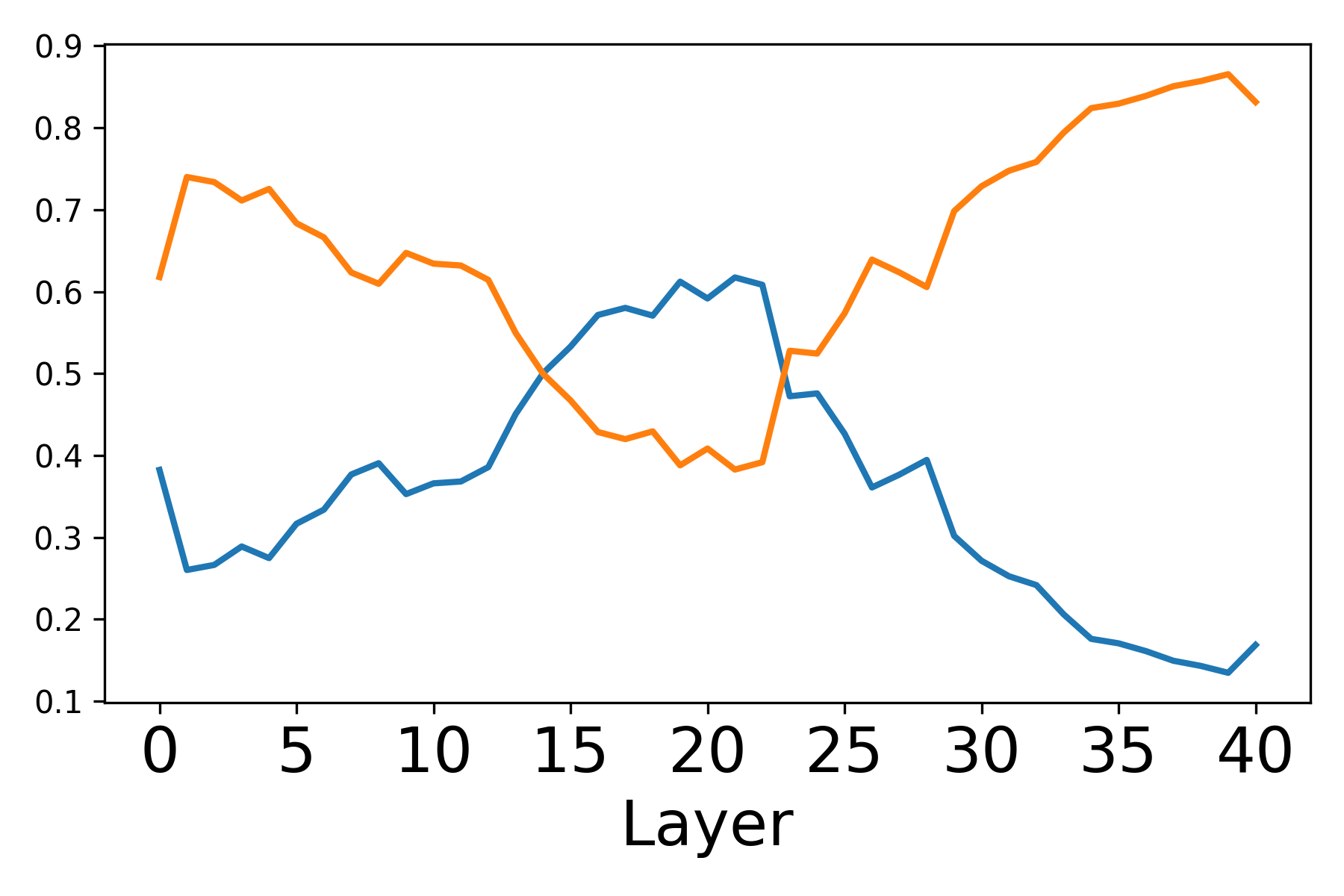}
    \caption{
Layer-selection curves showing the balance between multilingual alignment and language separability across layers.
\textbf{Blue} curves denote \emph{multilinguality} (shared cross-lingual alignment), and \textbf{orange} curves denote \emph{separability} (language-specific structure).
\textbf{Top:} LLaMA-3.1-8B. \textbf{Bottom:} Gemma-2-9B.
\textbf{Left:} Open-source SAEs (LLaMA-Scope, Gemma-Scope), where LLaMA-Scope shows no clear intersection and Gemma-Scope selects L14 and L23.
\textbf{Right:} Residual representations, which exhibit clear balance points at L15 (LLaMA) and L14/L23 (Gemma).
}

    \label{fig:layer_select_comparison}
\end{figure}

\begin{figure}[ht]

    \centering

    \includegraphics[width=1.0\linewidth]{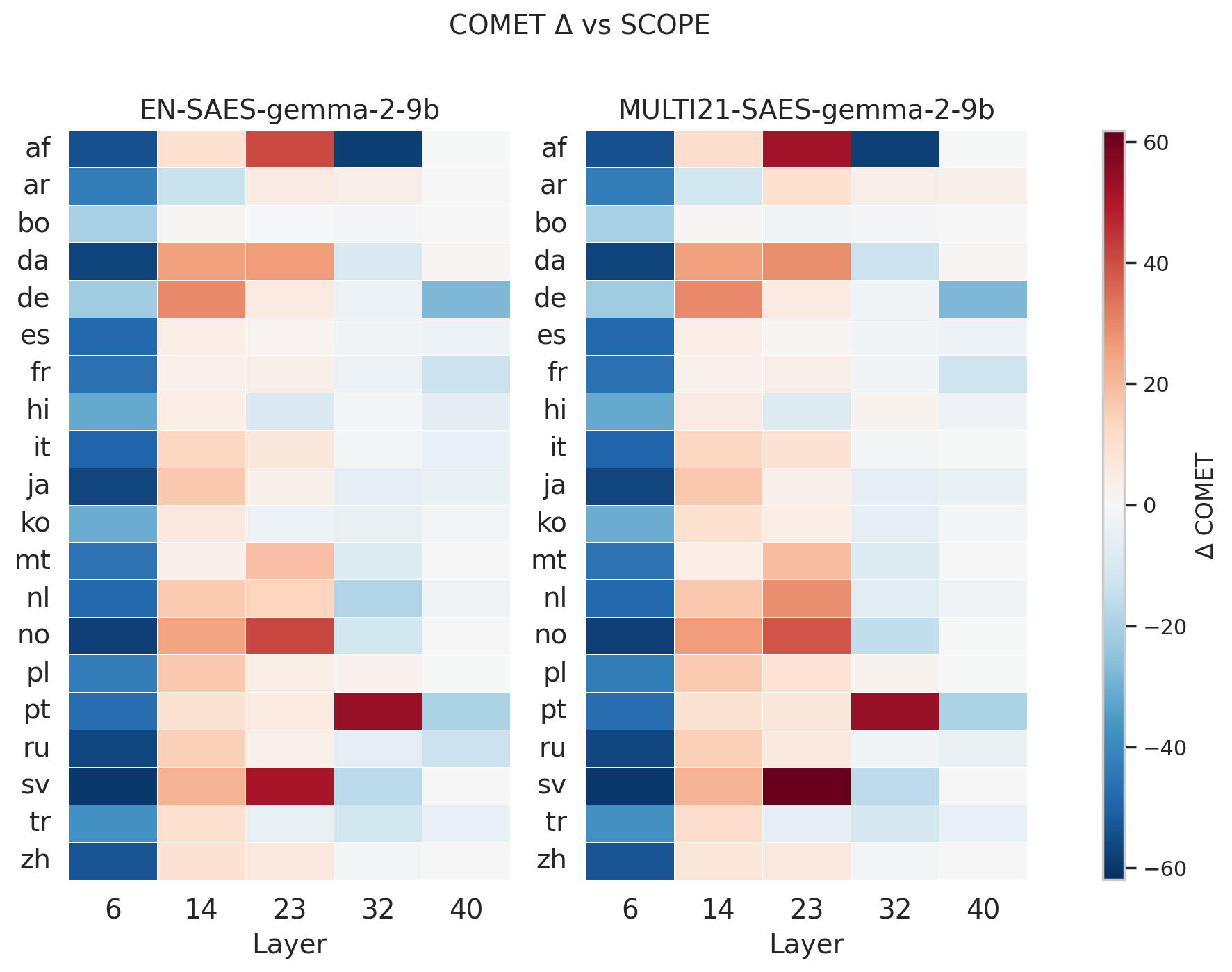}

    \caption{
Per-language, per-layer COMET score deltas for \textbf{Gemma-2-9B} on \textsc{FLORES} under cross-lingual steering (\texttt{tgt\_$i$} $\neq$ \texttt{steer\_$j$}). 
}

    \label{fig:gemma_flores_1_diff_big}
\end{figure}

\subsection{Optimal Layers}

For each model, we identify steering layers as \emph{intersection points} where multilingual alignment and language separability are jointly balanced (Figure~\ref{fig:layer_select_comparison}). Importantly, the multilinguality--separability curves are computed independently of any downstream generation metrics, making the predicted intersection layers a falsifiable, pre-intervention hypothesis.

For Gemma-2-9B, these curves exhibit a characteristic \emph{two-hump} shape, yielding intersection regions near \textbf{L14} and \textbf{L23}. Figure~\ref{fig:gemma_deltas} shows that these same layers achieve the strongest overall trade-offs between language identification accuracy and generation quality for both multilingual and English-only SAEs. Figure~\ref{fig:gemma_flores_1_diff_big} further confirms this pattern at the per-language level, where \textbf{L14} and \textbf{L23} consistently emerge as the best-performing steering depths. 

In LLaMA-3.1-8B, a pronounced increase in multilinguality near \textbf{L13} is followed by a rise in separability, yielding an intersection region spanning \textbf{L13--L15}. Figure~\ref{fig:llama_deltas} (Appendix) reports the best-performing layers across SpBLEU, COMET, LaSE, and LangID, and shows that steering within this intersection region achieves the strongest overall trade-offs. In particular, \textbf{L13} and \textbf{L15} consistently emerge as the empirically optimal steering depths across most metrics.

\begin{figure}[t]
    \centering

    \includegraphics[width=0.49\linewidth]{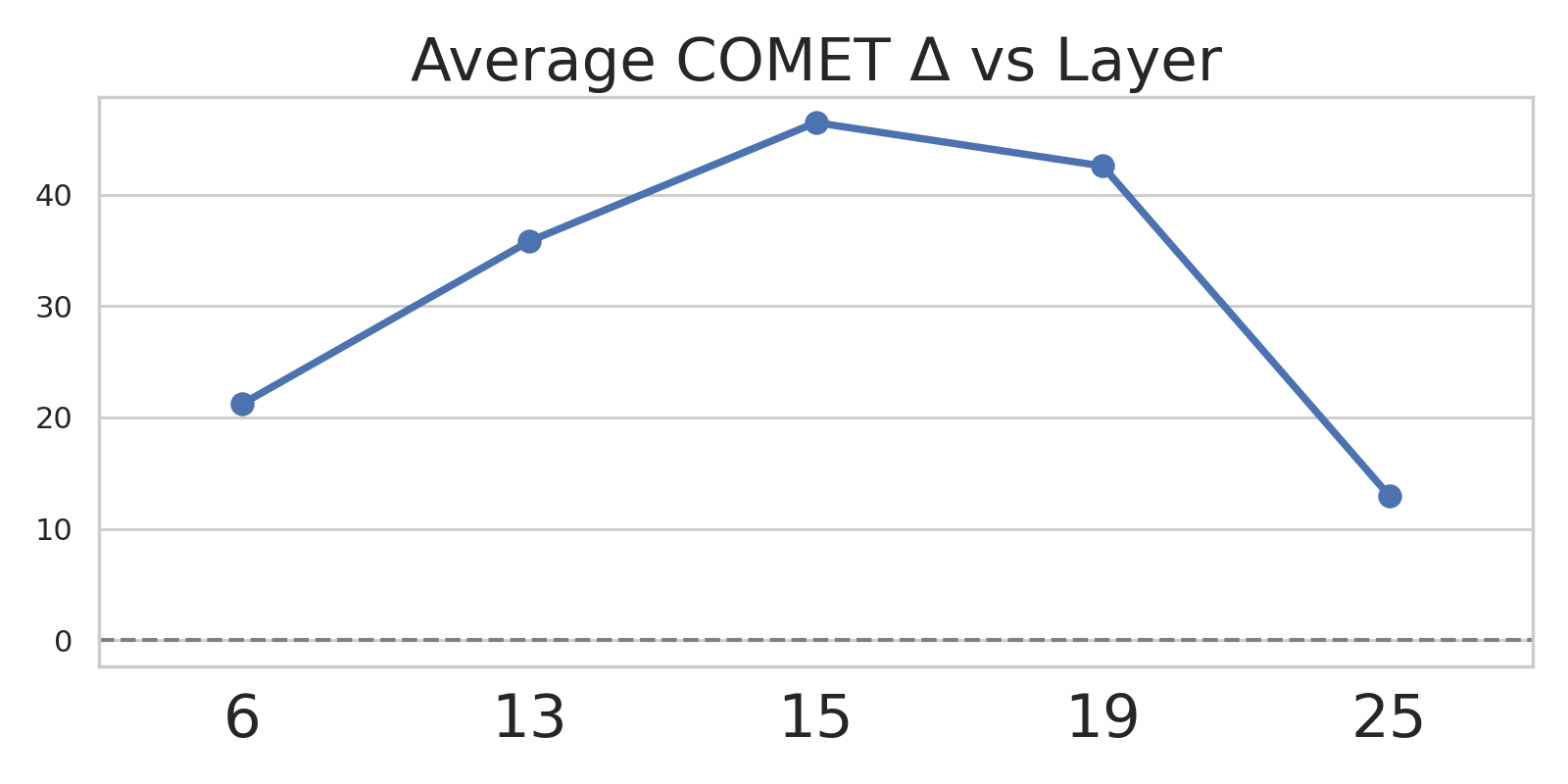}
    \includegraphics[width=0.49\linewidth]{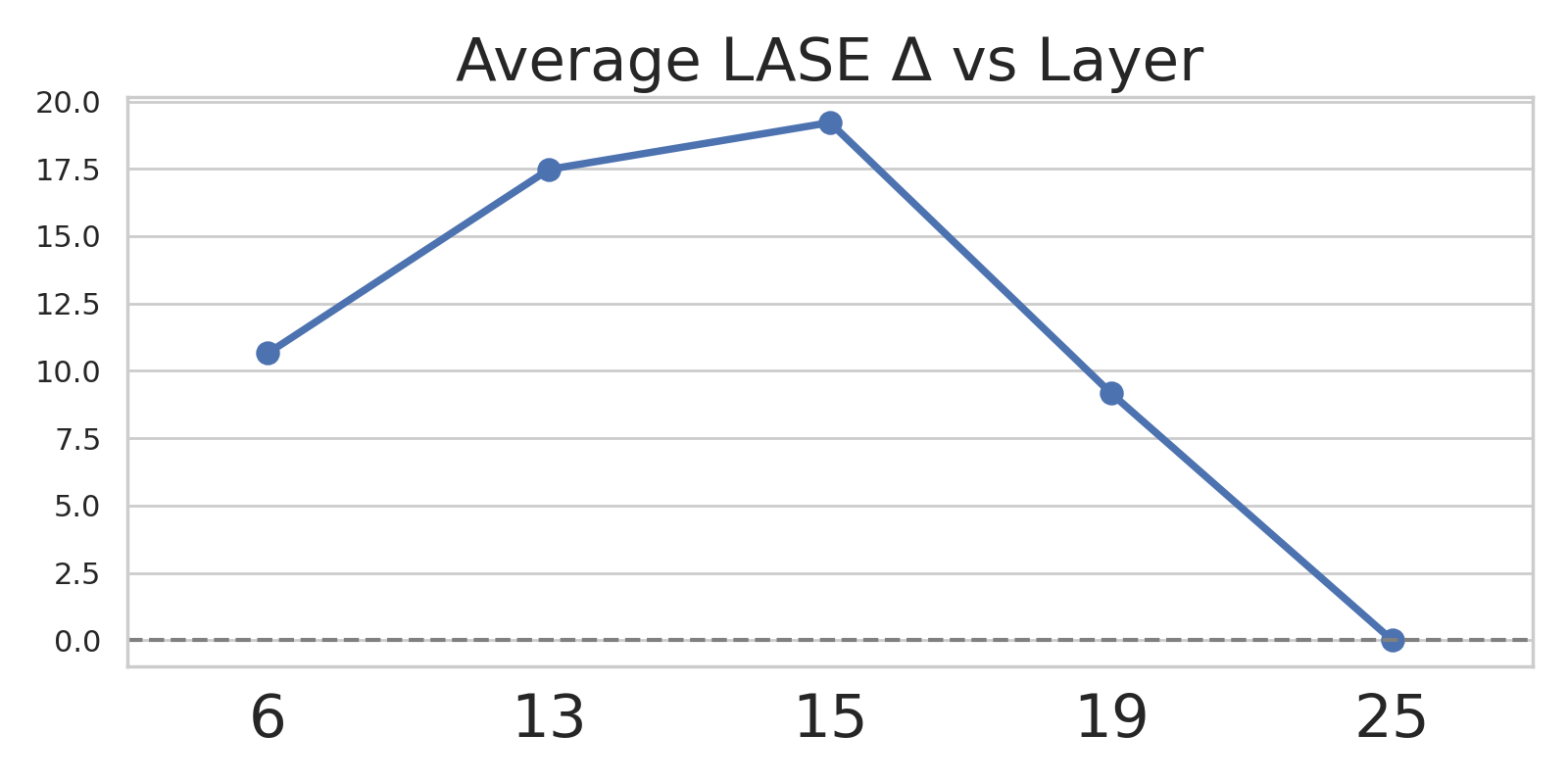} \\[2pt]
    \includegraphics[width=0.49\linewidth]{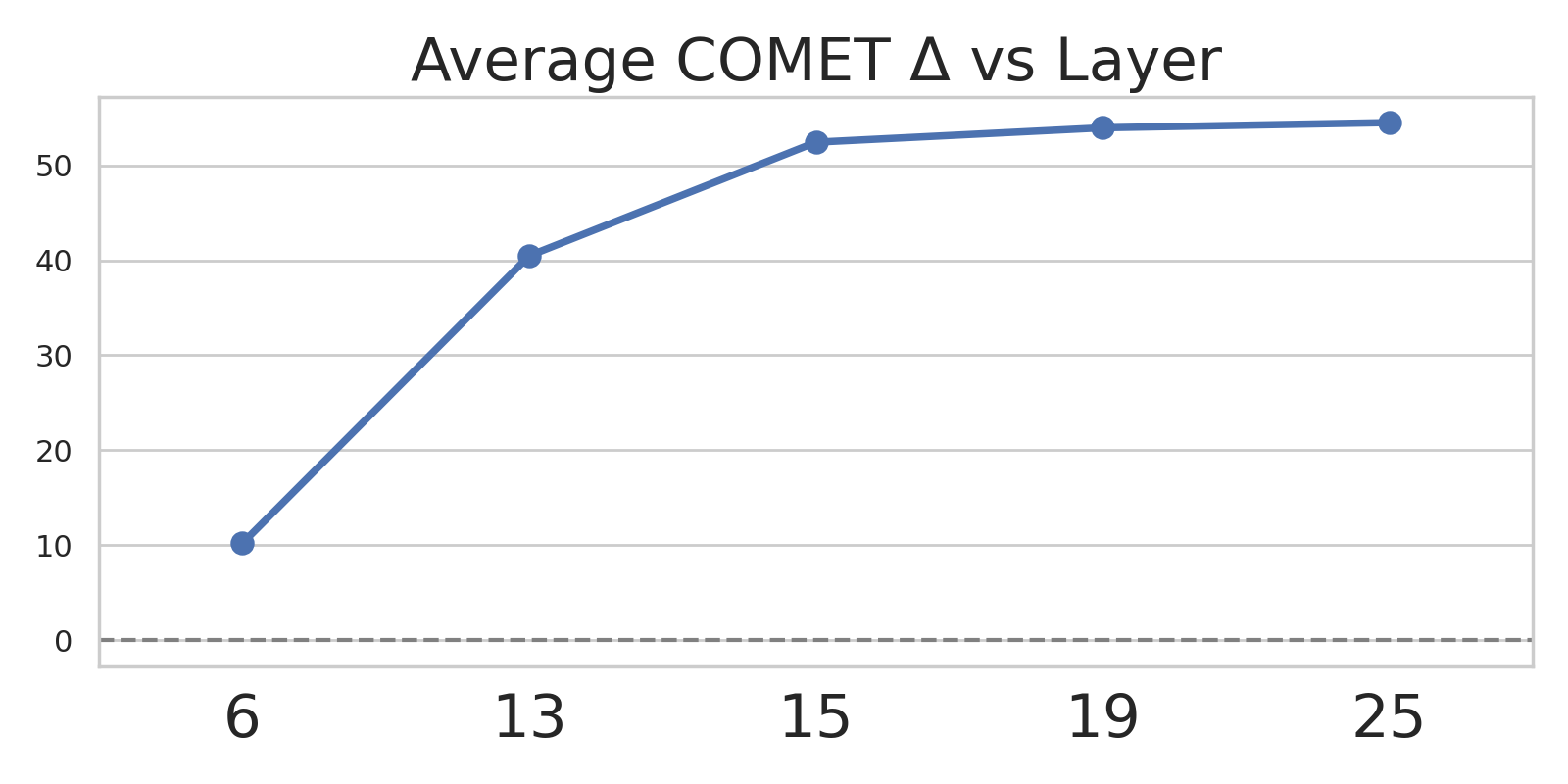}
    \includegraphics[width=0.49\linewidth]{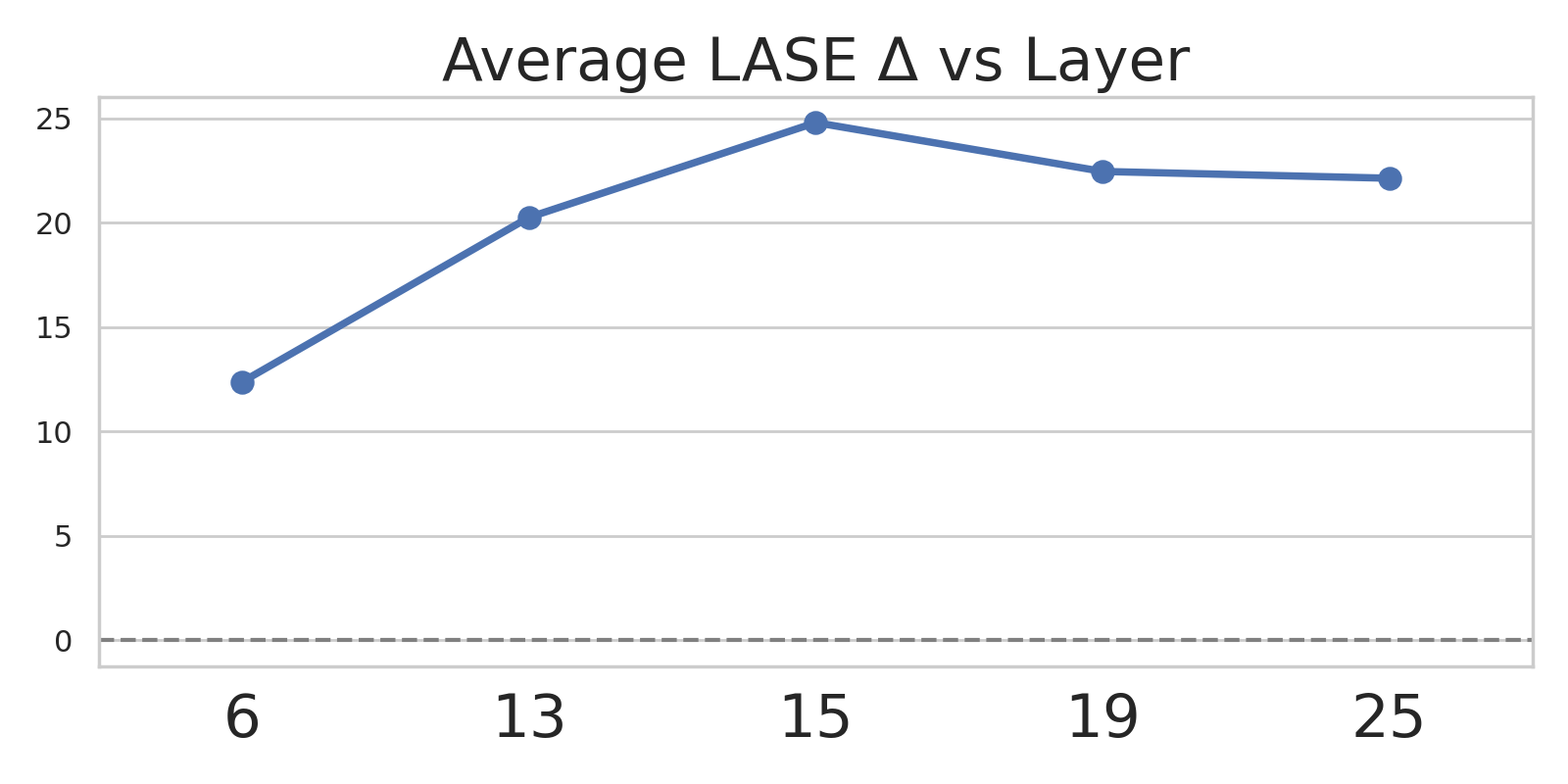}
    
    \caption{Layerwise $\Delta$COMET and $\Delta$LaSE trends for LLaMA-3.1-8B averaged across SAEs under two steering regimes. \textbf{Top:} \texttt{steer\_lang} $\neq$ \texttt{target\_lang}. \textbf{Bottom:} \texttt{steer\_lang} $=$ \texttt{target\_lang}.}
    \label{fig:layer_trend_spbleu}
\end{figure}

To further validate our layer-selection criterion, we analyze layerwise performance trends under different steering regimes. Figure~\ref{fig:layer_trend_spbleu} reports layerwise $\Delta$COMET and $\Delta$LaSE averaged across SAE variants for two settings: (i) when the steering language matches the prompted target language, $\Delta$COMET and $\Delta$LaSE increase monotonically with depth. This behavior is consistent with later layers exhibiting stronger language separability in LLaMA-3.1-8B and favoring same-language amplification. (ii) when steering toward a language different from the prompted target, performance follows a non-monotonic trend, peaking near the layers identified by our multilinguality--separability intersection.

This divergence highlights the role of representational balance: deeper layers benefit same-language reinforcement, whereas effective cross-language steering requires intervening at depths where shared cross-lingual structure is still preserved alongside language-specific distinctions. These results provide additional empirical support that the layers selected by our criterion correspond to optimal steering depths. 
Importantly, we show that layers selected by our criterion consistently outperform earlier and later layers when controlling for SAE architecture and training data.
This indicates that effective steering depth is a structural property of the base model rather than an artifact of a particular SAE. We observe a different pattern in Gemma-2-9B: same-language steering favors earlier layers, where language separability is high, while cross-language steering again peaks near the layers identified by our intersection criterion.

\subsection{Additional Experiments: Steering with Open‑Source SAEs}

We compare our multilingual SAEs against the open-source \emph{LLaMA-Scope} and \emph{Gemma-Scope} suites.  
For \emph{Gemma-Scope}, our criterion identifies two intersection layers at \textbf{L14} and \textbf{L23}. Steering performance at these depths remains consistently below that of our multilingual SAEs (Figure~\ref{fig:gemma_flores_1_diff_big}). At other layers, Gemma-Scope can exhibit competitive or occasionally stronger results, highlighting the sensitivity of multilingual steering to intervention depth and suggesting that multilingual SAEs, even when trained on comparatively less multilingual data, can surpass Gemma-Scope when applied at mechanistically appropriate layers.

In contrast, for \emph{LLaMA-Scope}, we observe consistently negligible downstream gains across layers. Applying our multilinguality-separability analysis in the sparse space reveals that LLaMA-Scope does not exhibit a meaningful intersection layer: language separability remains weak across depth (Figures~\ref{fig:layer_select_comparison} and~\ref{fig:correlation_matrices}). The absence of a balance point between shared cross-lingual alignment and language-specific structure aligns with its poor steering performance.

\begin{figure}[t]
    \centering
    \includegraphics[width=1.0\linewidth]{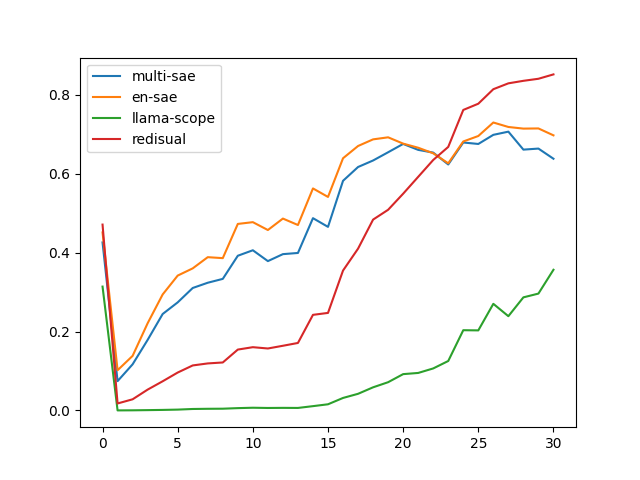}
    \caption{Separability across layers for LLaMA‑3.1‑8B, comparing different SAEs.}
    \label{fig:llama_separability}
\end{figure}

Figure~\ref{fig:llama_separability} further reproduces the early--late dynamics of multilingual representations previously reported for LLaMA-3.1-8B \citep{gurgurov2025languagearithmeticssystematiclanguage,tan-etal-2024-neuron}: shared cross-lingual structure is strongest in early-to-mid layers, while language separability increases toward later depths. Notably, LLaMA-Scope exhibits substantially lower separability than even the dense residual stream across all layers, which likely reflects the combined effects of English-skewed training data and architectural choices in the SAE design, and helps explain its failure to support effective language control despite operating at similar depths. More generally, when the separability score approaches zero, we consistently observe steering failure, indicating that separability provides a simple and predictive signal of multilingual steering capability at a specific layer (Figure~\ref{fig:correlation_matrices} in appendix).

\section{Take-Aways and Conclusion}

Our results show that reliable SAE-based multilingual steering emerges from the combination of multilingual training and principled layer selection. Crucially, our findings show that multilingual SAE steering is promising, and that its success can be predicted from representation-level structure.

\paragraph{(I) Multilingual SAE training strengthens language representations.}
Across both models, multilingual SAEs consistently outperform English-only SAEs on language identification accuracy and generation quality. These gains indicate that multilingual training does more than expand language coverage: it induces richer shared cross-lingual structure while preserving cleaner language-specific signals in the sparse feature space, yielding more reliable steering directions.

\paragraph{(II) Intersection points predict optimal steering depths.}
Balancing multilingual alignment and language separability identifies layers where language control and generation quality are jointly maximized. The intersection of these signals provides an \emph{a priori} rule for layer selection that replaces heuristic mid–late choices and avoids exhaustive layer sweeps and repeated SAE training. Across both base models, the layers identified by this criterion consistently coincide with those yielding the strongest LangID–quality trade-offs, and outperform earlier and later layers even when controlling for SAE architecture and training data.

\paragraph{(III) Open-source SAEs highlight the limits of heuristic depth choices.}
Open-source SAEs provide useful baselines but illustrate the importance of principled layer selection and multilingual training. \emph{LLaMA-Scope} does not exhibit a clear intersection between multilinguality and separability (Figure~\ref{fig:layer_select_comparison}) and yields negligible steering gains across layers. Its sparse representations show weak language separability, often worse than the dense residual stream, suggesting that English-skewed training data and architectural choices collapse multilingual features (Figure~\ref{fig:llama_separability}).

\section*{Limitations}

We evaluated two base models (LLaMA-3.1-8B and Gemma-2-9B); larger, instruction-tuned, or decoder–encoder architectures may exhibit different cross-lingual dynamics. Our evaluation focuses on automated metrics (LangID, SpBLEU, COMET, ROUGE-L, LaSE), which do not capture stylistic fidelity, code-switching behavior or robustness to ambiguous prompts. Additionally, our findings are based on JumpReLU SAEs trained on the residual stream; extending this analysis to other sparse architectures, intervention sites (e.g., attention or MLP activations), or alternative steering constructions remains an open direction. We do not claim that the intersection criterion is unique; alternative representational statistics may identify similar balance points. Future work should complement these automated evaluations with manual translation-error analysis and stronger comparisons to existing steering methods and state-of-the-art multilingual systems, in order to better characterize failure modes and clarify the remaining performance gap for SAE-based language control. Similarly, the 0.5 intersection threshold should be understood as an operational definition of equal multilingual alignment and language separability rather than as a uniquely optimal cutoff; future work should study adaptive or model-specific thresholds.

\section*{Acknowledgments}
This research was supported by the German Federal Ministry for Economic Affairs and Energy (BMWE) as part of the project \textit{“Souveräne KI für Europa (SOOFI)”} (13IPC040H), and by the German Federal Ministry of Research, Technology and Space (BMFTR) as part of the project TRAILS (01IW24005).

\bibliography{custom}

\clearpage

\appendix
\newpage

\section*{Appendix}
\section{SAE Training}
\label{app:SAEtraining}

\subsection{SAE Training Data}
Following the mechanistic motivation outlined in Section~\ref{sec:why_multilingual_sae}, we train parallel English-only and multilingual SAE suites under a controlled setup to isolate the effect of training data on language steering.
We train two SAE suites per base model using Wikipedia data \cite{wikimedia_wikipedia_dump_20231101}. For the multilingual suite (\emph{MULTI21-SAEs}), we construct a balanced corpus covering the same 21 languages, and select a total of 2.1B tokens with a uniform distribution across languages. For the English-only suite (\emph{EN-SAEs}), we select the same number of tokens (2.1B), drawn from English Wikipedia. This controlled setup ensures that both suites are trained on identical data volume with identical optimization parameters, isolating the effect of multilingual versus monolingual training data from corpus size or training duration.

\subsection{SAE Training Procedure}
\label{sec:sae-training}
For each layer of each base model, we train JumpReLU SAEs \citep{rajamanoharan2407jumping} on the residual stream, matching the expansion factors used by the corresponding open-source SAE suites (8$\times$ for LLaMA-Scope and 16k for Gemma-Scope). We use identical architectures and optimization hyperparameters for our \emph{EN-SAEs} and \emph{MULTI21-SAEs}. 
To ensure a controlled comparison, we fix the number of optimization steps and therefore the total number of parameter updates across both suites. This setup cleanly isolates the impact of multilingual training from architectural and optimization confounds.

\subsection{Hyperparameters}
We train SAEs with \texttt{SAELens}\footnote{\url{https://github.com/jbloomAus/SAELens}} using a JumpReLU architecture on the residual stream at multiple layers. The base model is loaded in \texttt{float16}, while SAE training uses \texttt{float32}. Hook sites follow \texttt{blocks.\{layer\}.hook\_resid\_post}

\paragraph{Key hyperparameters (from code).}
\begin{itemize}
    \item \textbf{Architecture:} \texttt{jumprelu} (expansion factor $=8$), $L_1$ coefficient $=5.0$, JumpReLU bandwidth $=10^{-3}$, init threshold $=10^{-3}$, decoder init = zeros, transpose encoder init, decoder heuristic init enabled, sparsity penalty scaled by decoder norm.
    \item \textbf{Training length (\#steps):} $30{,}000$.
    \item \textbf{Batch size (tokens/step):} $4{,}096$.
    \item \textbf{Context size:} $512$.
    \item \textbf{Optimizer / schedule:} Adam ($\beta_1=0.9,\;\beta_2=0.999$), LR $=5\!\times\!10^{-5}$, LR warmup $=1{,}500$ steps, LR decay steps $=3{,}000$, $L_1$ warmup $=1{,}500$ steps.
    \item \textbf{Dead/feature refresh:} feature sampling window $=2000$, dead-feature window $=1000$, dead threshold $=10^{-4}$.
    \item \textbf{Data loader:} streaming enabled, \texttt{prepend\_bos=True}.
\end{itemize}

\subsection{Computational Budget}
For all experiments we used 1× H100 80GB. Each SAE model was trained for 30,000 optimization steps with a batch size of 4,096 tokens per step, corresponding to approximately 123M training tokens and about 3 GPU hours per run.

\subsection{License and Availability}

All trained SAE checkpoints produced in this work, including both English-only and multilingual variants for \emph{LLaMA-3.1-8B} and \emph{Gemma-2-9B}, are released under the \textbf{Apache License 2.0}. This license permits use, modification, and distribution of the models for both research and commercial purposes, provided that proper attribution is maintained.  

We emphasize that the underlying base models (\emph{LLaMA-3.1-8B}, \emph{Gemma-2-9B}) remain subject to their original licenses as released by Meta and Google, respectively. Users of our checkpoints must therefore comply with both the Apache 2.0 license governing our SAEs and the terms of the corresponding base model licenses.  


\clearpage

\begin{table*}[ht]
\section{Language Labels}
\label{app:lang_labels}
\centering
\begin{tabular}{ll}
\toprule
\textbf{Language} & \textbf{Code} \\
\midrule
English & \texttt{eng\_Latn} \\
Tibetan & \texttt{bod\_Tibt} \\
Maltese & \texttt{mlt\_Latn} \\
Italian & \texttt{ita\_Latn} \\
Spanish & \texttt{spa\_Latn} \\
German & \texttt{deu\_Latn} \\
Japanese & \texttt{jpn\_Jpan} \\
Arabic & \texttt{arb\_Arab} \\
Chinese (Simplified) & \texttt{zho\_Hans} \\
Afrikaans & \texttt{afr\_Latn} \\
Dutch & \texttt{nld\_Latn} \\
French & \texttt{fra\_Latn} \\
Portuguese & \texttt{por\_Latn} \\
Russian & \texttt{rus\_Cyrl} \\
Korean & \texttt{kor\_Hang} \\
Hindi & \texttt{hin\_Deva} \\
Turkish & \texttt{tur\_Latn} \\
Polish & \texttt{pol\_Latn} \\
Swedish & \texttt{swe\_Latn} \\
Danish & \texttt{dan\_Latn} \\
Norwegian Bokmål & \texttt{nob\_Latn} \\
\bottomrule
\end{tabular}
\caption{List of 21 target languages from FLORES–200 and their language codes.}
\label{tab:flores_languages}
\end{table*}

\clearpage

\section{Formal Definitions of Language Vectors and Layer Selection}
\label{app:language_vectors}

This appendix provides the full mathematical formulation of the language vectors, steering procedure, and layer-selection metrics summarized in the main text.

\subsection{Representation Extraction}
\label{sec:sae-rep-app}

At each transformer layer $\ell$, we extract the dense hidden representation from the residual stream,
\[
h_{\ell}(x) \in \mathbb{R}^D,
\]
for input $x$. To obtain a sparse and interpretable representation, we apply an encoder--decoder sparse autoencoder (SAE) trained at layer $\ell$. The encoder maps dense activations to a high-dimensional sparse code,
\[
z_{\ell}(x) = \mathrm{Encoder}_{\ell}(h_{\ell}(x)), z_{\ell}(x) \in \mathbb{R}^K,\; K \gg D
\]

and the decoder reconstructs the activation,
\[
\hat{h}_{\ell}(x) = \mathrm{Decoder}_{\ell}(z_{\ell}(x)).
\]
Sparsity is enforced via the SAE objective, yielding sparse codes that isolate a small number of active features for each input.

\subsection{DiffMean Steering Vectors}

We construct language steering vectors using the DiffMean method \citep{wu2025axbench}. For a given target language at layer $\ell$, let $\mathcal{Z}^+$ denote the set of sparse codes corresponding to examples in the target language, and $\mathcal{Z}^-$ the set corresponding to all other languages. We compute the mean sparse representations
\[
\bar z^+_{\ell} = \frac{1}{|\mathcal Z^+|} \sum_{z \in \mathcal Z^+} z,
\qquad
\bar z^-_{\ell} = \frac{1}{|\mathcal Z^-|} \sum_{z \in \mathcal Z^-} z,
\]
and define the steering vector as
\[
w_{\mathrm{DiffMean}}(\ell) = \bar z^+_{\ell} - \bar z^-_{\ell}.
\]

This vector amplifies features that are characteristic of the target language while suppressing features shared with other languages. Prior work applies DiffMean directly in the dense residual stream; in contrast, we primarily apply it in the SAE sparse space, which yields more disentangled and controllable steering directions.

\subsection{Inference-Time Steering}

Given a hidden activation $h_{\ell}(x)$ at inference time, we apply steering as follows:
\begin{enumerate}
    \item Encode the activation into sparse space:
    \[
    z_{\ell}(x) = \mathrm{Encoder}_{\ell}(h_{\ell}(x)).
    \]
    \item Apply the steering vector:
    \[
    z'_{\ell}(x) = z_{\ell}(x) + \alpha \, w_{\mathrm{DiffMean}}(\ell),
    \]
    where $\alpha$ controls steering strength. We use fixed steering coefficients for all test examples within each model setting, with $\alpha = 5.0$ for LLaMA and $\alpha = 100.0$ for Gemma. These values were chosen in preliminary experiments as conservative values that improved target-language identification, and were fixed before final evaluation; they were not tuned per language, layer, or test example.
    \item Decode back to dense space:
    \[
    \hat h'_{\ell}(x) = \mathrm{Decoder}_{\ell}(z'_{\ell}(x)).
    \]
    \item Correct for reconstruction error by adding the residual:
    \[
    \tilde h_{\ell}(x) =
    \hat h'_{\ell}(x) +
    \big(h_{\ell}(x) - \mathrm{Decoder}_{\ell}(z_{\ell}(x))\big).
    \]
\end{enumerate}
The corrected activation $\tilde h_{\ell}(x)$ is then passed to subsequent layers. This procedure preserves the original activation outside the SAE subspace while applying a targeted intervention along the language direction.

\section{Language Correlation and Intersection-Based Layer Selection}
\label{app:layer_selection}

\subsection{Per-Language Contrast Vectors}

For each language $i$ and layer $\ell$, we construct a contrast vector using DiffMean. Let $\mathcal{H}_i^+$ denote dense codes from language $i$, and $\mathcal{H}_i^-$ dense codes from all other languages. The per-language vector is
\[
\mathbf{v}_i =
\frac{1}{|\mathcal{H}_i^+|} \sum_{h \in \mathcal{H}_i^+} h
-
\frac{1}{|\mathcal{H}_i^-|} \sum_{h \in \mathcal{H}_i^-} h.
\]
These vectors represent languages in a shared feature space by emphasizing language-specific features and suppressing shared ones.

\subsection{Correlation Matrix Across Languages}

Given the set of language vectors $\{\mathbf{v}_i\}_{i=1}^N$ at layer $\ell$, where $N$ is the number of languages, we compute a pairwise Pearson correlation matrix
\[
C_{\ell} \in \mathbb{R}^{N \times N},
\qquad
C_{ij} = \mathrm{corr}(\mathbf{v}_i, \mathbf{v}_j).
\]
This matrix captures how similarly different languages are represented at a given depth.

\subsection{Multilinguality and Separability Metrics}

Let $\{\lambda_j\}_{j=1}^N$ be the eigenvalues of $C_{\ell}$. We define the \emph{multilinguality} score as the explained-variance ratio of the first principal component,
\[
f_{\ell} = \frac{\max_j \lambda_j}{\sum_{k=1}^N \lambda_k},
\]
which measures the degree of shared alignment across languages. We define \emph{separability} as the complementary quantity
\[
s_{\ell} = 1 - f_{\ell},
\]
which reflects how distinct the language representations remain.

\subsection{Intersection-Based Layer Selection}

We select steering layers at depths where multilinguality and separability are balanced. Since $s_{\ell} = 1 - f_{\ell}$, an intersection occurs when $f_{\ell} \approx 0.5$, or equivalently when $2f_{\ell} - 1$ changes sign between adjacent layers. In practice, we detect these sign changes with a small tolerance and linearly interpolate between layer indices. These intersection points serve as \emph{a priori} candidates for effective steering depths and consistently correspond to layers that yield strong language control while preserving generation quality.

\clearpage

\begin{figure*}[t]
\section{CrossSum Prompts}
\label{app:cross_sum_prompts}
    \centering
    \includegraphics[width=0.8\linewidth]{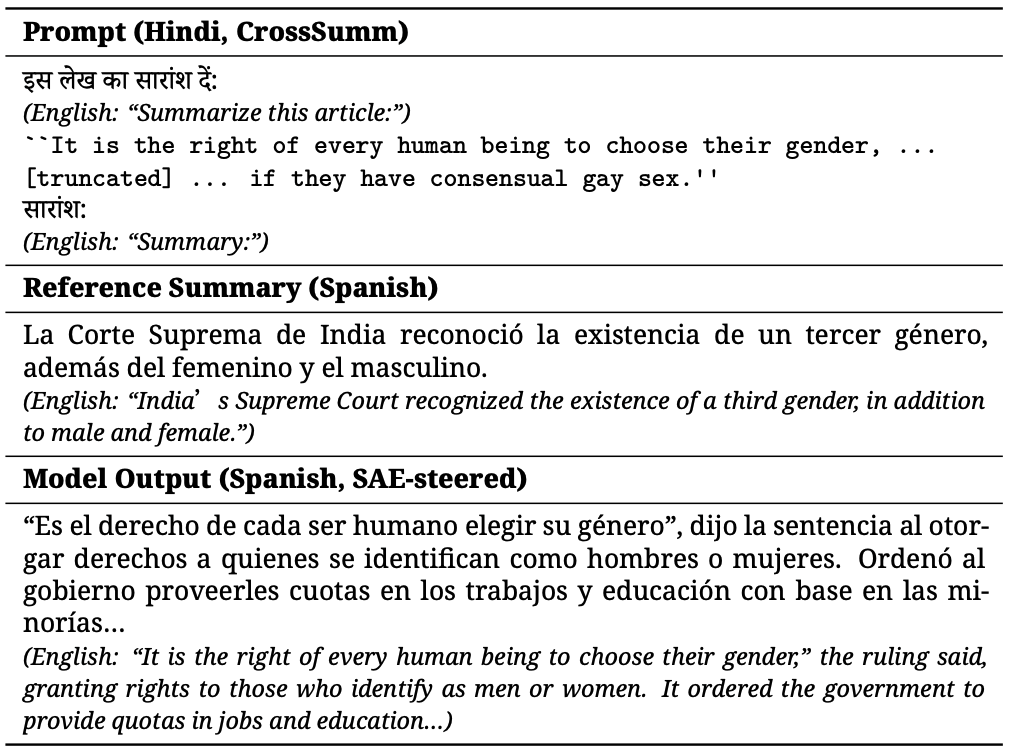}
    \caption{Example prompt and outputs for cross-lingual summarization (CrossSum). The model is prompted in Hindi and steered to generate a Spanish summary. English glosses are provided for clarity. Full article text omitted for readability.}
    \label{app:hindi_images}
\end{figure*}

\begin{figure*}[t]
    \centering
    \includegraphics[width=0.8\linewidth]{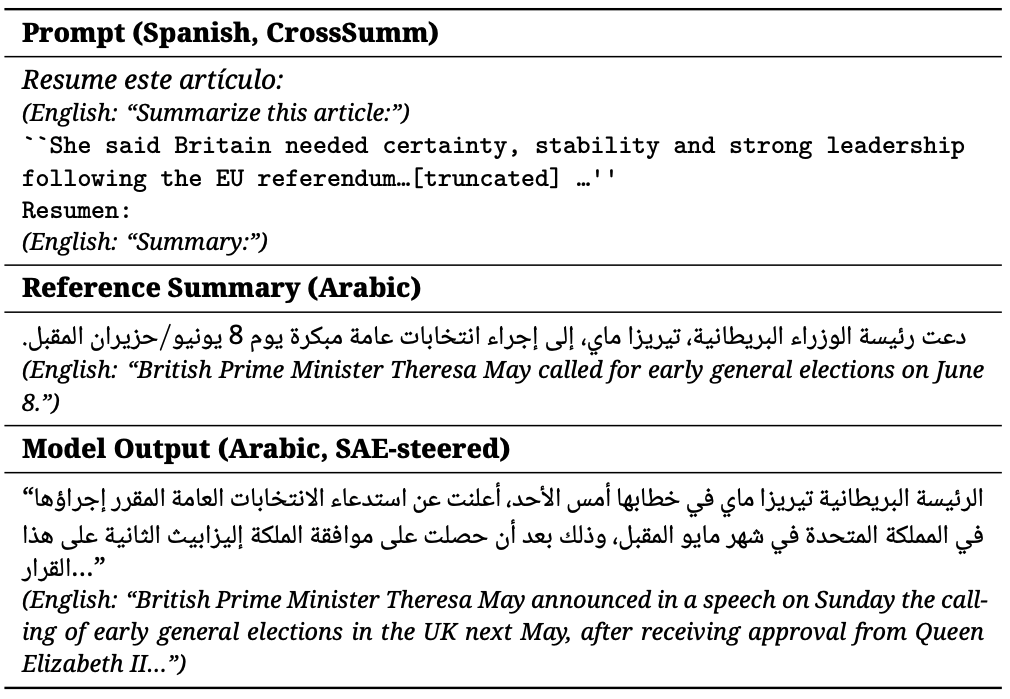}
    \caption{Example prompt and outputs for cross-lingual summarization (CrossSum). The model is prompted in Spanish and steered to generate an Arabic summary. English glosses are provided for clarity. Full article text omitted for readability.}
    \label{app:hindi_images}
\end{figure*}

\begin{figure*}[t]
\section{Flores Prompts}
\label{app:flores_prompts}
    \centering
    \includegraphics[width=0.8\linewidth]{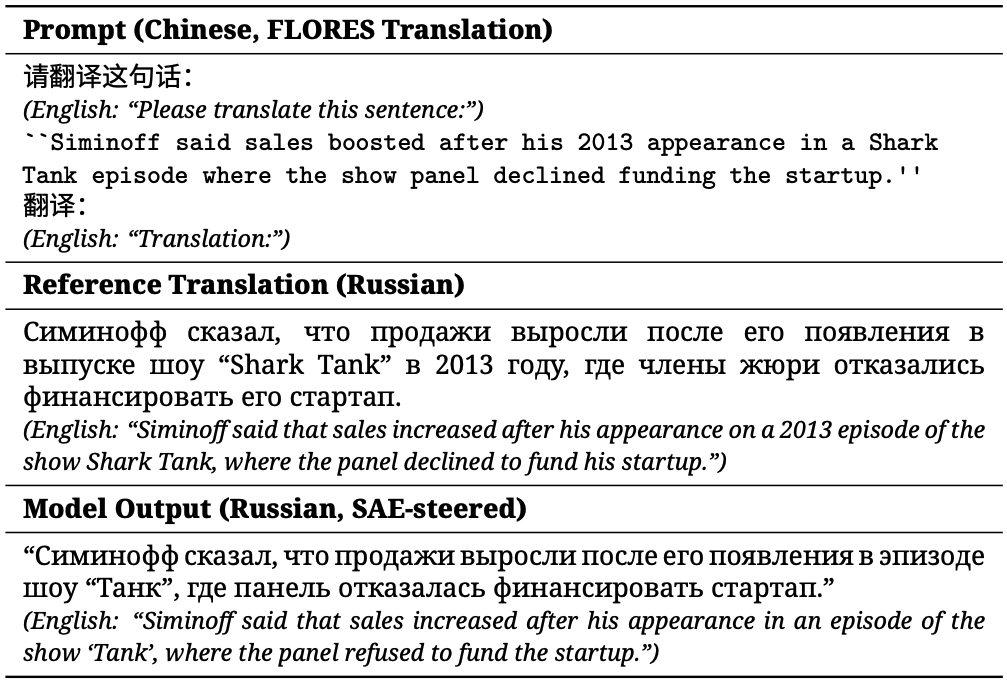}
    \caption{Example prompt and outputs for machine translation. The model is prompted in Chinese and steered to generate a Russian translation. English glosses are provided for readability.}
    \label{app:hindi_images}
\end{figure*}

\begin{figure*}[t]
    \centering
    \includegraphics[width=0.8\linewidth]{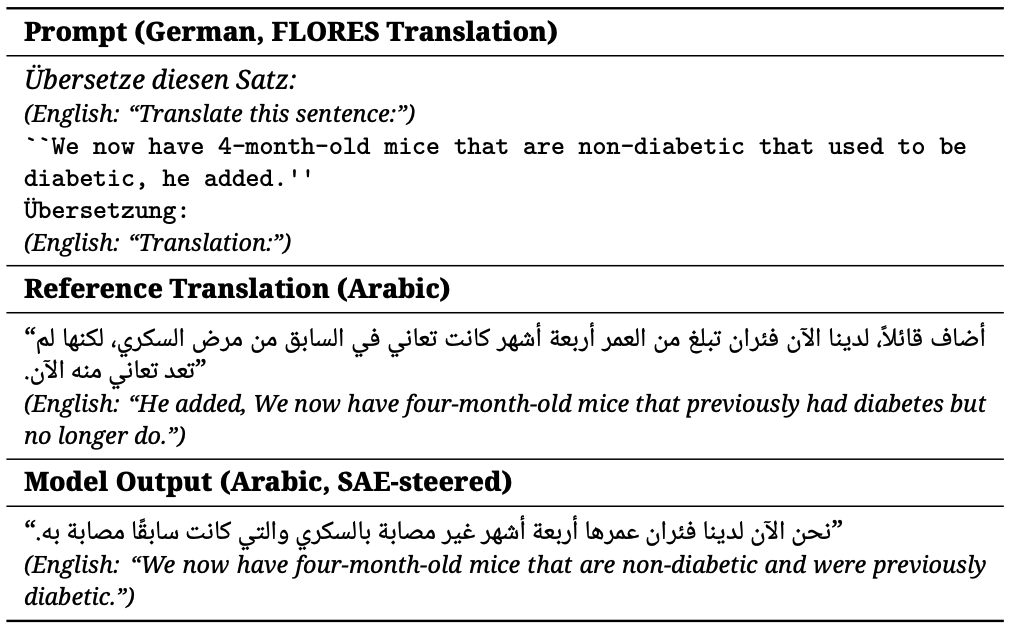}
    \caption{Example prompt and outputs for machine translation. The model is prompted in German and steered to generate an Arabic translation. English glosses are provided for readability.}
    \label{app:hindi_images}
\end{figure*}

\clearpage

\begin{figure*}[t]
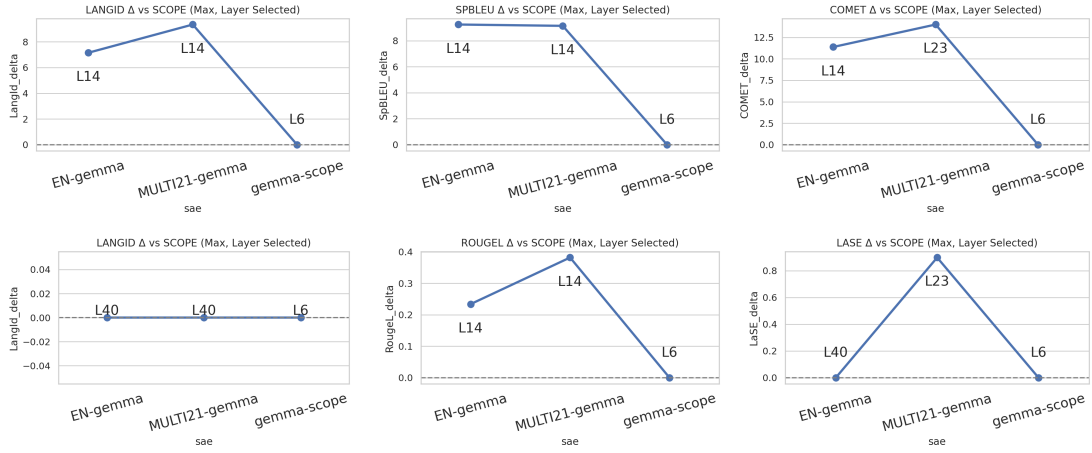

\section{Multilingual Results for Gemma-2-9B}

    \centering

    \includegraphics[width=0.3\linewidth]{assets/new_exp/plots/flores/gemma/diff/sae_point_LangId_delta_max.png}
    \includegraphics[width=0.3\linewidth]{assets/new_exp/plots/flores/gemma/diff/sae_point_SpBLEU_delta_max.png}
    \includegraphics[width=0.3\linewidth]{assets/new_exp/plots/flores/gemma/diff/sae_point_COMET_delta_max.png}  \\
    
    \includegraphics[width=0.3\linewidth]{assets/new_exp/plots/cross_sum/gemma/diff/sae_point_LangId_delta_max.png}
    \includegraphics[width=0.3\linewidth]{assets/new_exp/plots/cross_sum/gemma/diff/sae_point_RougeL_delta_max.png}
    \includegraphics[width=0.3\linewidth]{assets/new_exp/plots/cross_sum/gemma/diff/sae_point_LaSE_delta_max.png}

    \caption{Performance deltas relative to Scope baselines for \textbf{Gemma-2-9B} at the selected steering layer. \textbf{Top:} FLORES machine translation (LangID, SpBLEU, COMET). \textbf{Bottom:} Cross-lingual summarization (LangID, ROUGE-L, LaSE). Improvements from multilingual training are smaller than in LLaMA but remain directionally consistent across tasks and metrics.}
    \label{fig:gemma_deltas2}
\end{figure*}

\begin{figure*}[t]

    \centering

    \includegraphics[width=0.3\linewidth]{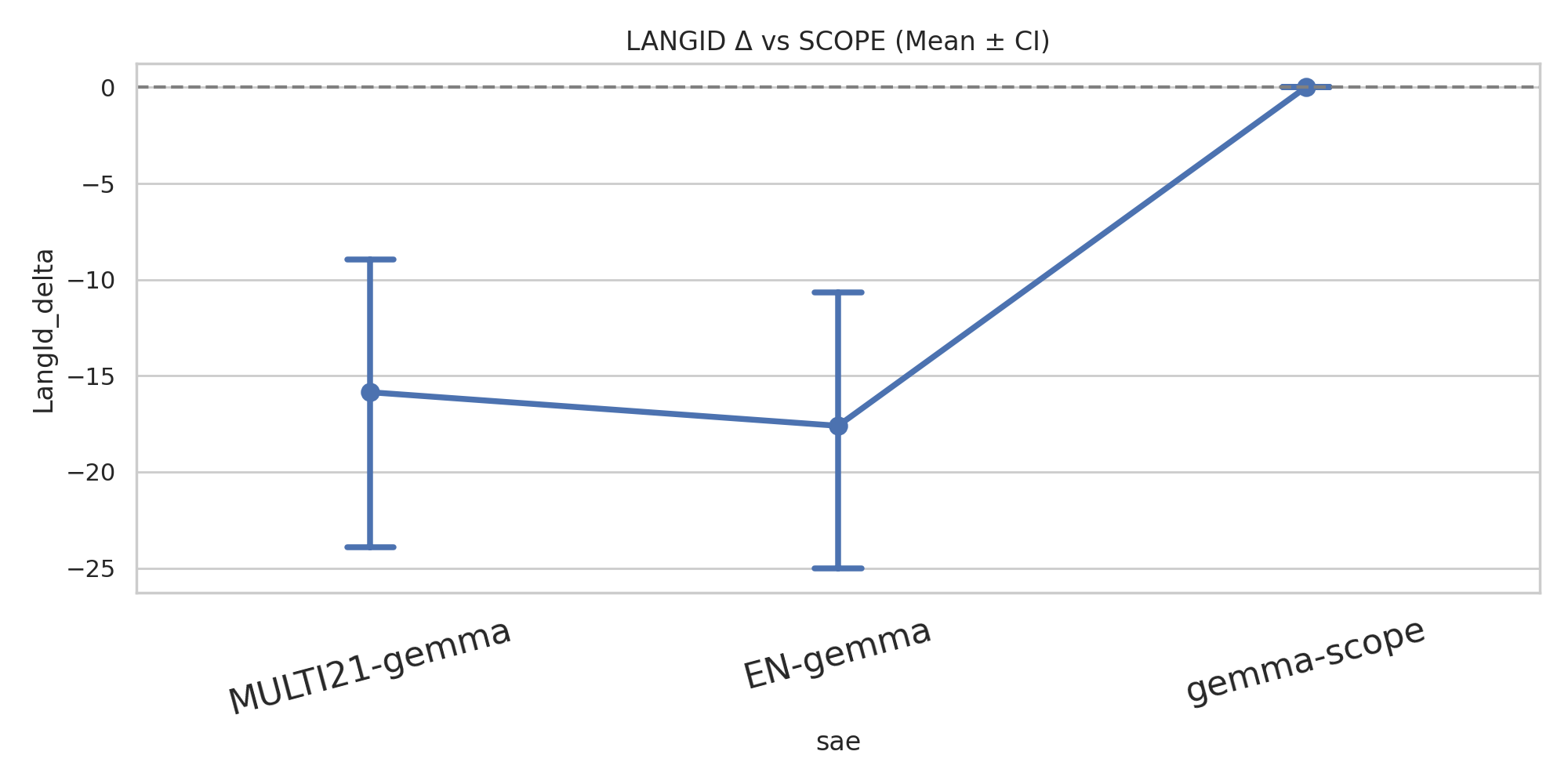}
    \includegraphics[width=0.3\linewidth]{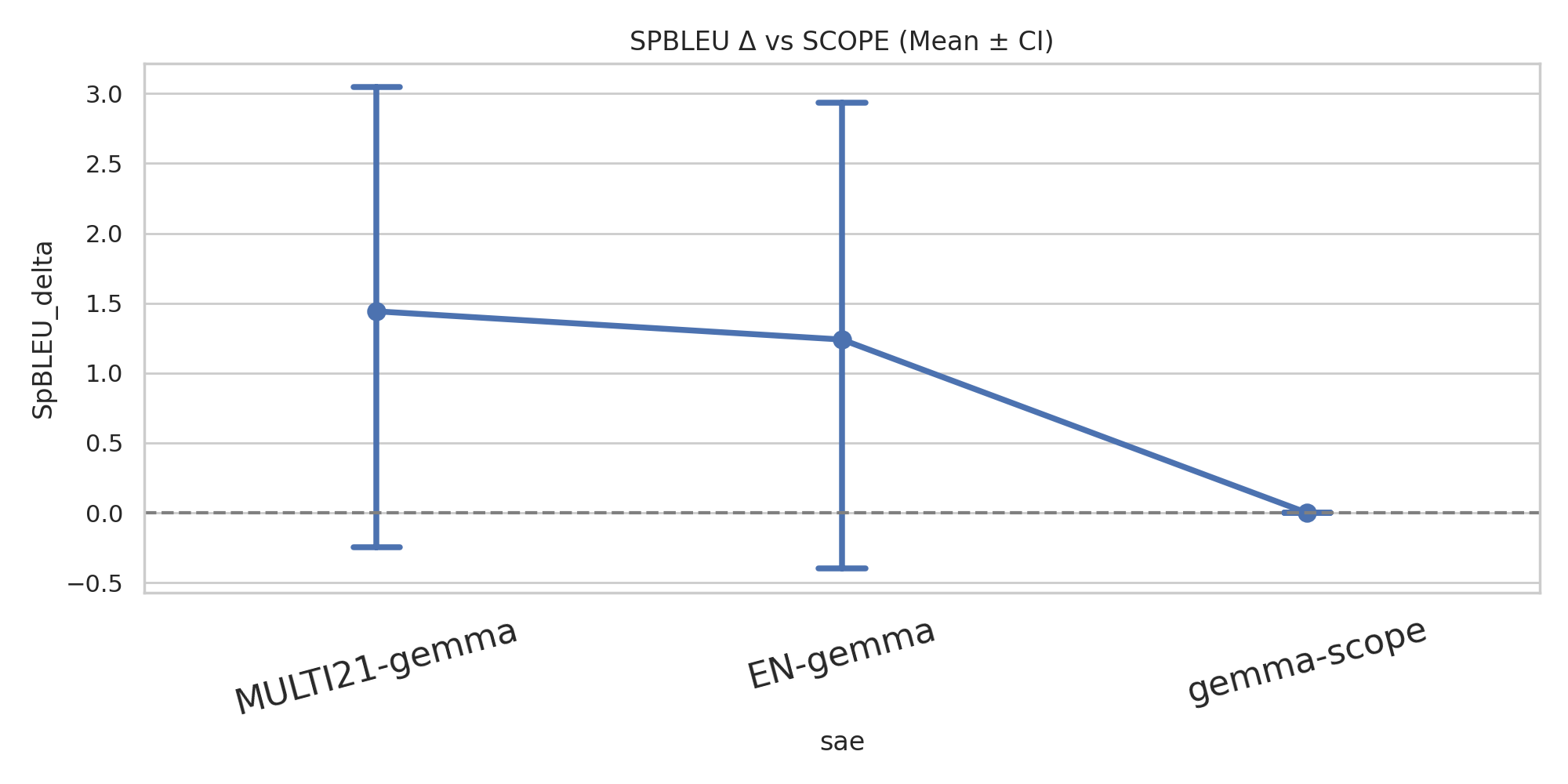}
    \includegraphics[width=0.3\linewidth]{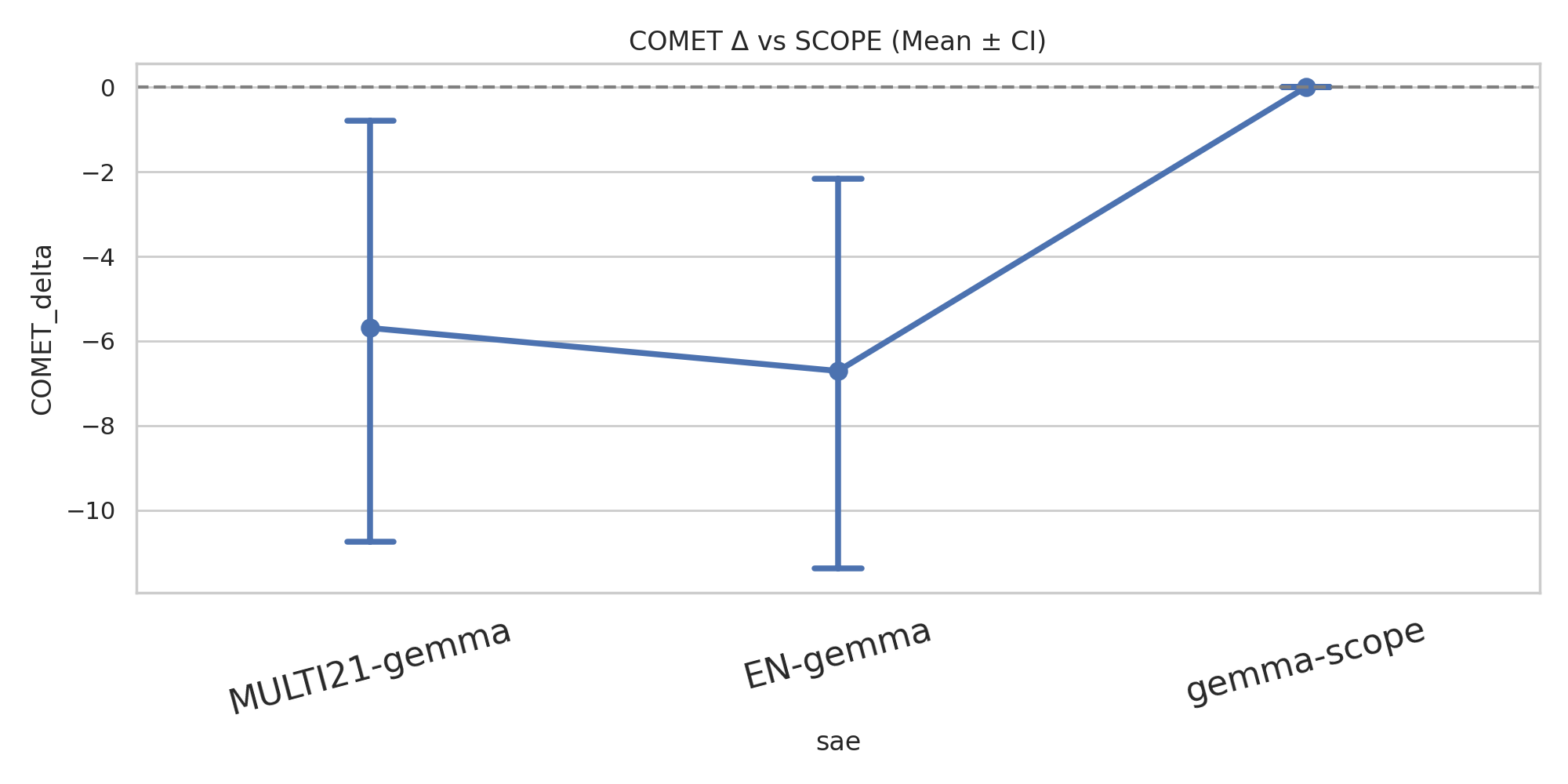}  \\
    
    \includegraphics[width=0.3\linewidth]{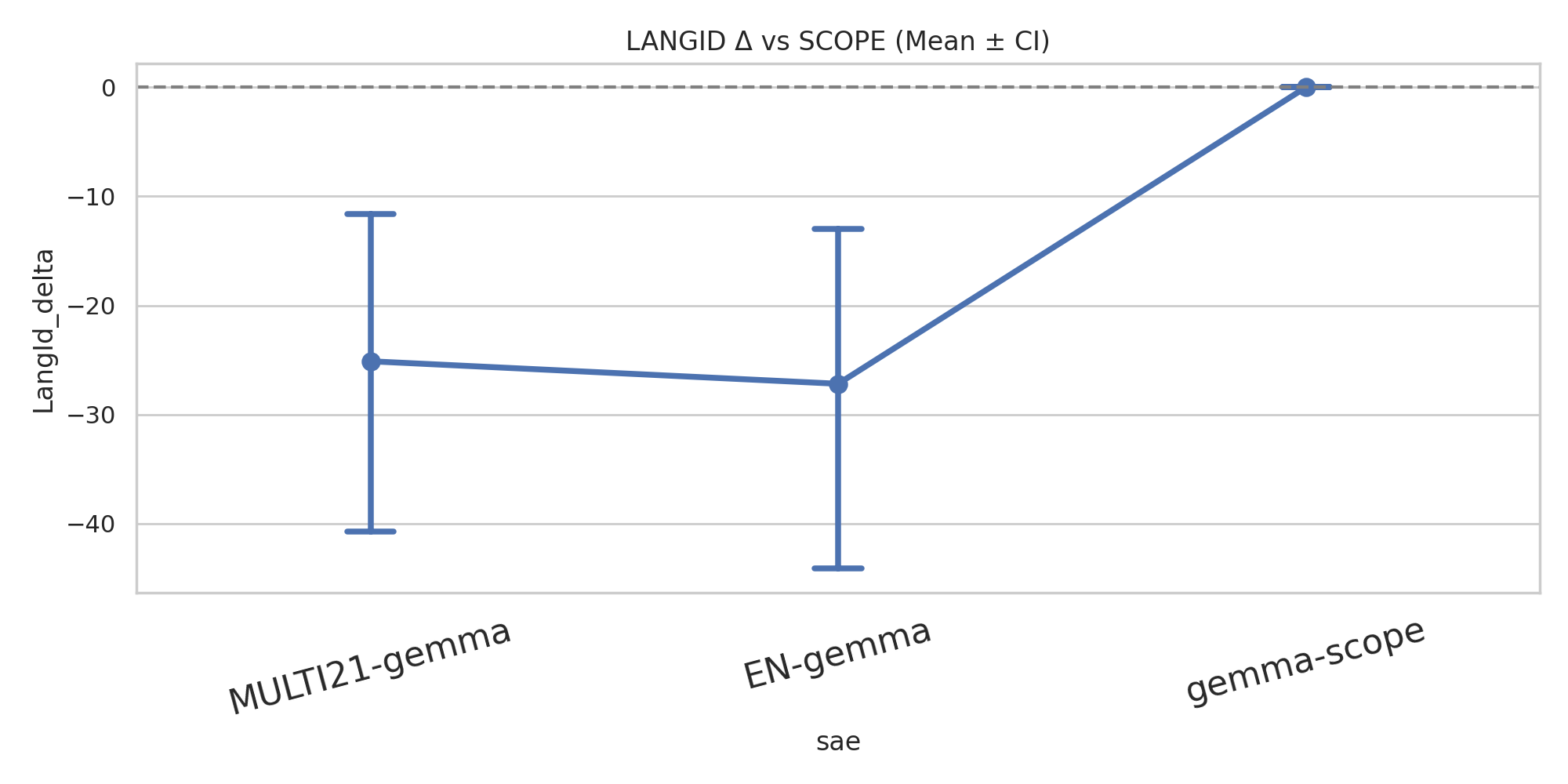}
    \includegraphics[width=0.3\linewidth]{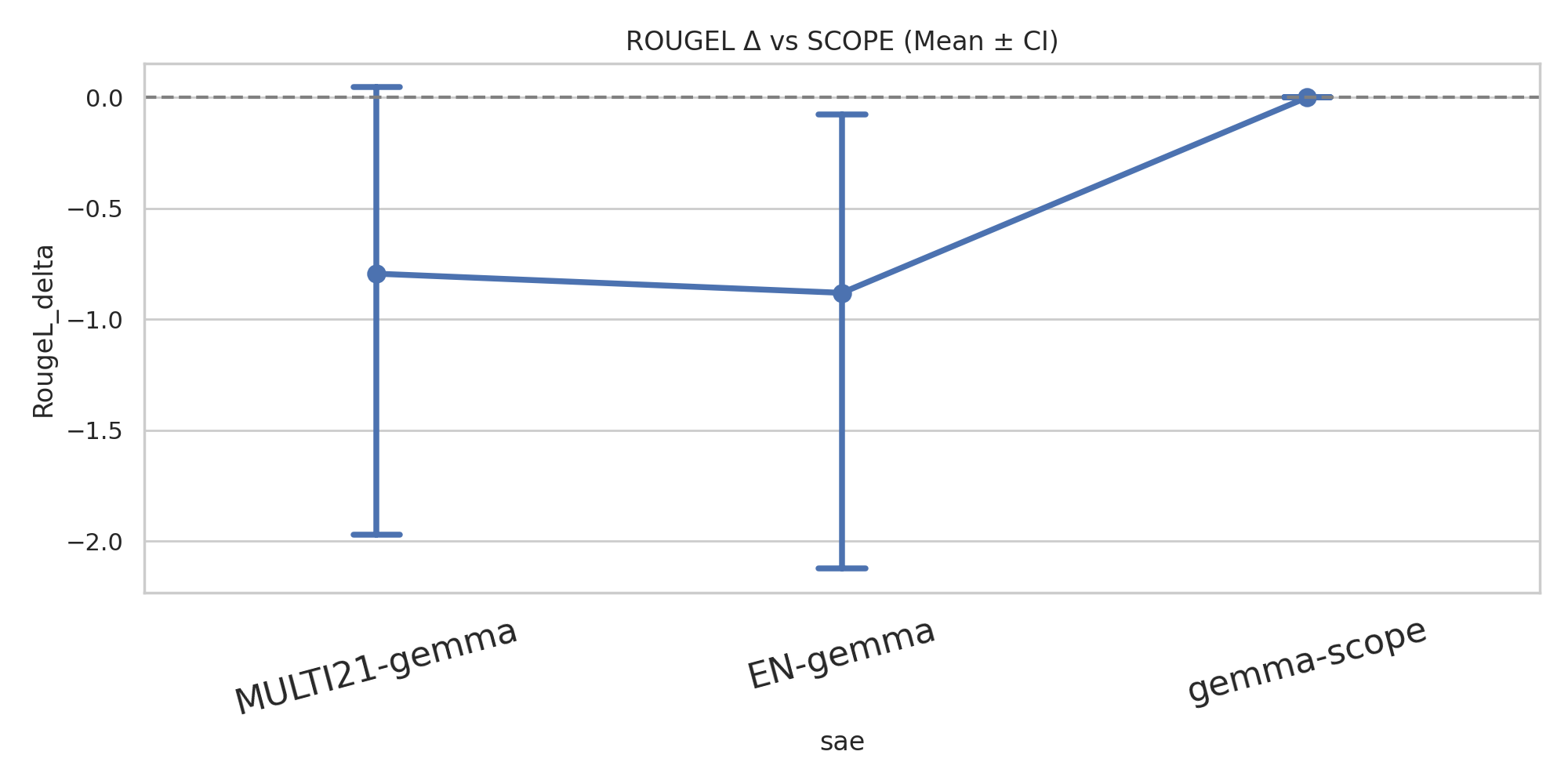}
    \includegraphics[width=0.3\linewidth]{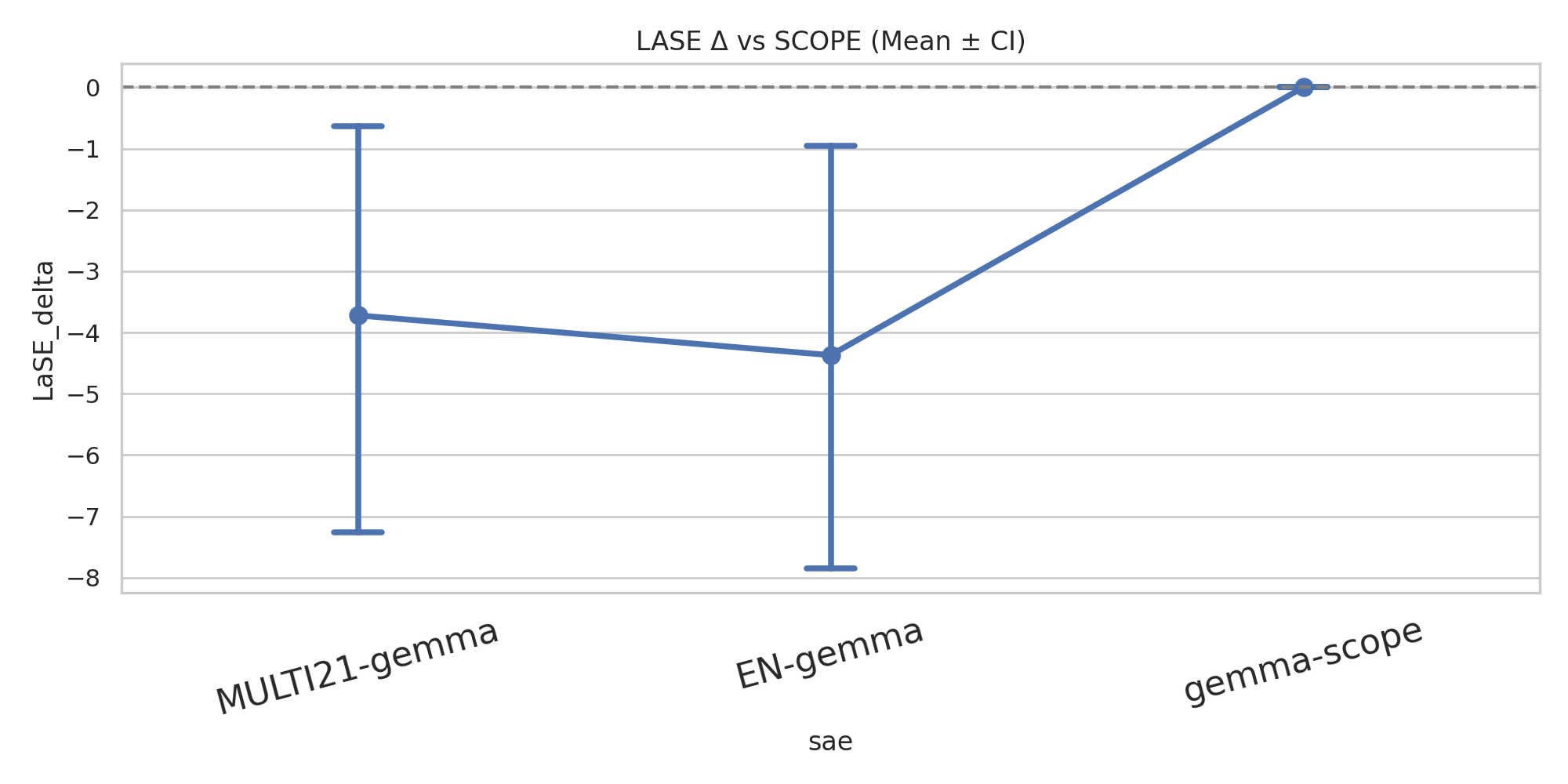}

    \caption{Performance deltas relative to Scope baselines for \textbf{Gemma-2-9B} averaged across layers. \textbf{Top:} FLORES machine translation (LangID, SpBLEU, COMET). \textbf{Bottom:} Cross-lingual summarization (LangID, ROUGE-L, LaSE).}
    \label{fig:gemma_deltas2_avg}
\end{figure*}

\begin{figure*}[t]

    \centering
    \includegraphics[width=0.9\linewidth]{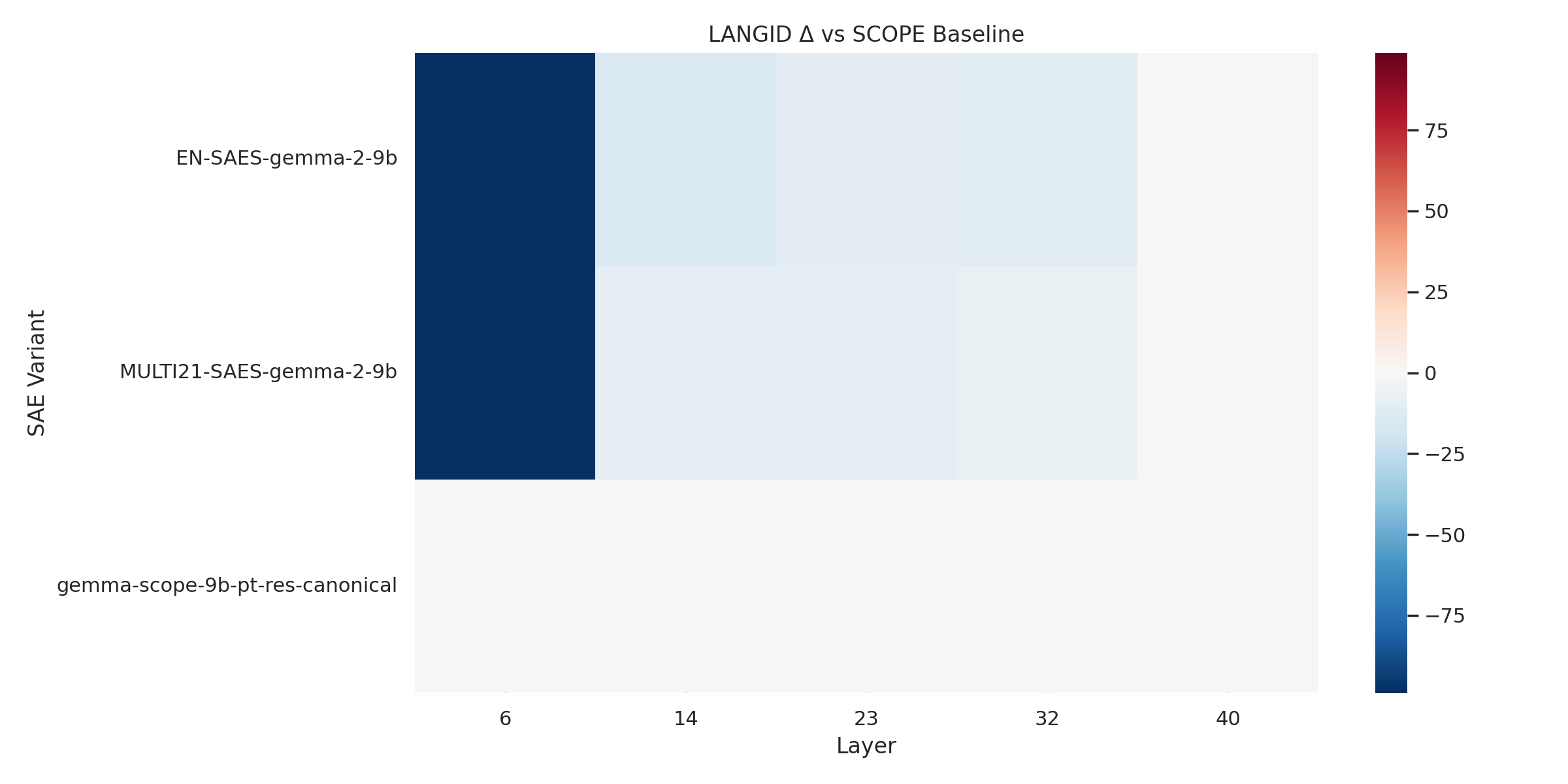}
    \includegraphics[width=0.9\linewidth]{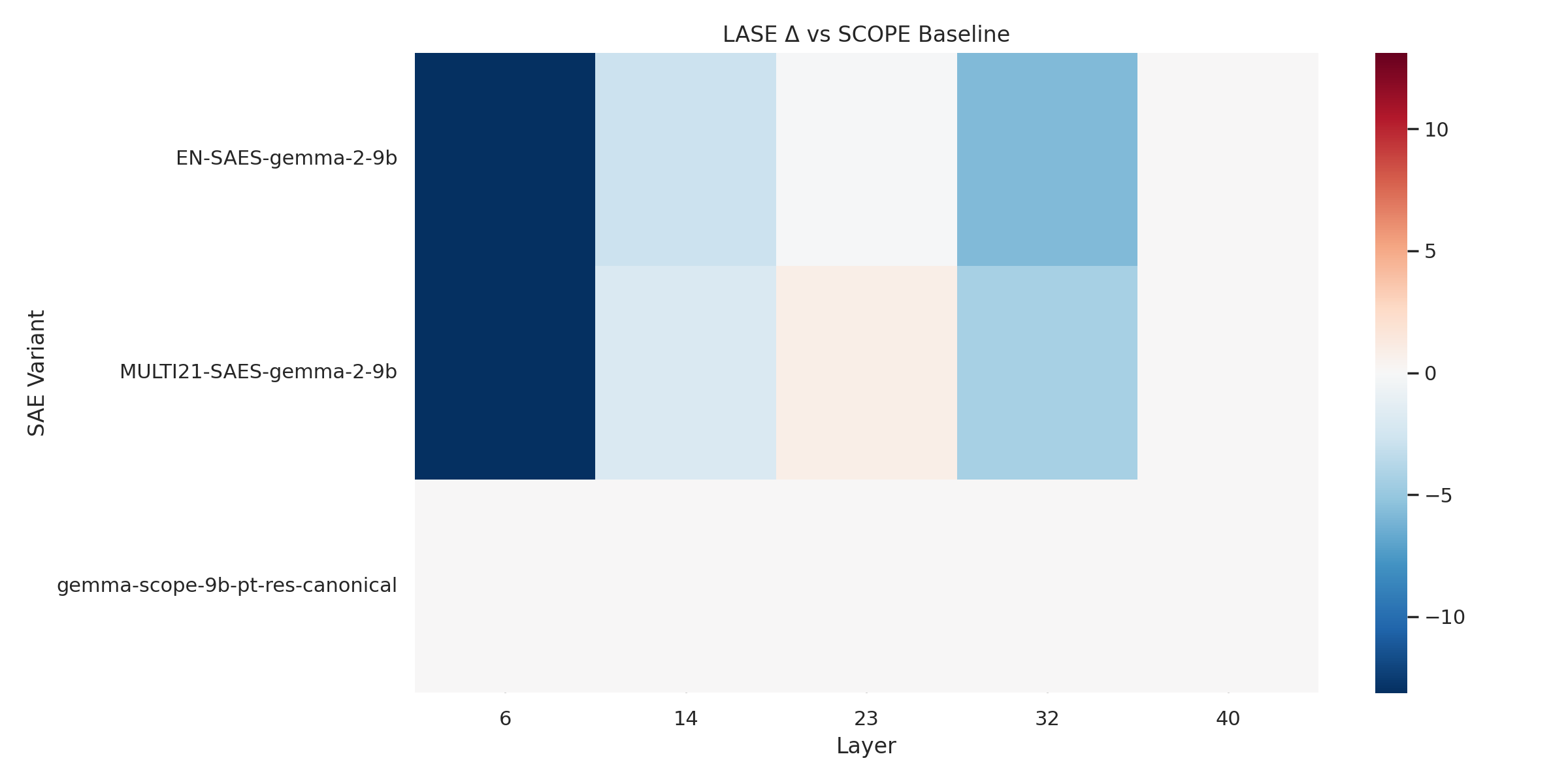}
    \includegraphics[width=0.9\linewidth]{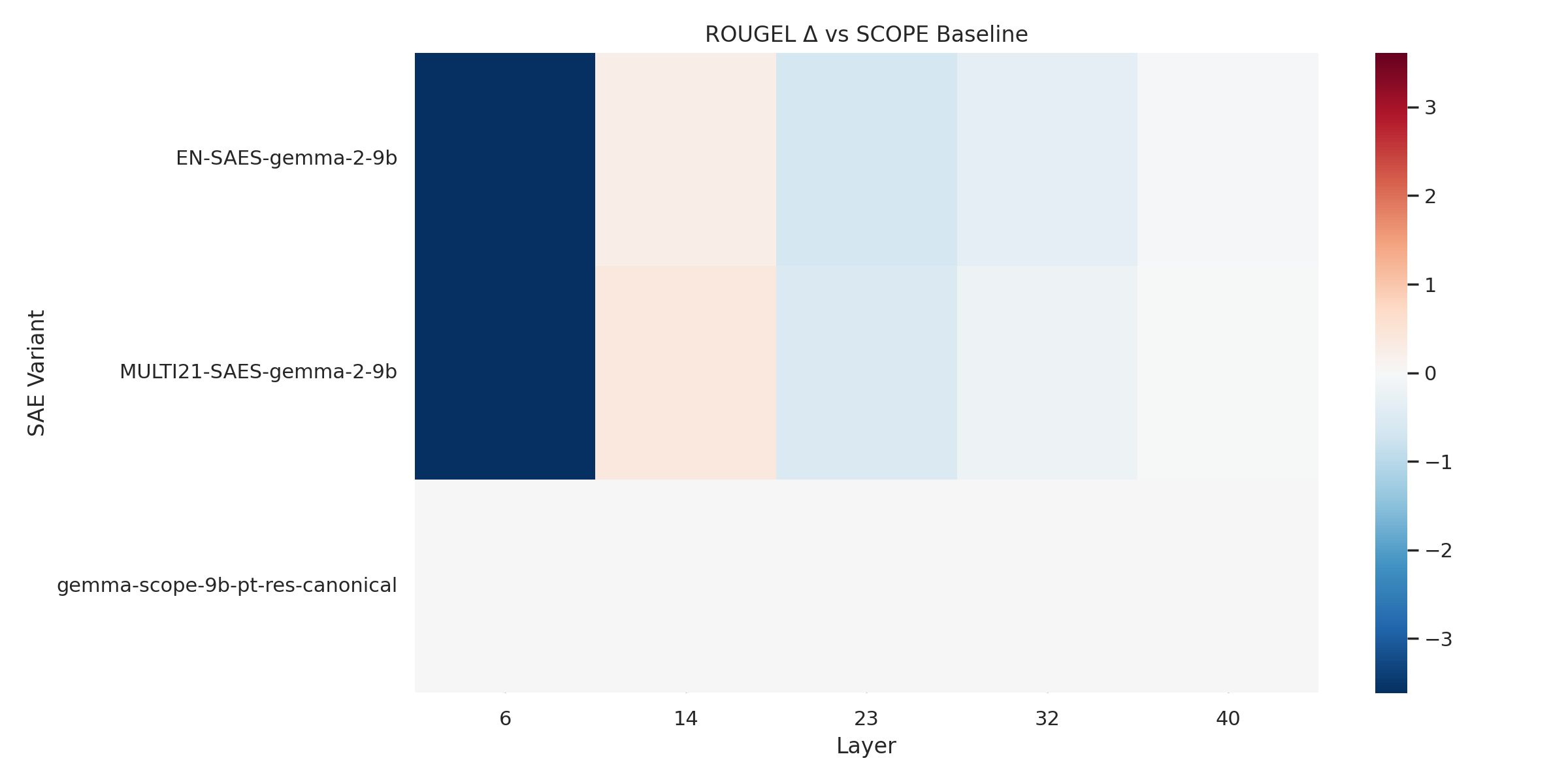}
    \caption{Layerwise heatmaps of performance deltas relative to \emph{Gemma-Scope} for \textbf{Gemma-2-9B} on \textbf{cross-lingual summarization (CrossSumm)}. Columns show deltas in \textbf{LangID}, \textbf{LaSE}, and \textbf{ROUGE-L} as a function of steering layer. Regions of positive gain cluster around the intersection layers identified by our multilinguality–separability criterion (\textbf{L14} and \textbf{L23}), indicating that these depths support more reliable language control and semantic preservation, though gains remain smaller than those achieved by our multilingual SAEs.}

    \label{fig:gemma_heatmap}
\end{figure*}

\begin{figure*}[t]
    \centering
    \includegraphics[width=0.9\linewidth]{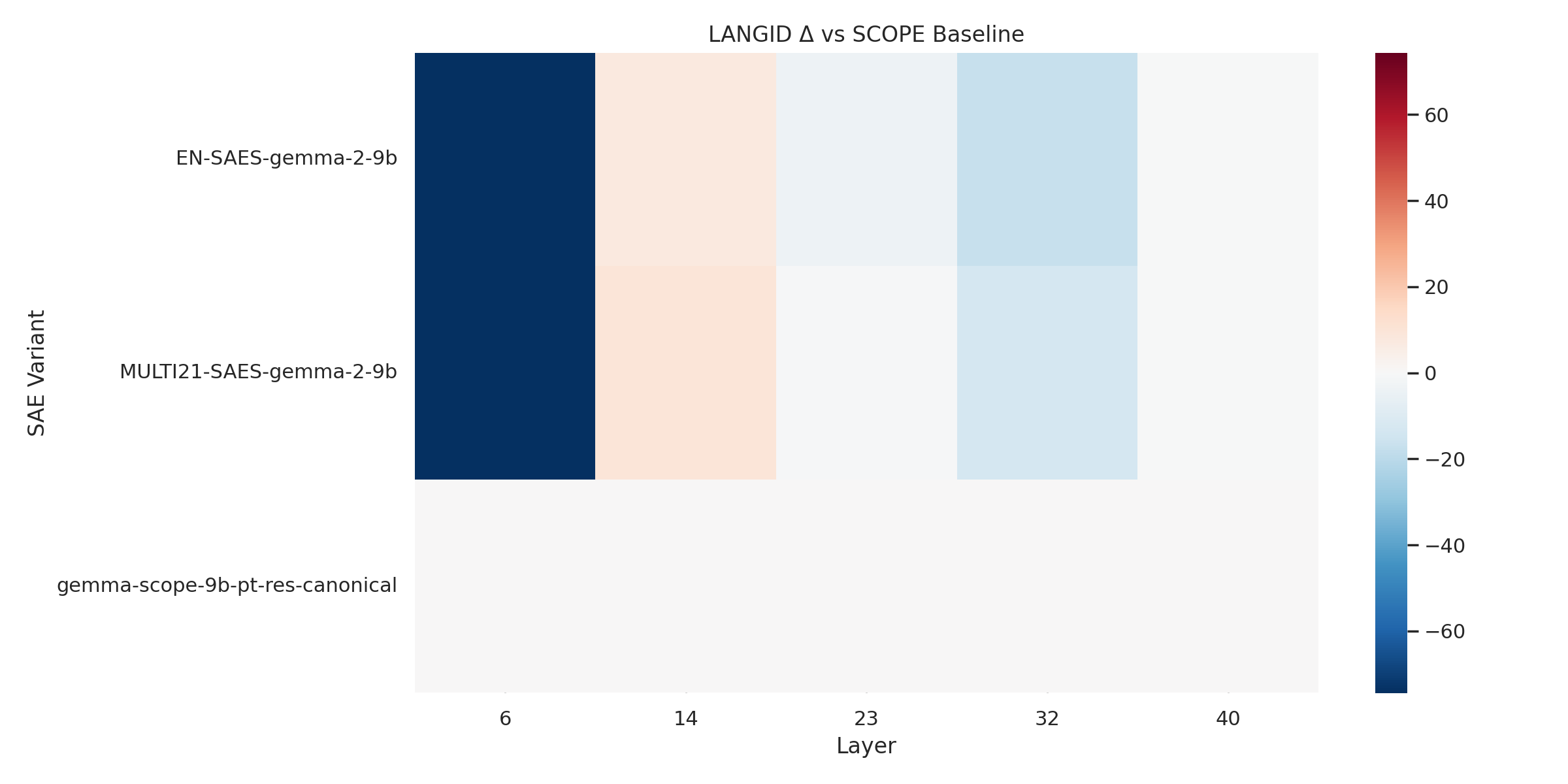}
    \includegraphics[width=0.9\linewidth]{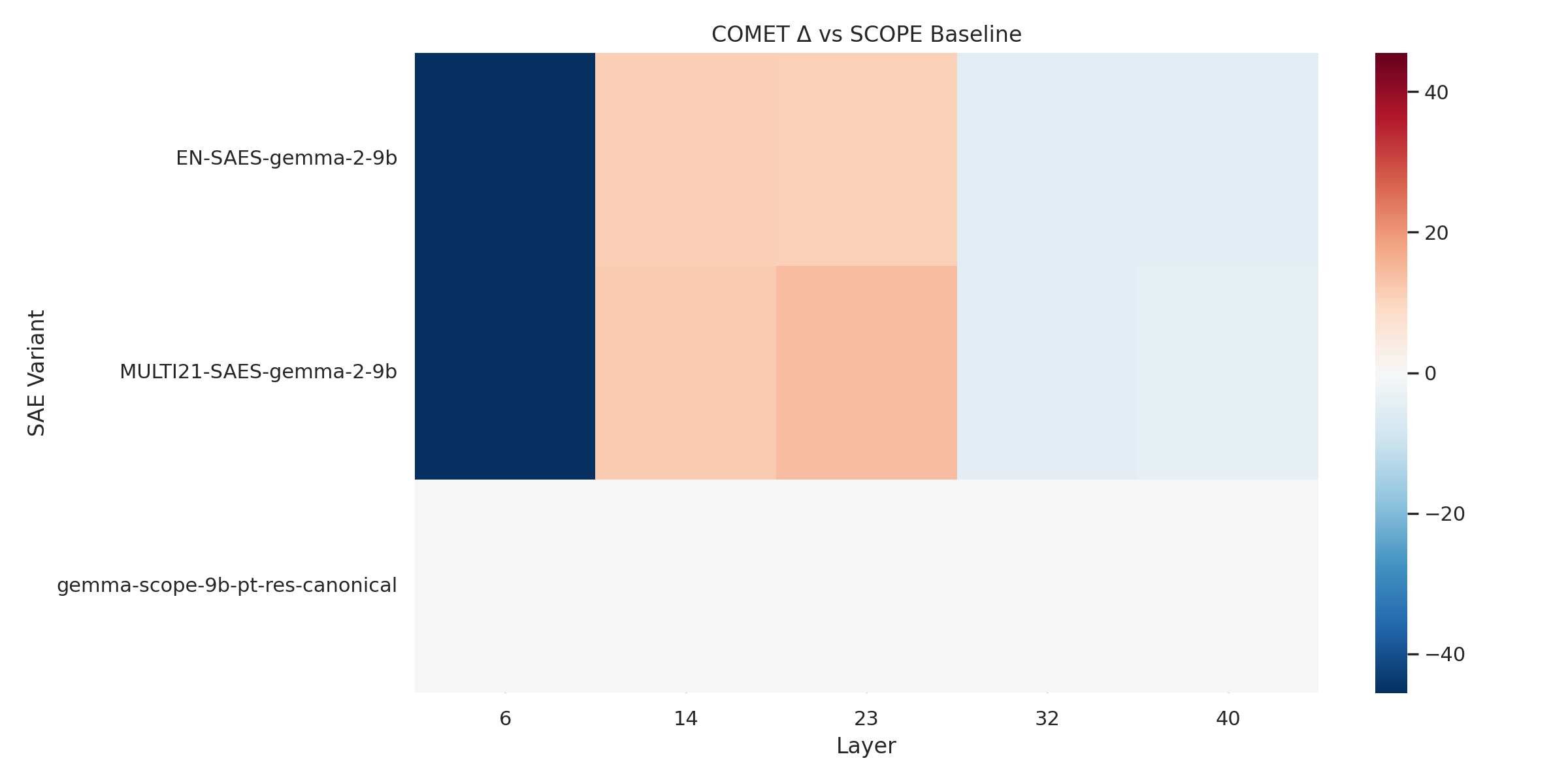}
    \includegraphics[width=0.9\linewidth]{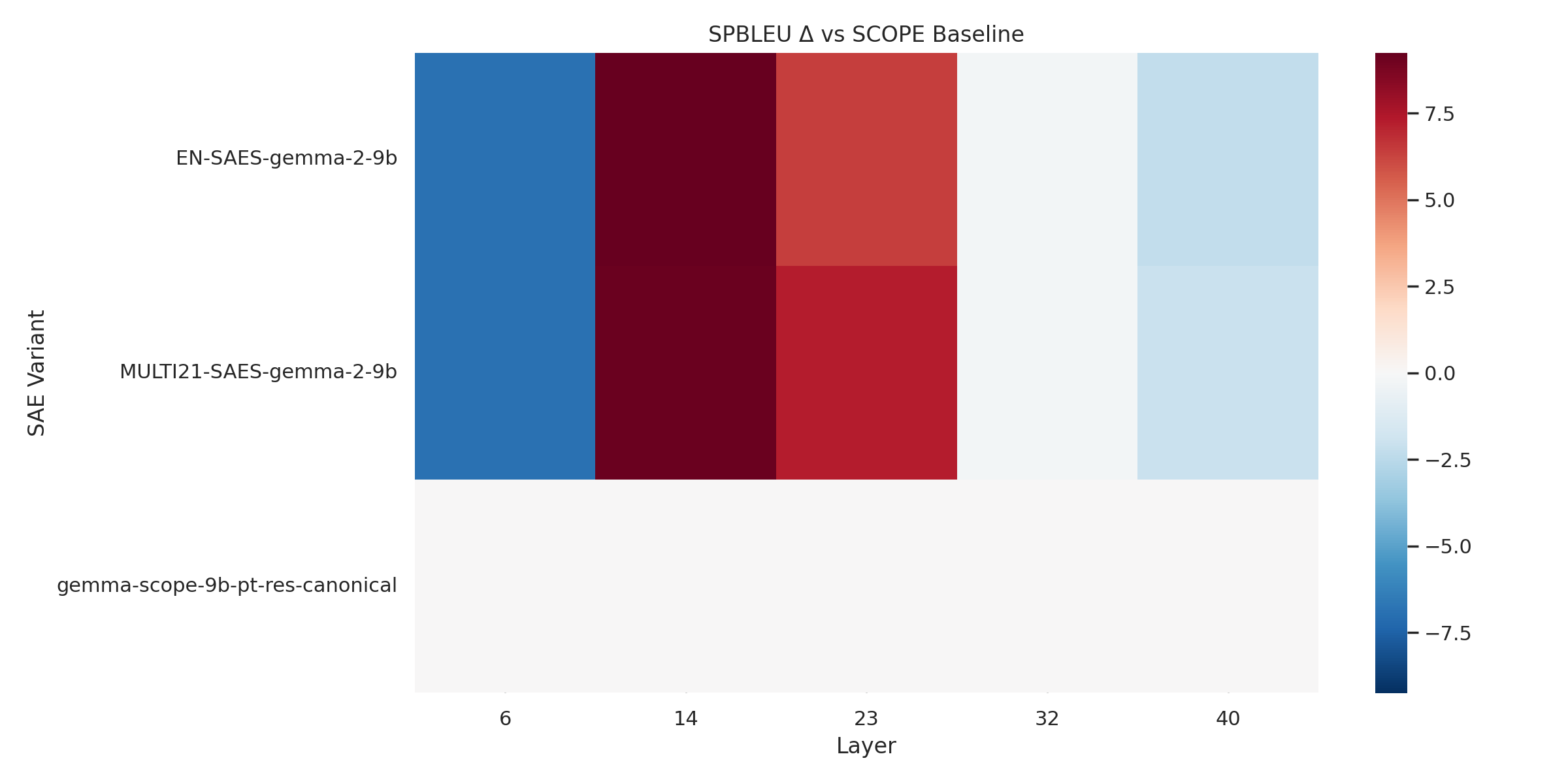}
    \caption{Layerwise heatmaps of performance deltas relative to \emph{Gemma-Scope} for \textbf{Gemma-2-9B} on \textbf{machine translation (FLORES)}. Columns report deltas in \textbf{LangID}, \textbf{COMET}, and \textbf{SpBLEU} across steering layers. Improved performance concentrates near the predicted intersection layers (\textbf{L14} and \textbf{L23}), validating that these depths balance cross-lingual alignment and language separability, but still underperform compared to multilingual SAEs trained in our framework.}

    \label{fig:gemma_heatmap}
\end{figure*}

\clearpage

\begin{figure*}[t]
\section{Multilingual Results for LLaMA-3.1-8B}

    \centering

    \includegraphics[width=0.3\linewidth]{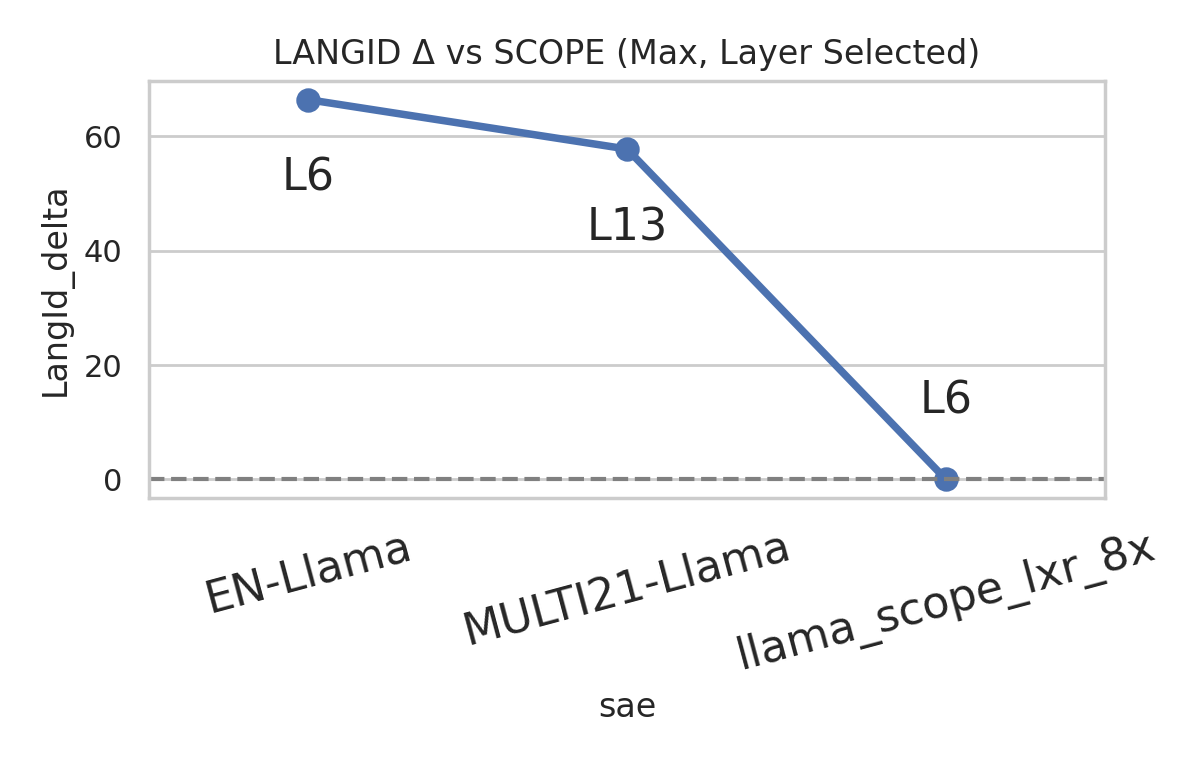}
    \includegraphics[width=0.3\linewidth]{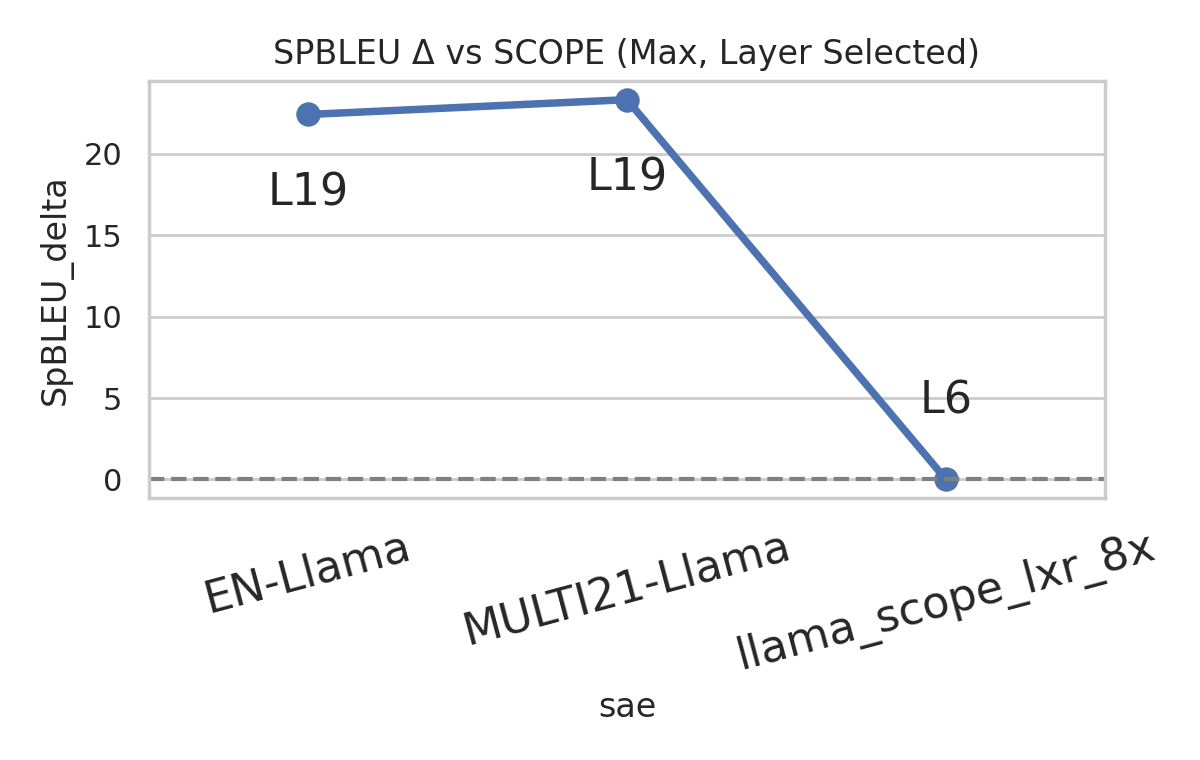}
    \includegraphics[width=0.3\linewidth]{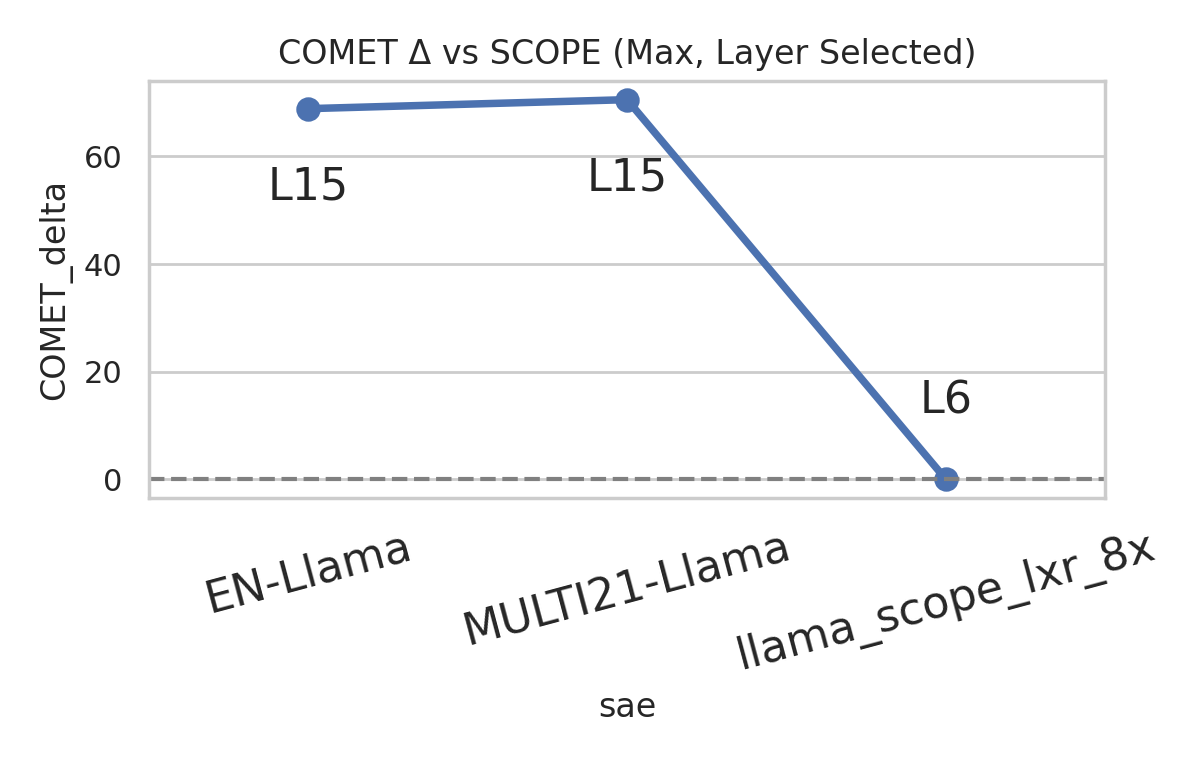}  \\
    
    \includegraphics[width=0.3\linewidth]{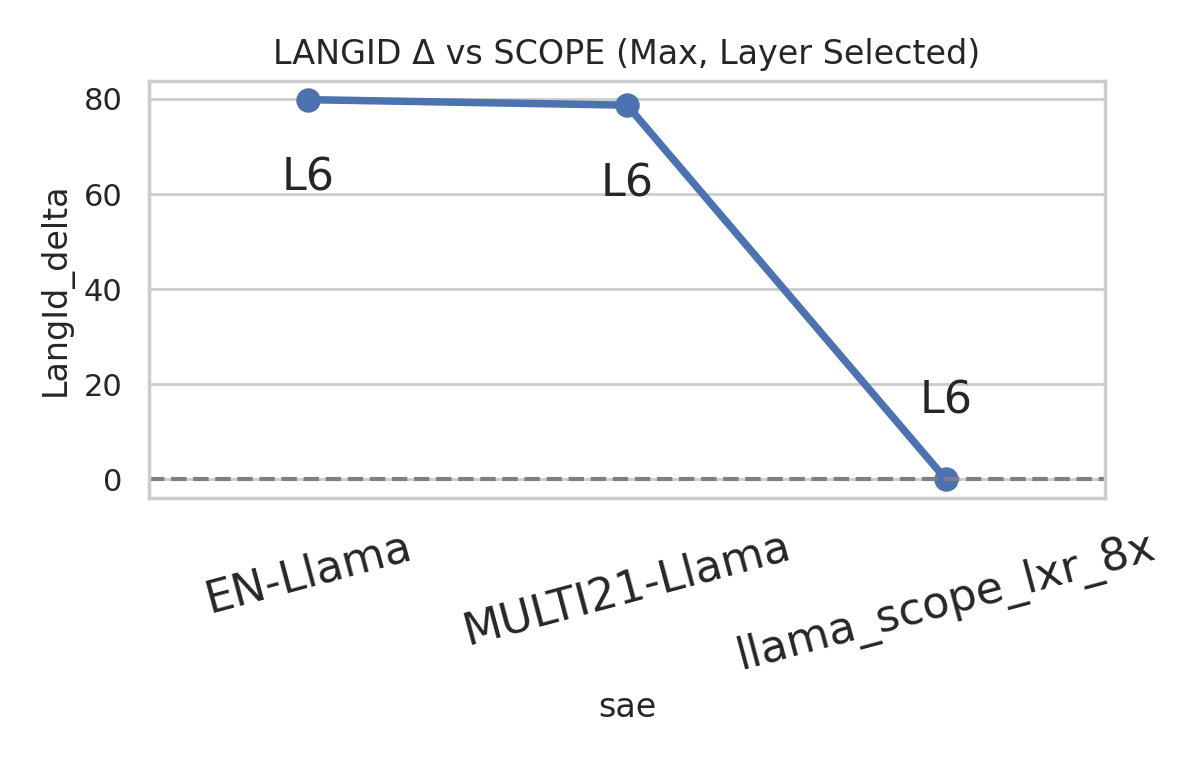}
    \includegraphics[width=0.3\linewidth]{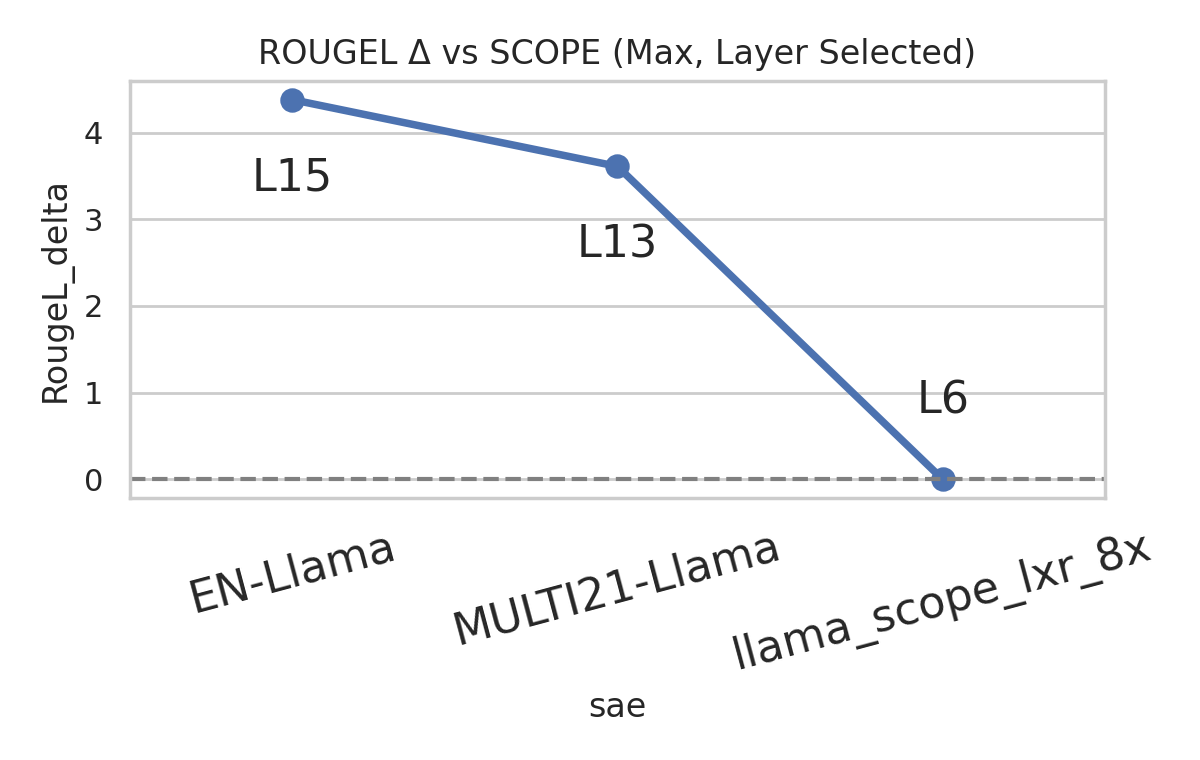}
    \includegraphics[width=0.3\linewidth]{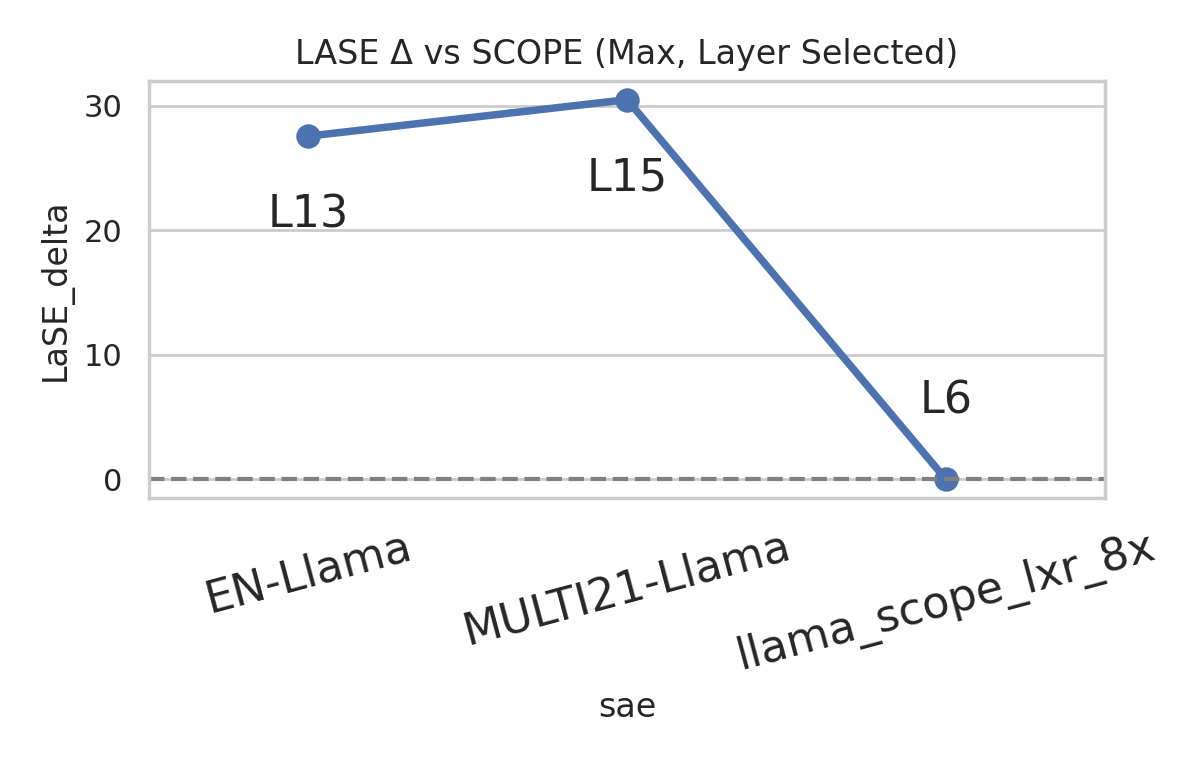}

    \caption{Performance deltas relative to Scope baselines for \textbf{LLaMA-3.1-8B} at the best-performing steering layer. \textbf{Top:} FLORES machine translation (LangID, SpBLEU, COMET). \textbf{Bottom:} Cross-lingual summarization (LangID, ROUGE-L, LaSE). Multilingual SAEs consistently outperform English-only SAEs across both tasks, with larger gains on semantic quality metrics.}
    \label{fig:llama_deltas}
\end{figure*}

\begin{figure*}[t]

    \centering

    \includegraphics[width=0.3\linewidth]{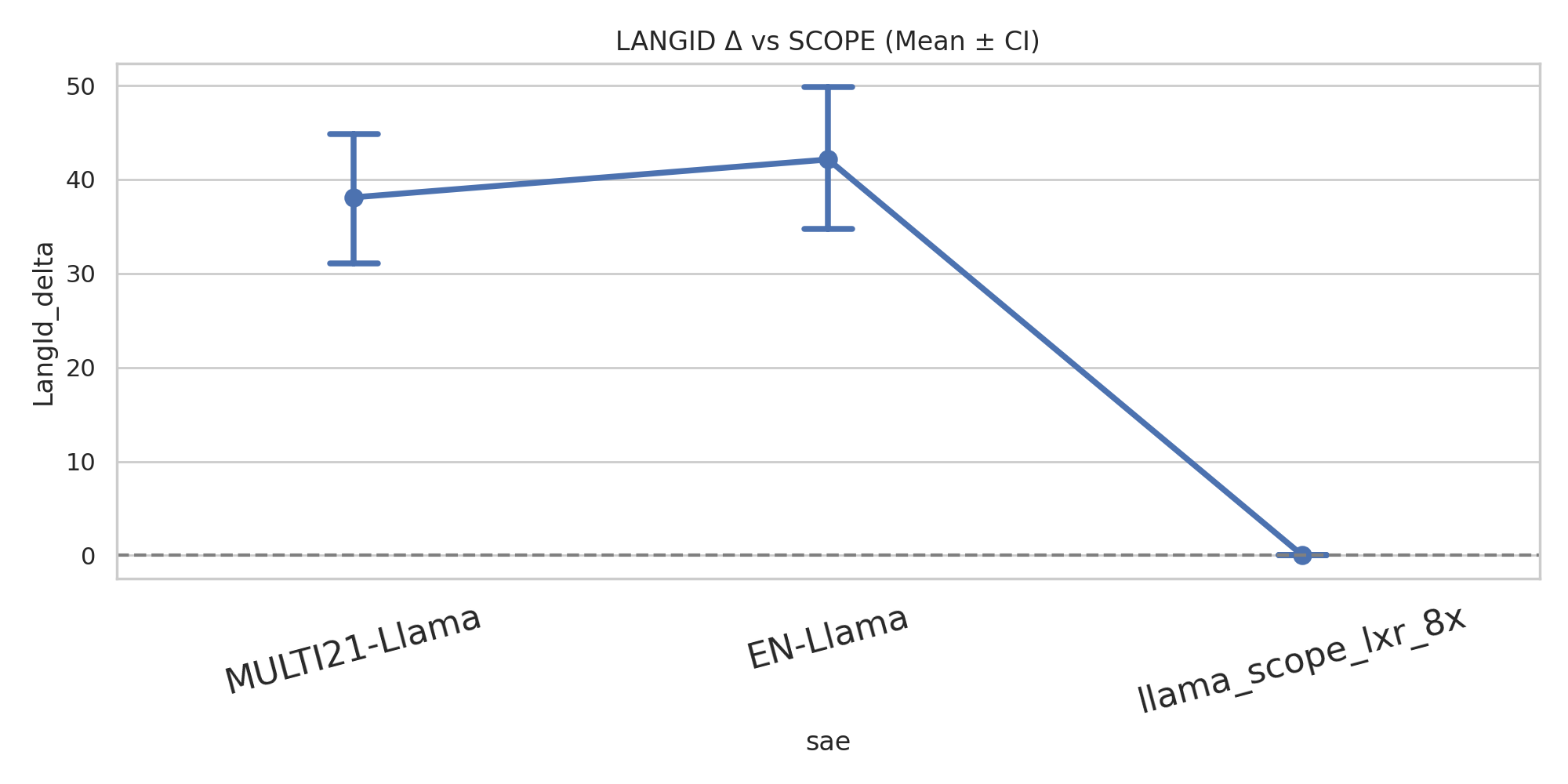}
    \includegraphics[width=0.3\linewidth]{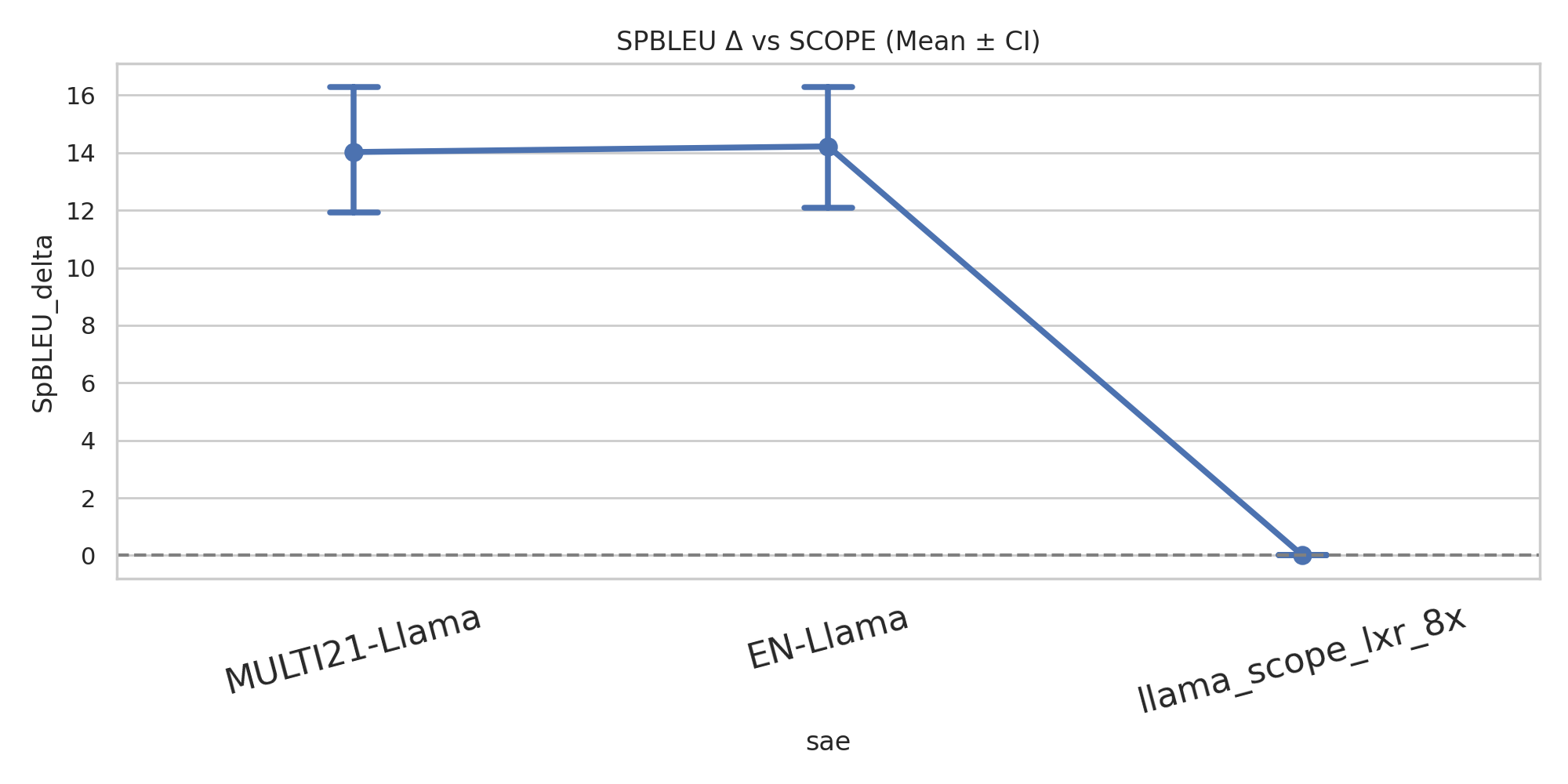}
    \includegraphics[width=0.3\linewidth]{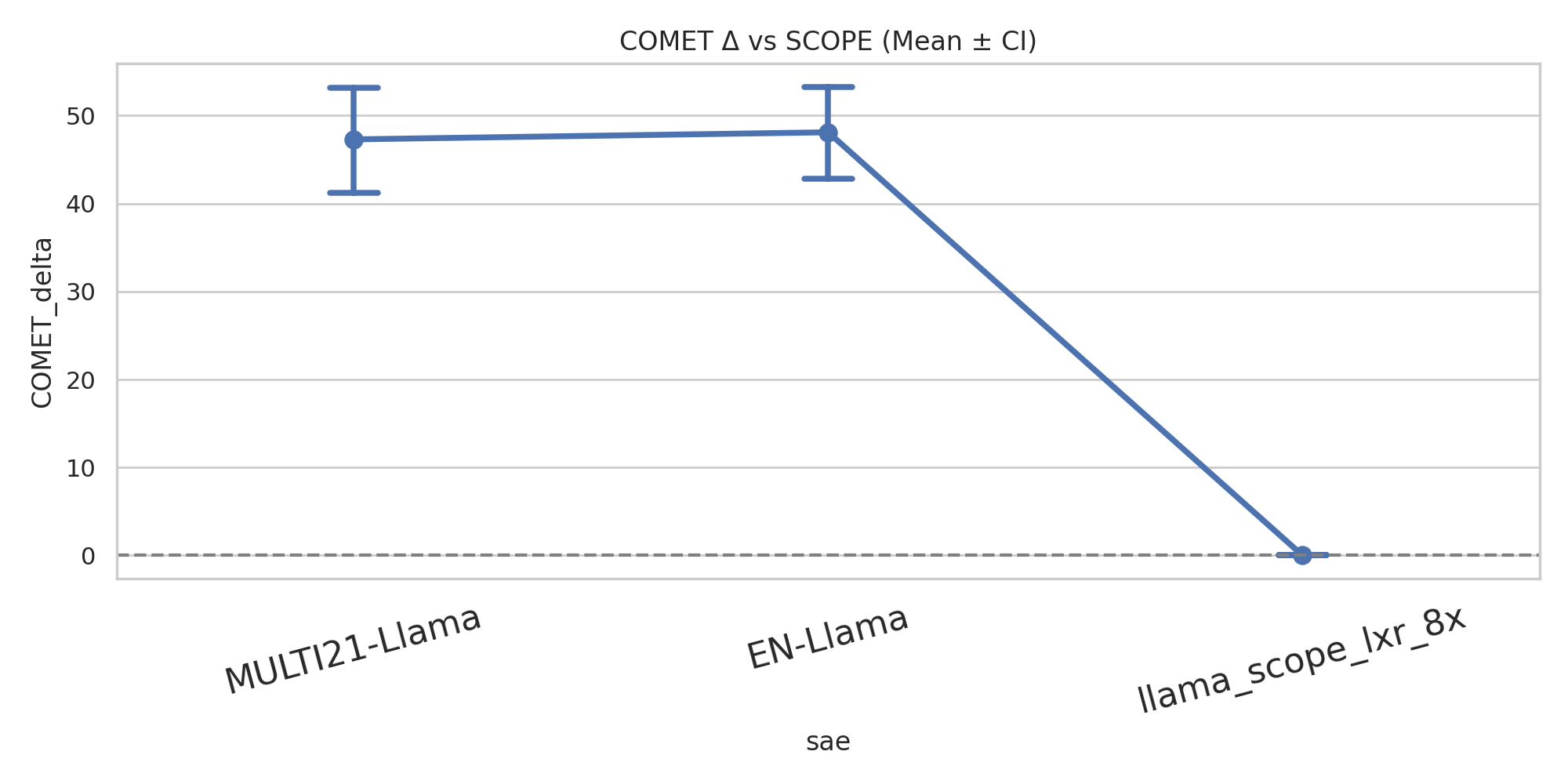}  \\
    
    \includegraphics[width=0.3\linewidth]{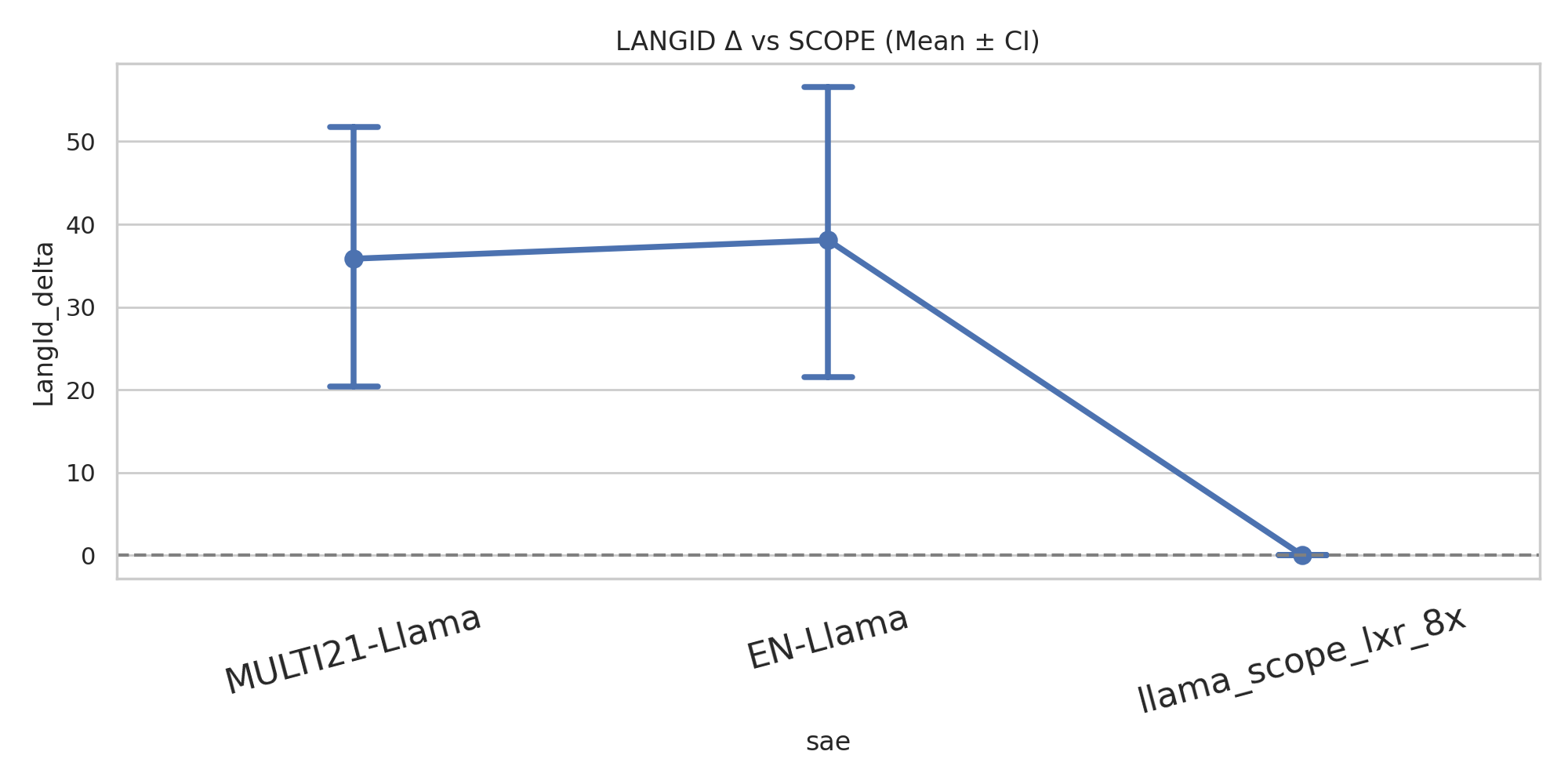}
    \includegraphics[width=0.3\linewidth]{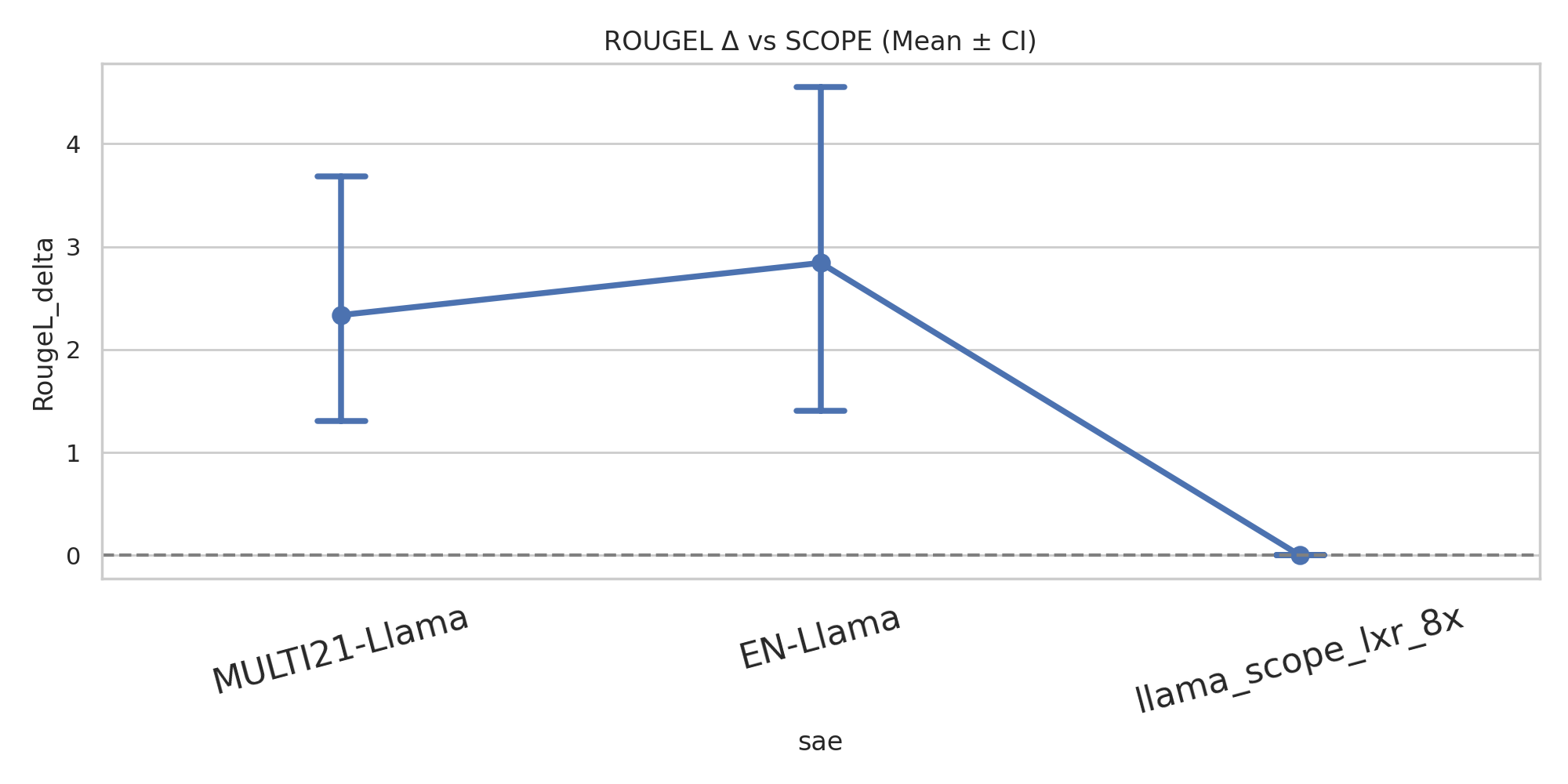}
    \includegraphics[width=0.3\linewidth]{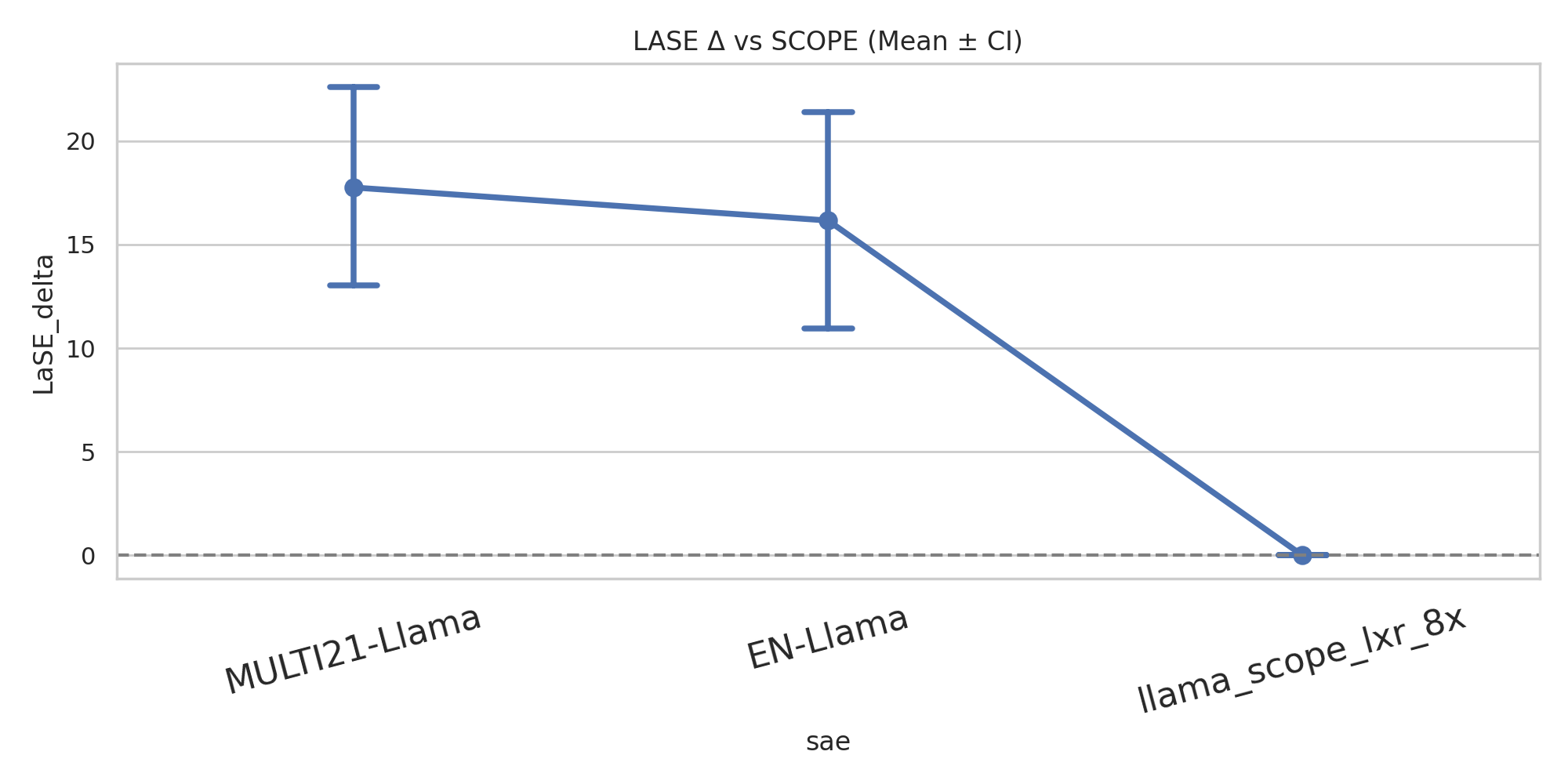}

    \caption{Performance deltas relative to Scope baselines for \textbf{LLaMA-3.1-8B} averaged across layers. \textbf{Top:} FLORES machine translation (LangID, SpBLEU, COMET). \textbf{Bottom:} Cross-lingual summarization (LangID, ROUGE-L, LaSE).}
    \label{fig:llama_deltas_avg}
\end{figure*}

\begin{figure*}[t]
    \centering
    \includegraphics[width=0.9\linewidth]{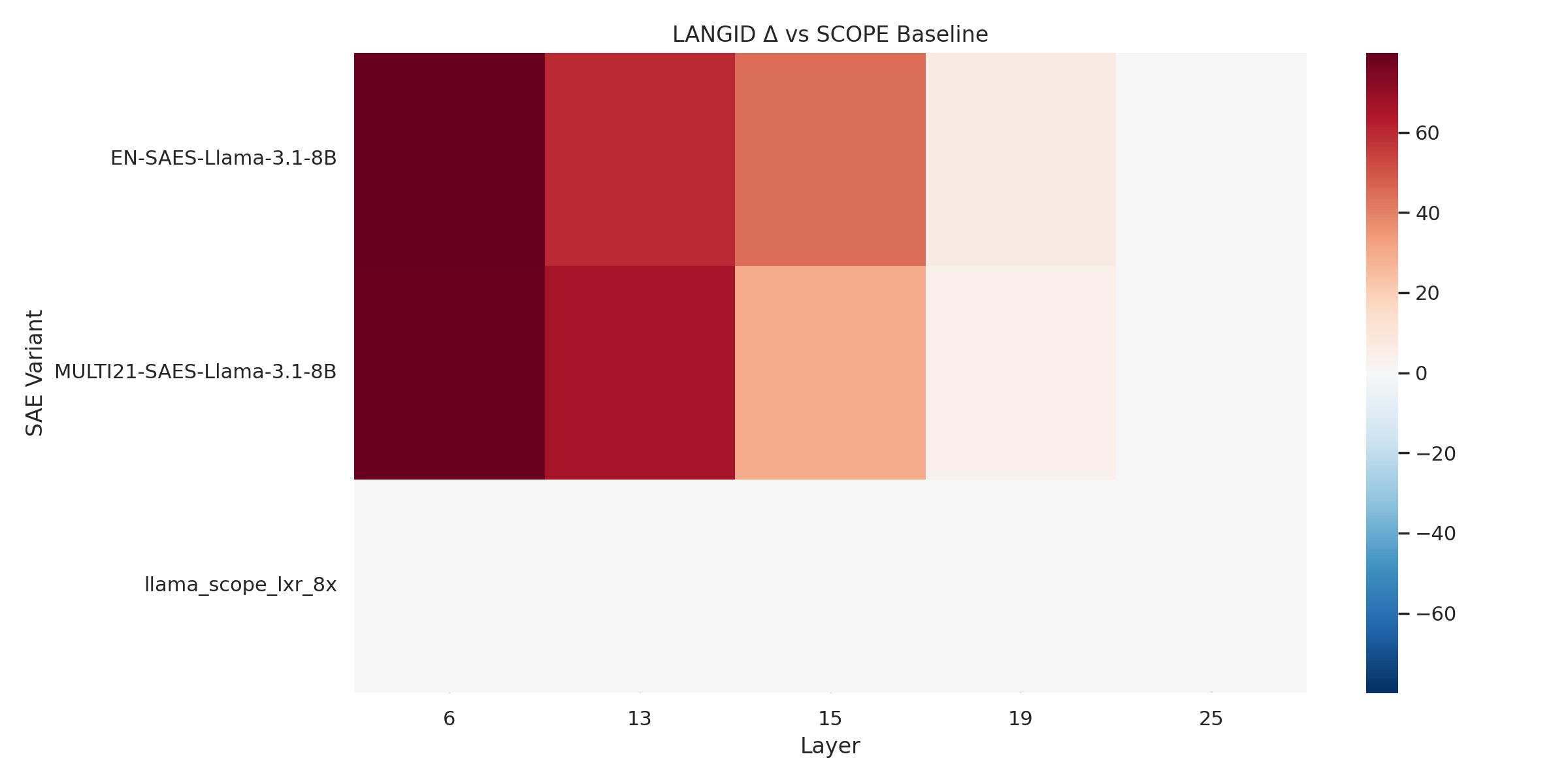}
    \includegraphics[width=0.9\linewidth]{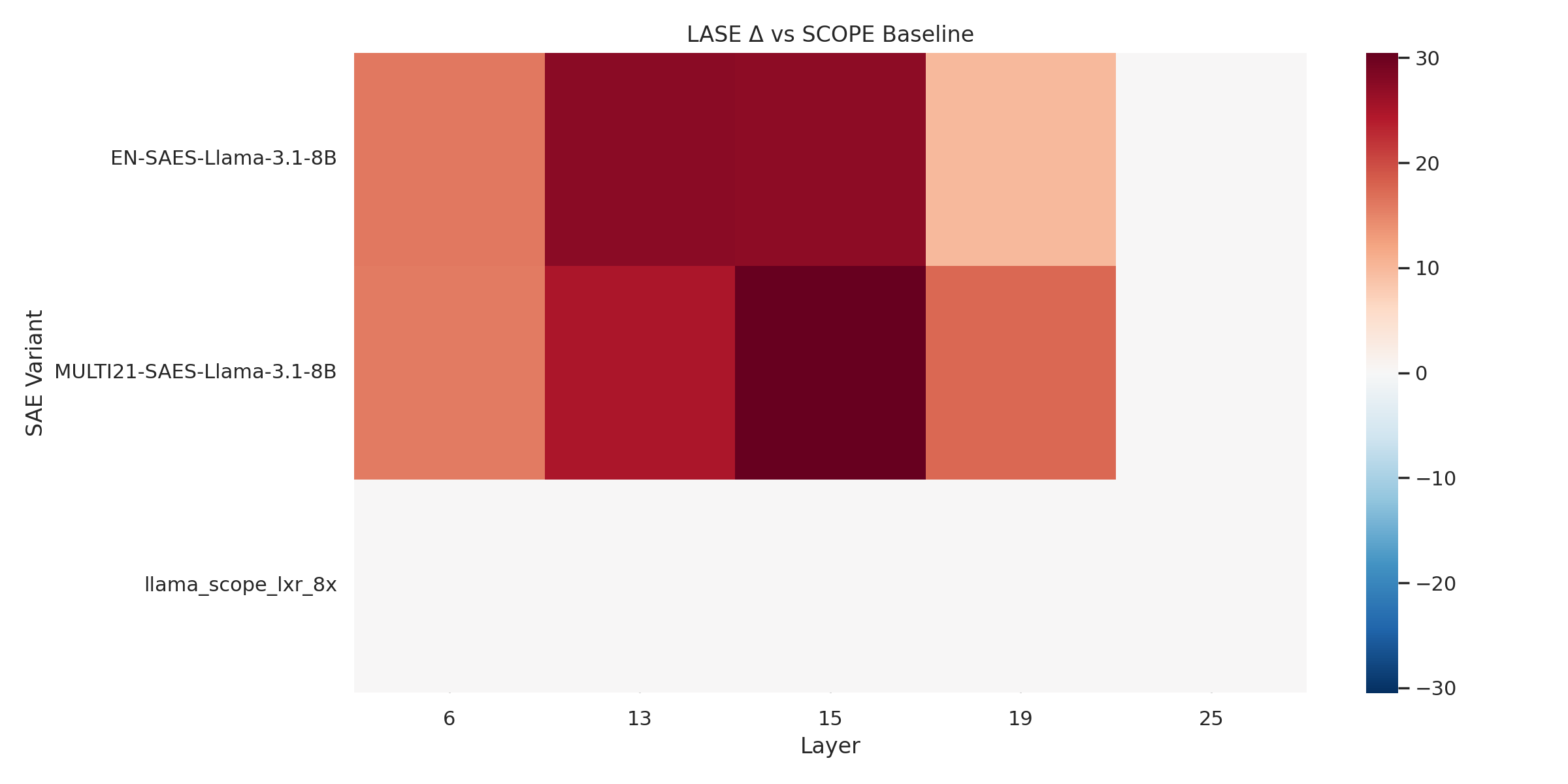}
    \includegraphics[width=0.9\linewidth]{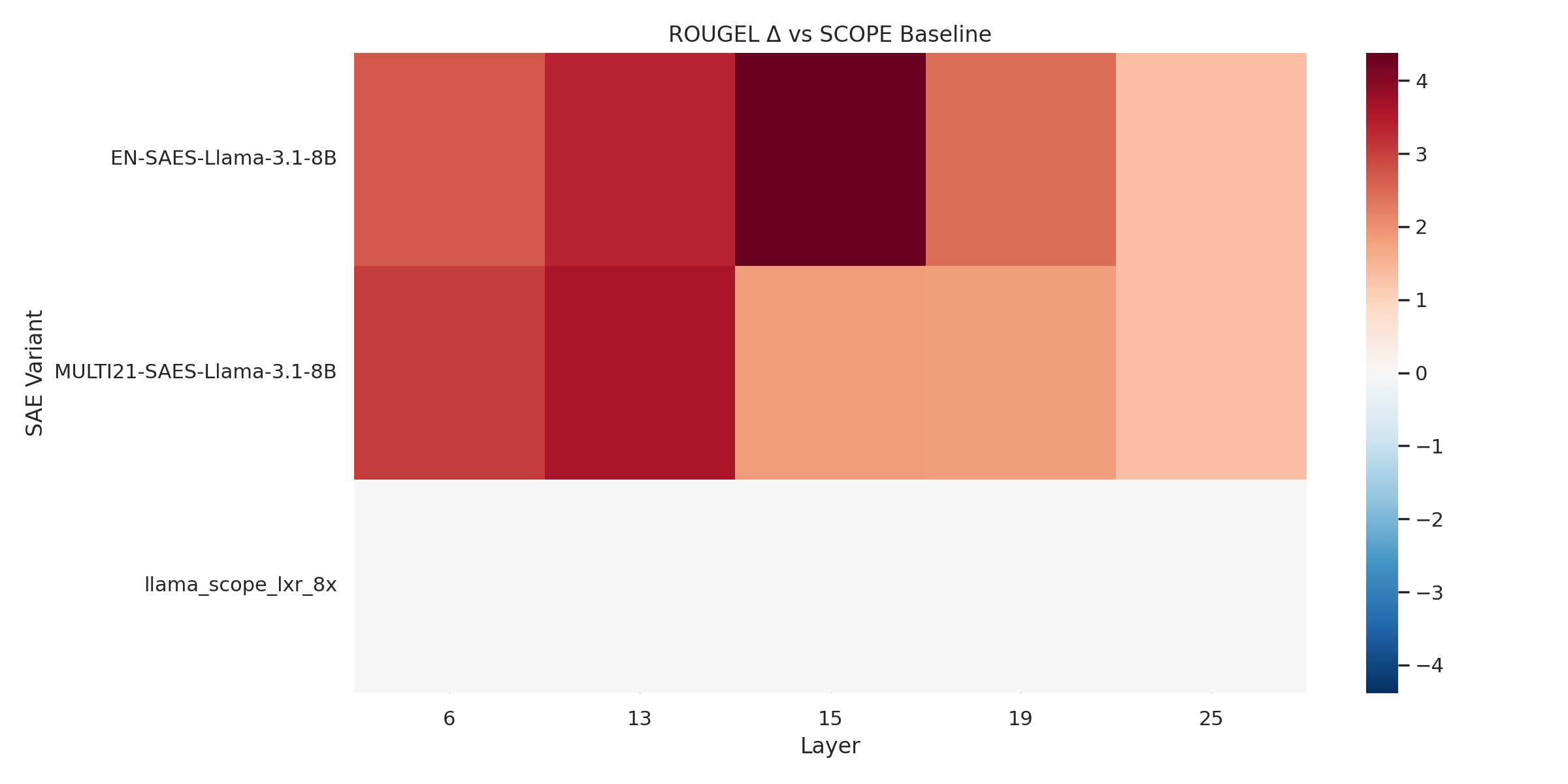}
    \caption{Layerwise heatmaps of performance deltas relative to \emph{LLaMA-Scope} for \textbf{LLaMA-3.1-8B} on \textbf{cross-lingual summarization (CrossSumm)}. Columns show deltas in \textbf{LangID}, \textbf{LaSE}, and \textbf{ROUGE-L} as a function of steering layer. Unlike Gemma-Scope, LLaMA-Scope exhibits weak and diffuse gains across layers, with no clear concentration around an intersection depth, consistent with the absence of a strong multilinguality–separability balance and its limited downstream steering effectiveness.}

    \label{fig:gemma_heatmap}
\end{figure*}

\begin{figure*}[t]
    \centering
    \includegraphics[width=0.9\linewidth]{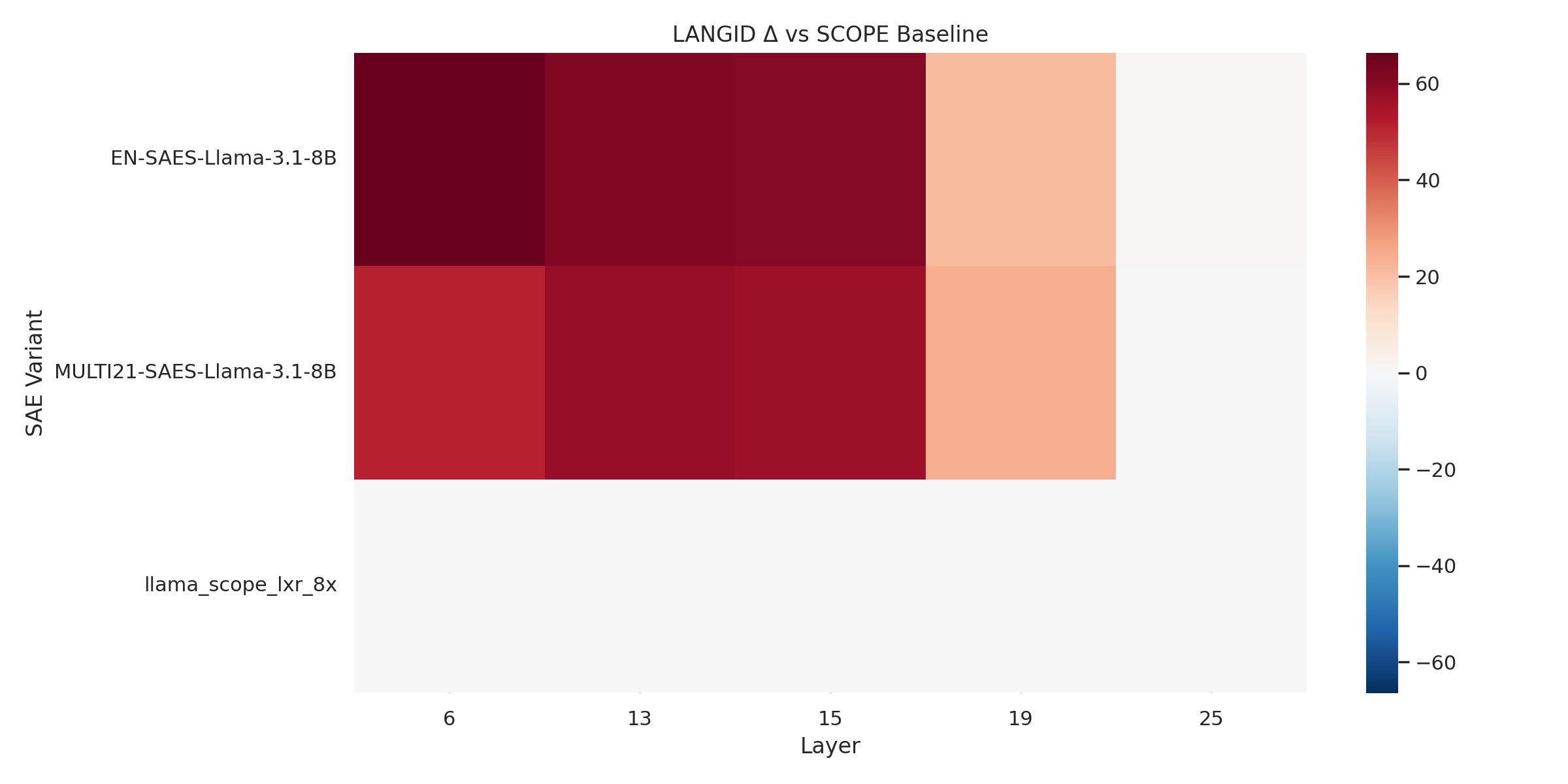}
    \includegraphics[width=0.9\linewidth]{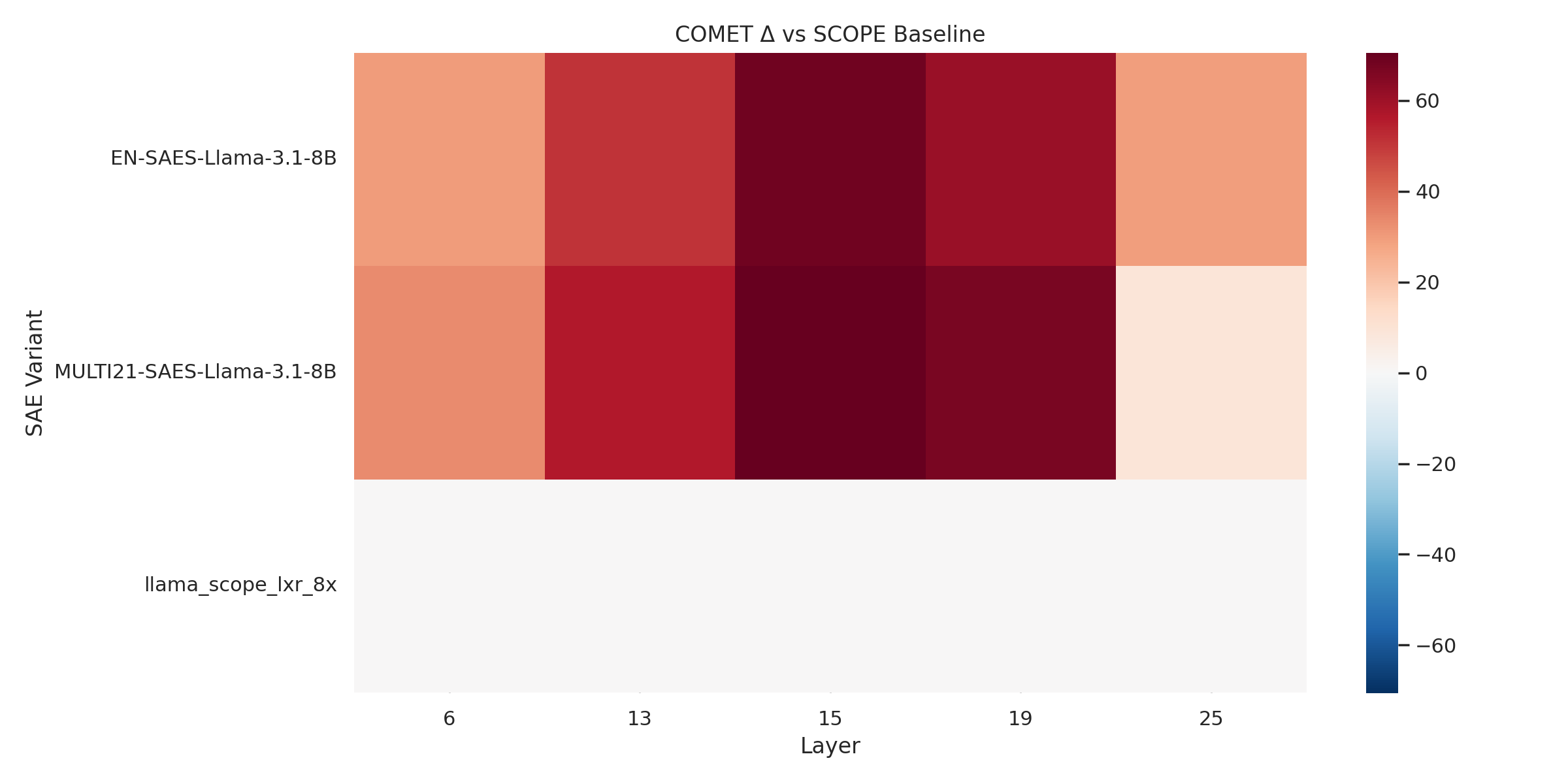}
    \includegraphics[width=0.9\linewidth]{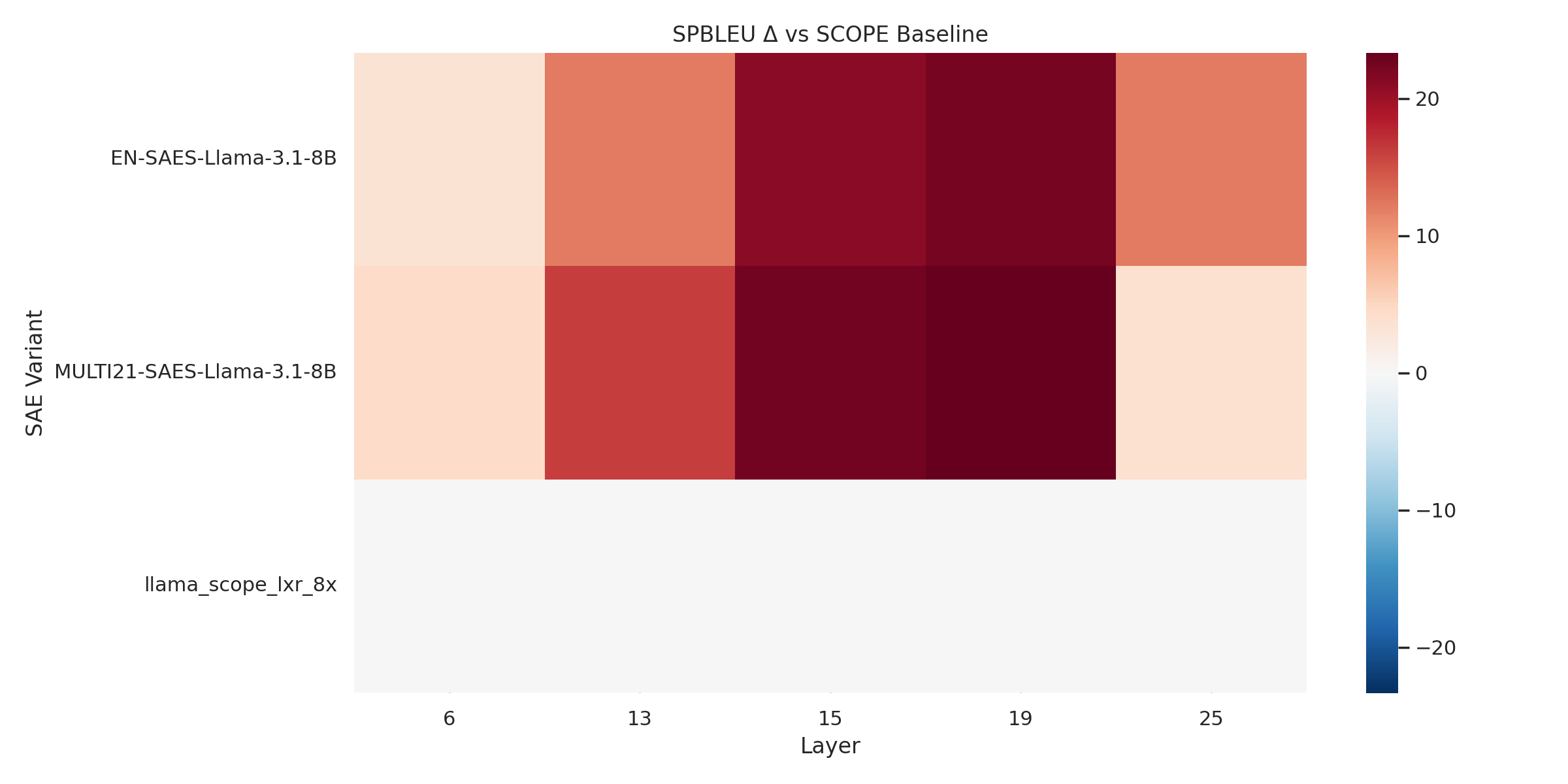}
    \caption{Layerwise heatmaps of performance deltas relative to \emph{LLaMA-Scope} for \textbf{LLaMA-3.1-8B} on \textbf{machine translation (FLORES)}. Columns report deltas in \textbf{LangID}, \textbf{COMET}, and \textbf{SpBLEU} across steering layers. Performance improvements remain small and scattered across depth, with no distinct layer emerging as consistently effective, mirroring the lack of a clear intersection between multilingual alignment and language separability in LLaMA-Scope.}

    \label{fig:gemma_heatmap}
\end{figure*}

\clearpage

\newpage





\begin{figure*}[htbp]
\section{Language vectors correlations and sparsity score}

\centering

\begin{tabularx}{0.7\textwidth}{@{}p{2.2cm}YYYY@{}}
& \multicolumn{1}{c}{\textbf{Residual Stream}} & \multicolumn{1}{c}{\textbf{Llama-Scope}} & \multicolumn{1}{c}{\textbf{Our-en-SAEs}} & \multicolumn{1}{c}{\textbf{Our-multi-SAEs}} \\[0.5em]

\textbf{Layer 1} &
\includegraphics[width=\linewidth]{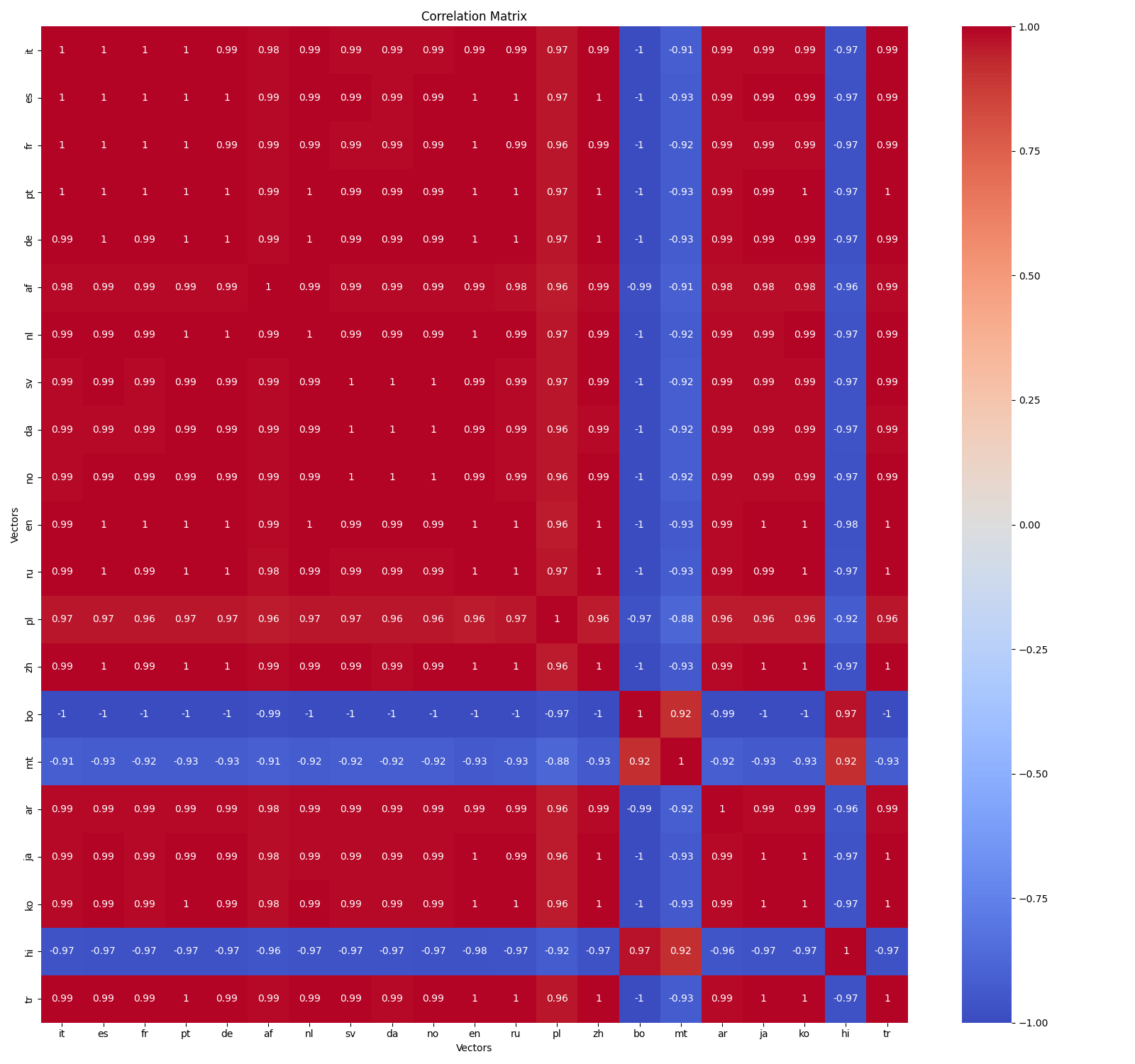} &
\includegraphics[width=\linewidth]{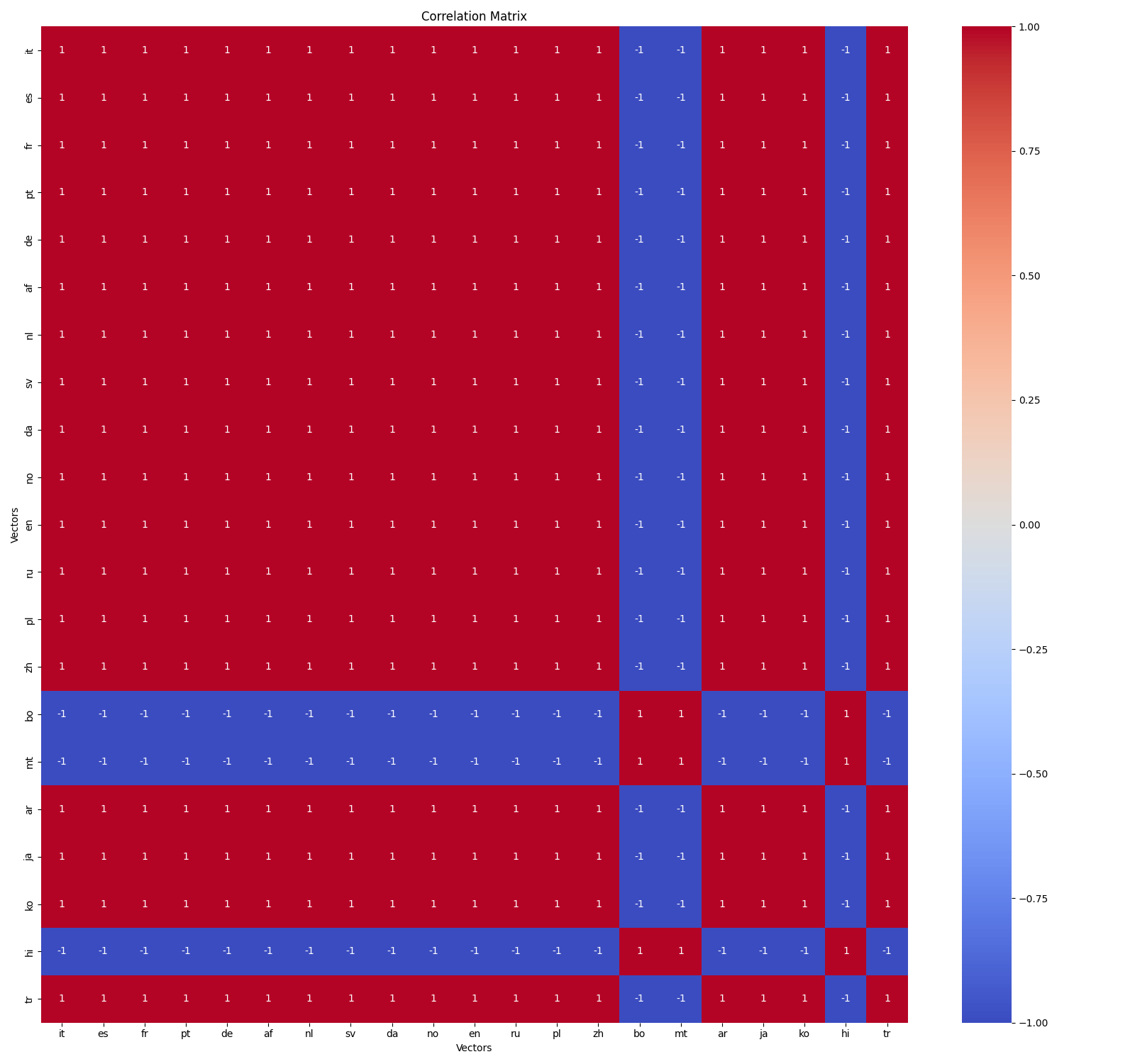} &
\includegraphics[width=\linewidth]{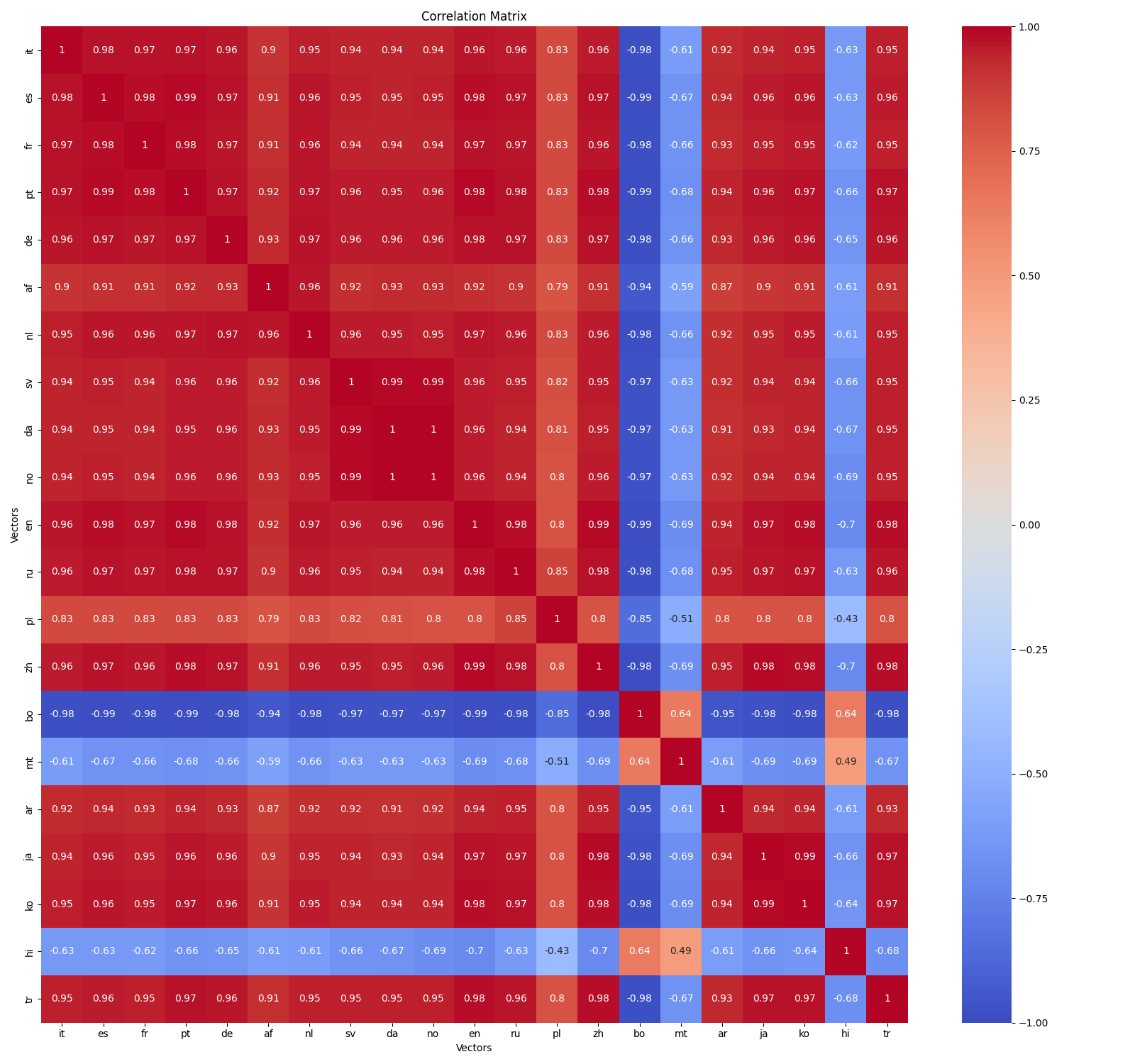} &
\includegraphics[width=\linewidth]{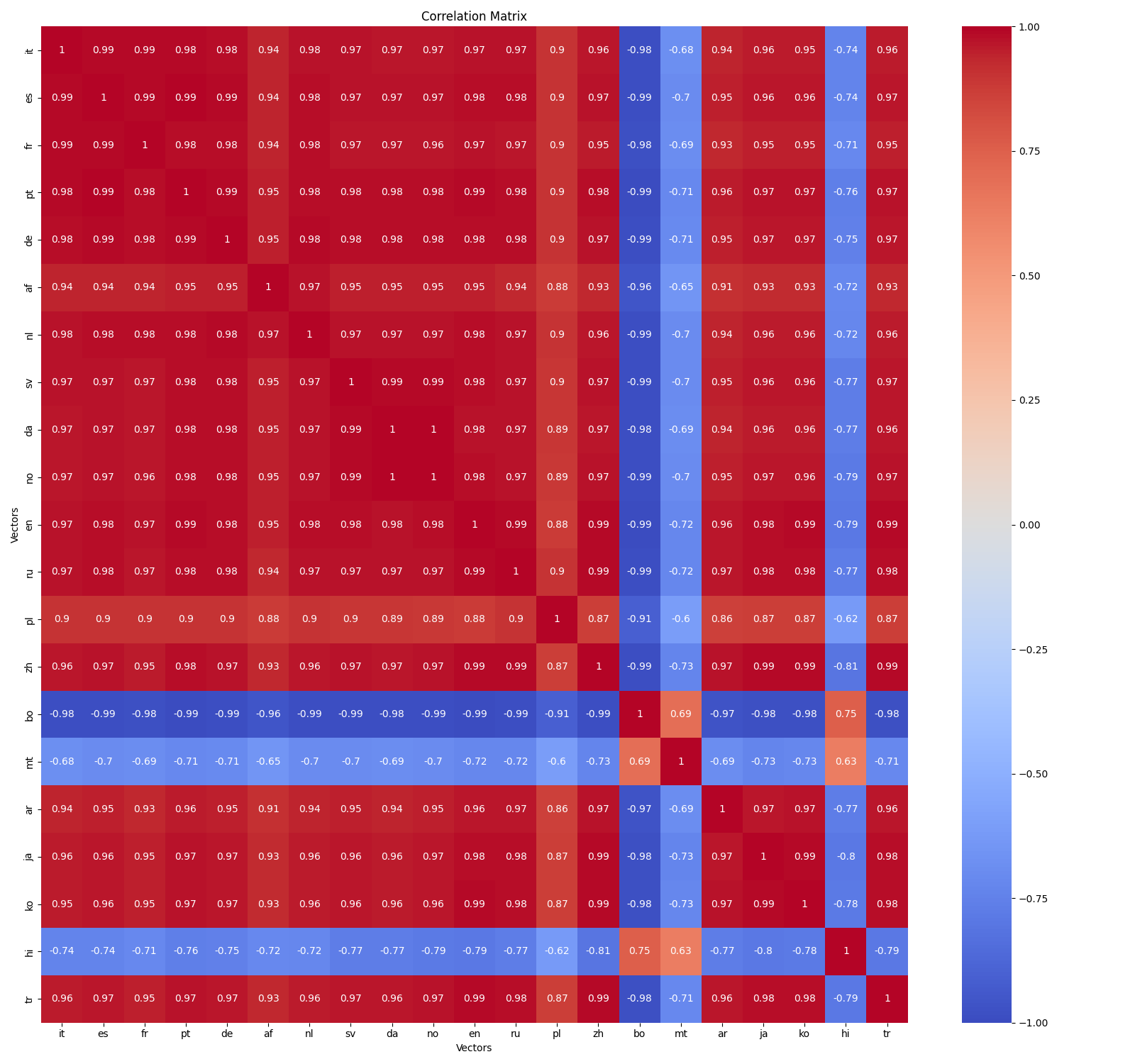} \\[0.1em]

Sep. score&0.01&0.00&0.00&0.07 \\[0.6em]

\textbf{Layer 11} &
\includegraphics[width=\linewidth]{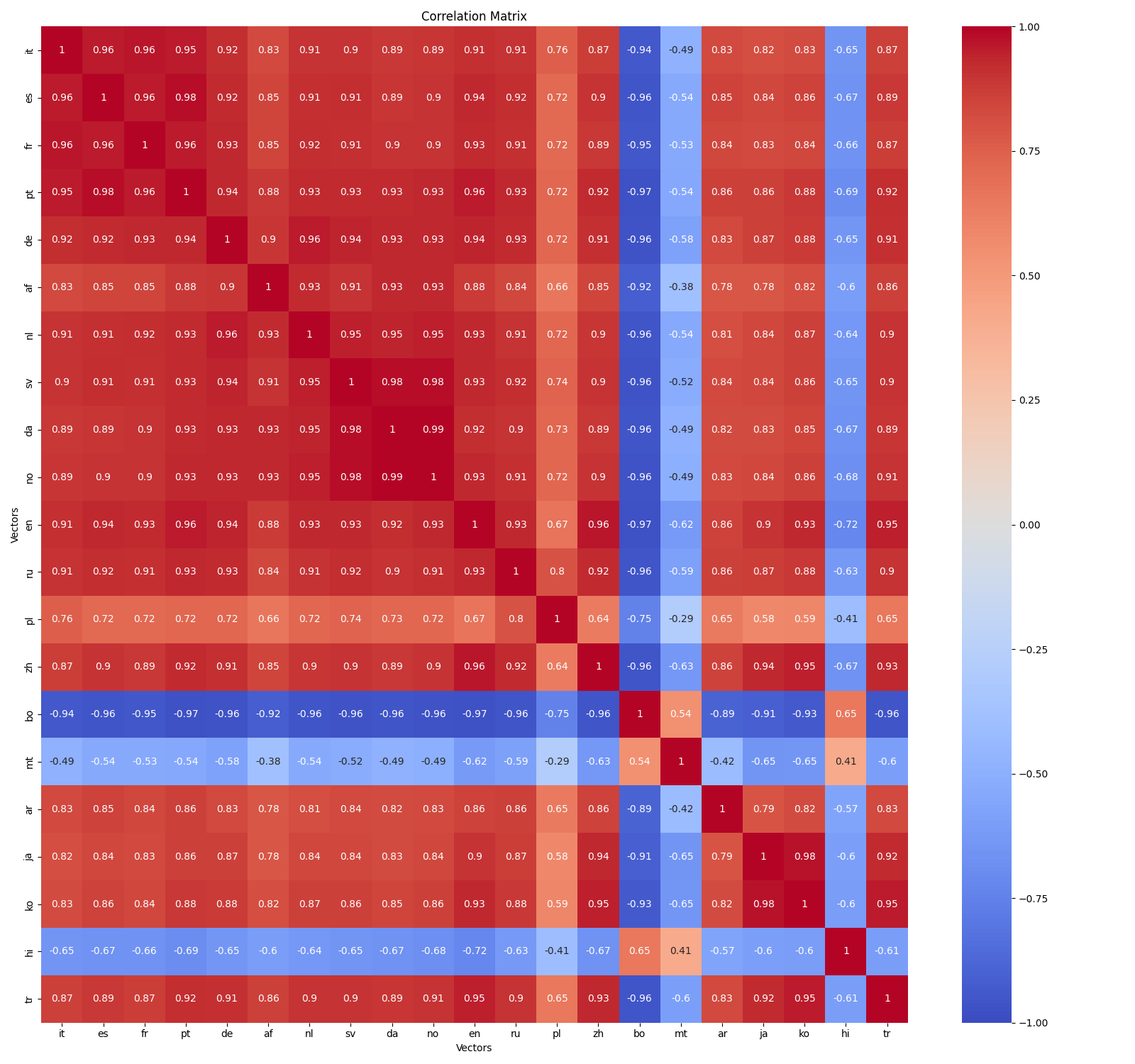} &
\includegraphics[width=\linewidth]{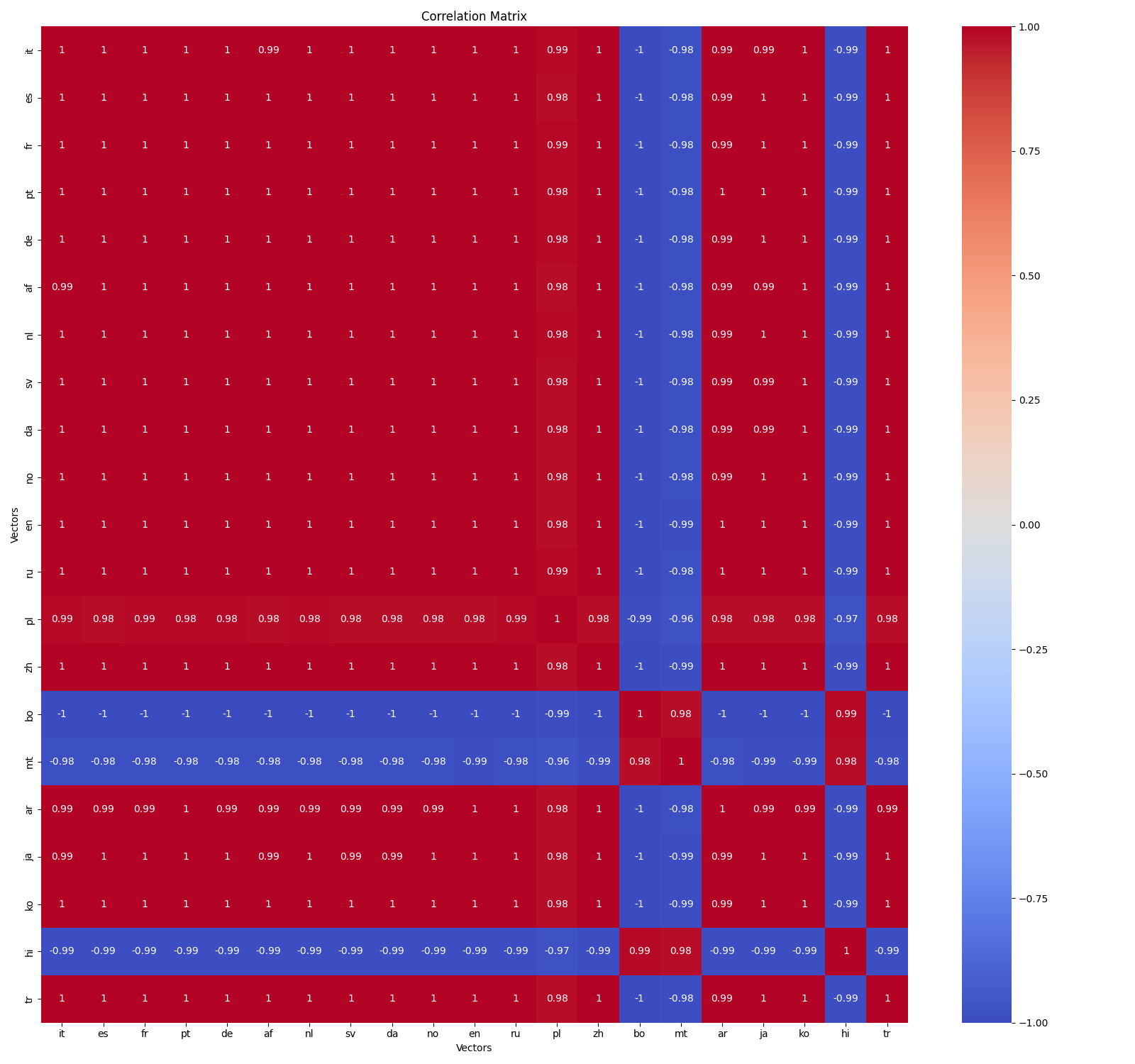} &
\includegraphics[width=\linewidth]{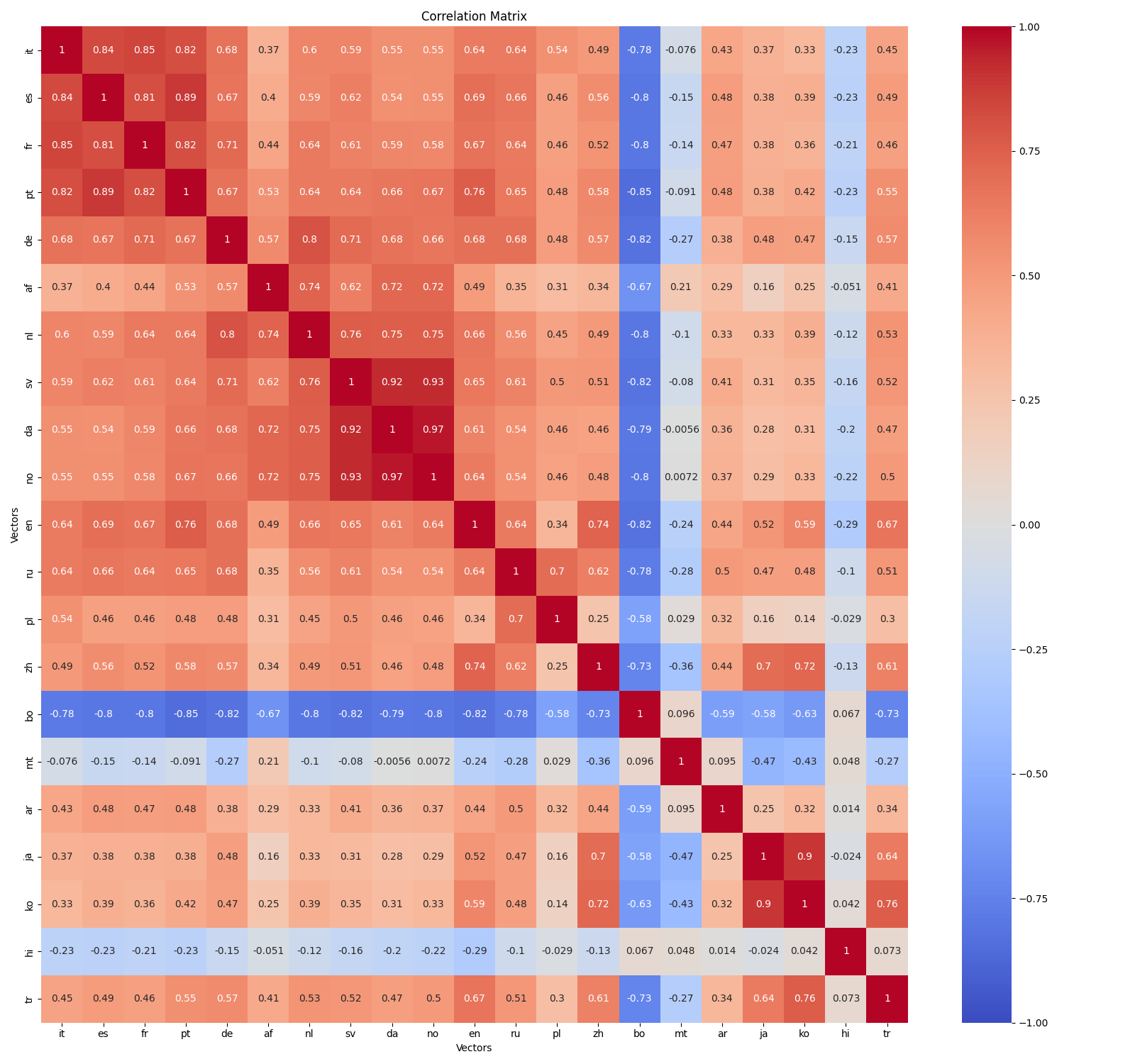} &
\includegraphics[width=\linewidth]{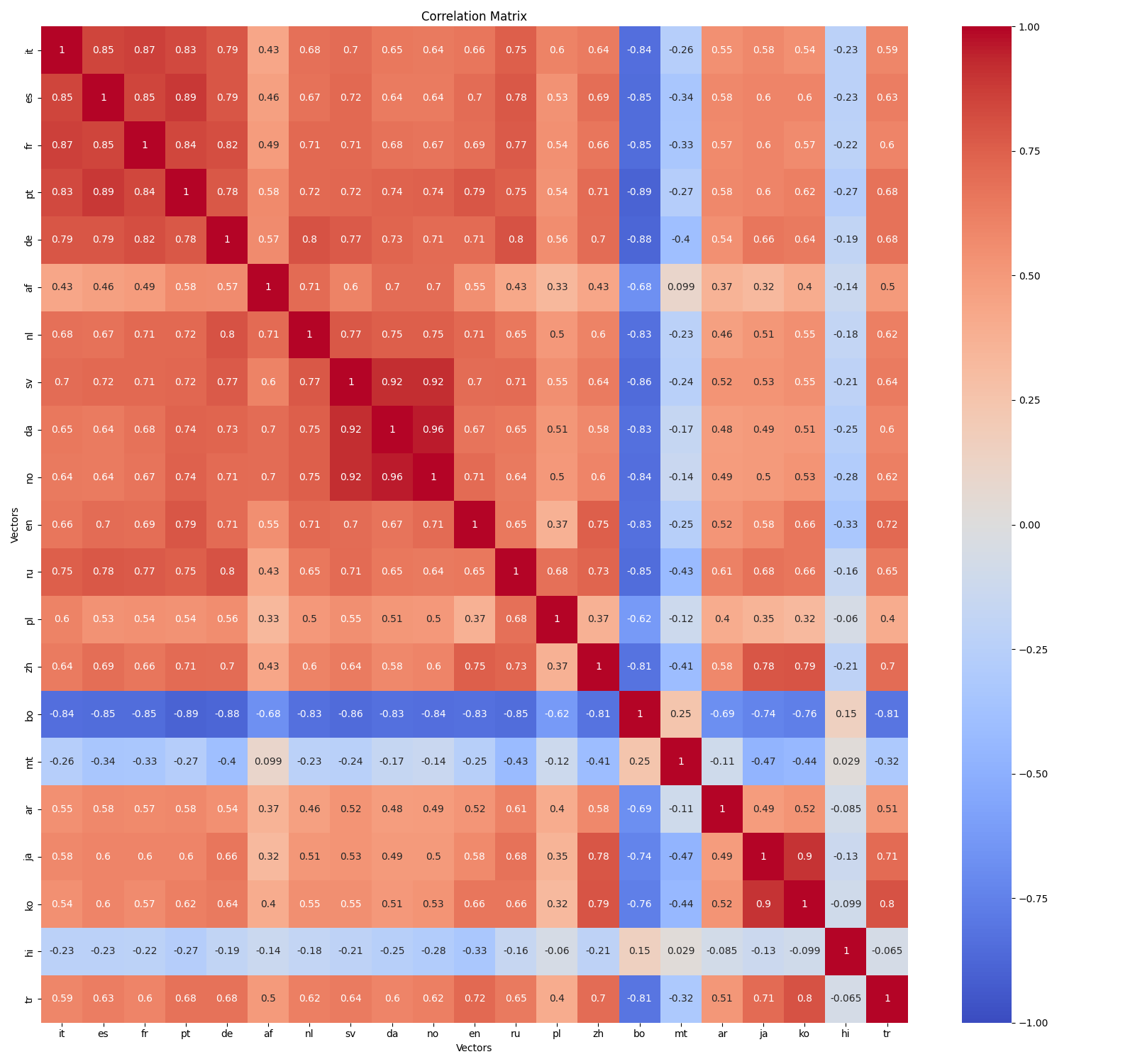} \\[0.1em]

Sep. score &0.16&0.01&0.50&0.37 \\[0.6em]

\textbf{Layer 23} &
\includegraphics[width=\linewidth]{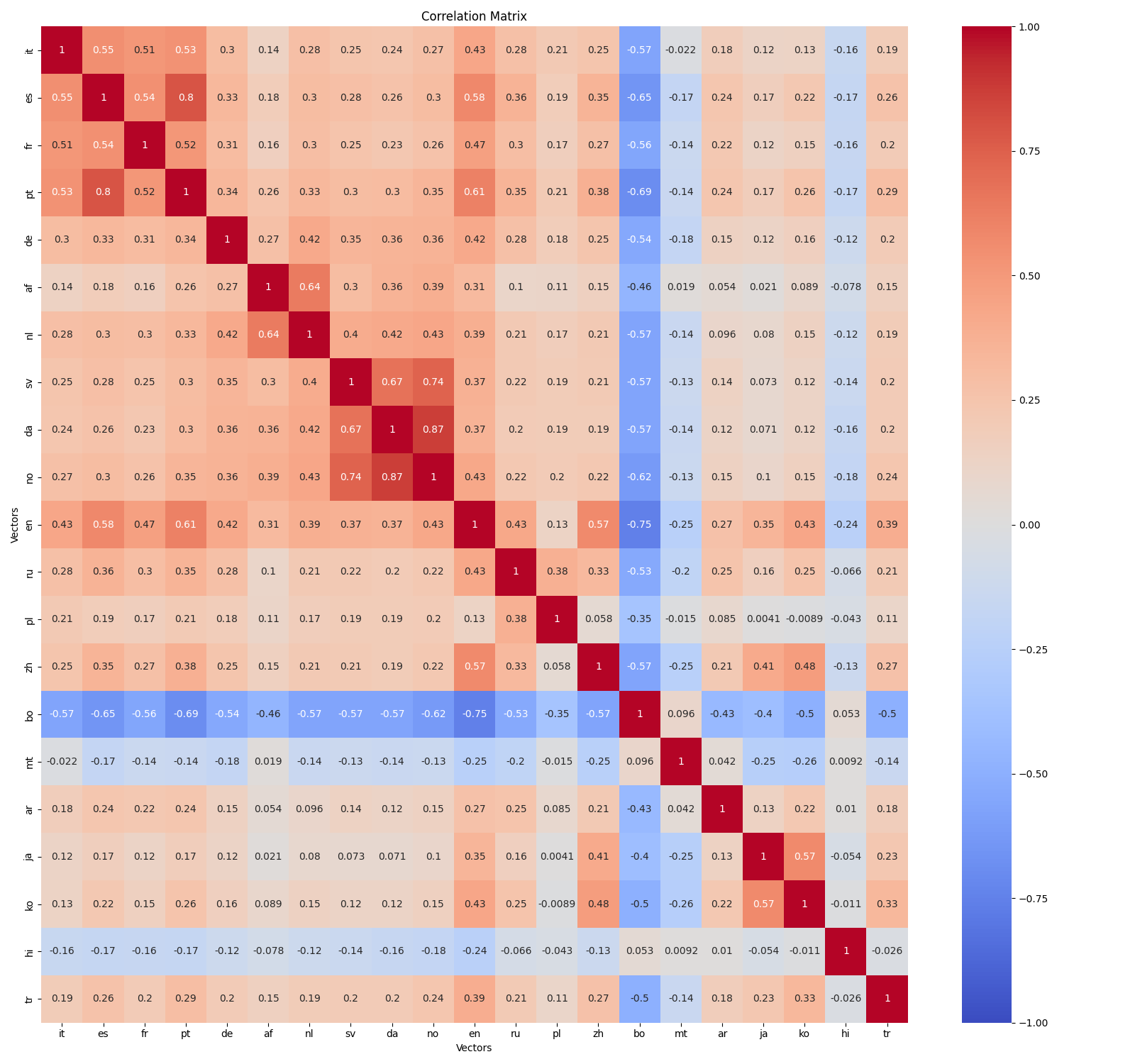} &
\includegraphics[width=\linewidth]{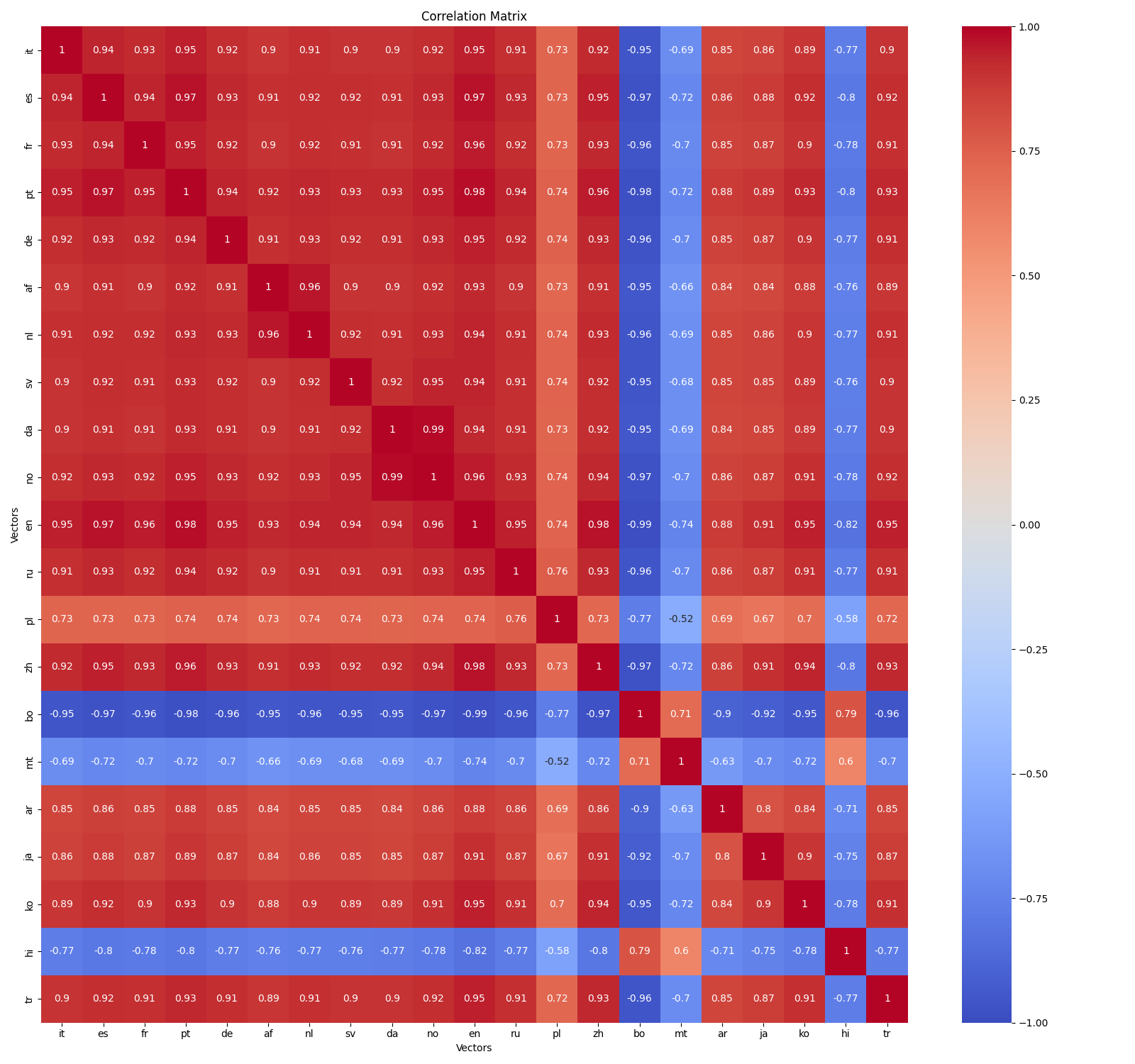} &
\includegraphics[width=\linewidth]{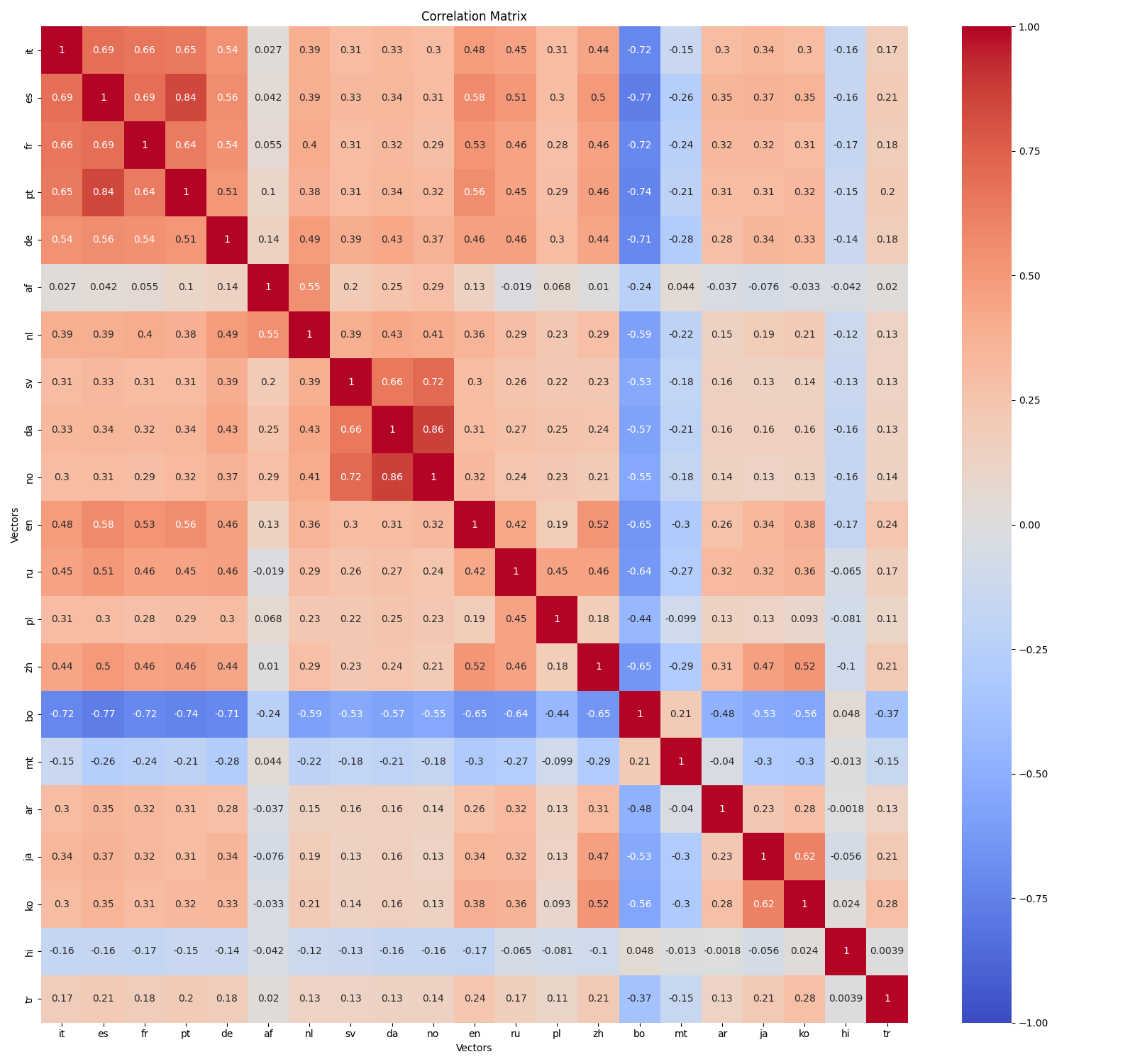} &
\includegraphics[width=\linewidth]{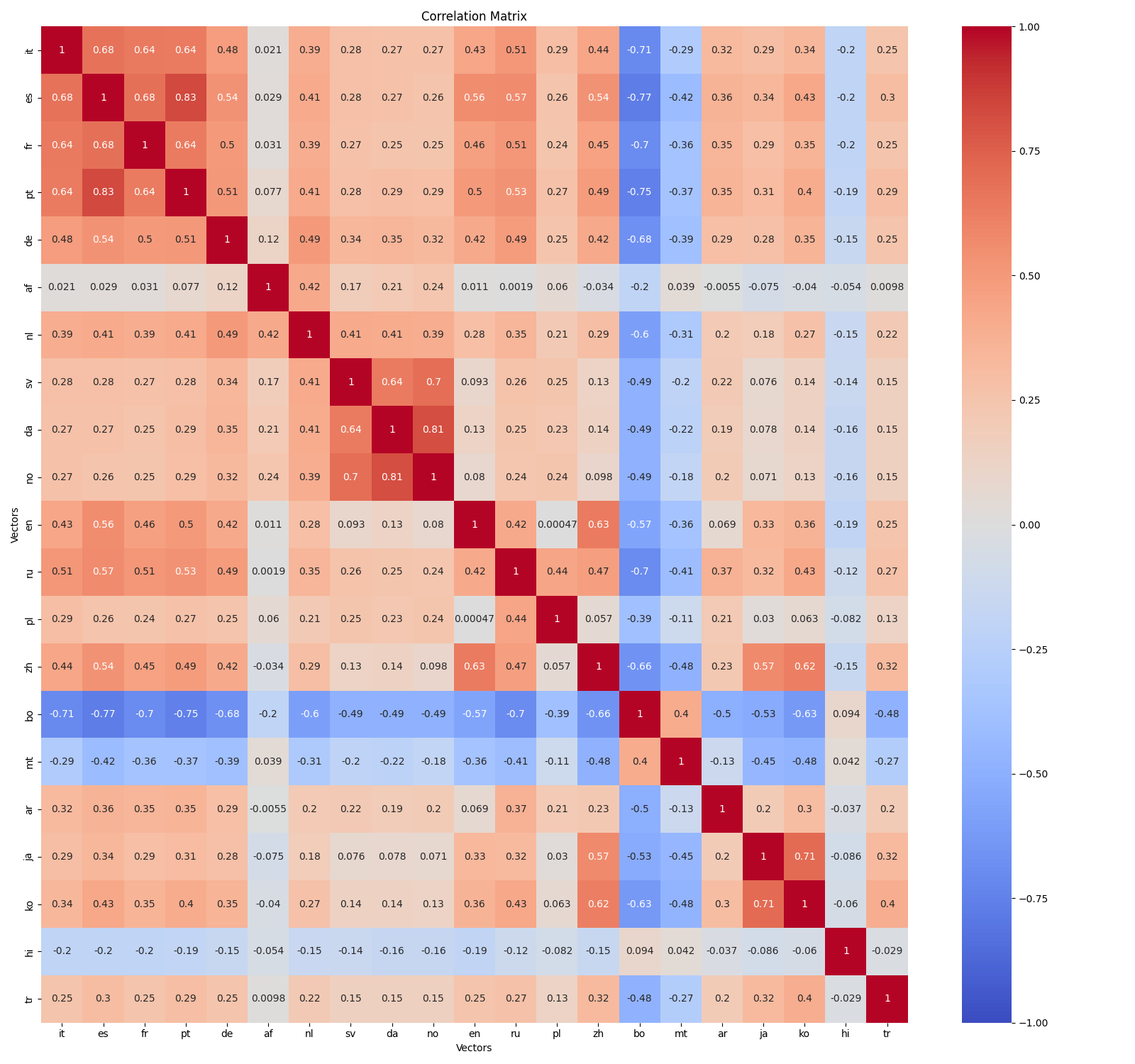} \\[0.1em]

Sep. score&0.67&0.13&0.28&0.62 \\[0.6em]

\end{tabularx}

\caption{Comparison of LLama-3.1-8B model representation space using residual stream vectors, LLama-Scope sparse space vectors. and our trained SAEs sparse space vectors.}
\label{fig:correlation_matrices}
\end{figure*}

\clearpage
\newpage

\begin{table*}[ht]
\section{Raw Results ( \texttt{tgt\_$i$} $\neq$ \texttt{steer\_$j$}) for Gemma-2-9B}
\centering
\small
\begin{tabular}{r|ccc|ccc|ccc|ccc}
\hline
\textbf{layer} &
\multicolumn{3}{c|}{\textbf{MULTI21-SAES}} &
\multicolumn{3}{c|}{\textbf{EN-SAES}} &
\multicolumn{3}{c|}{\textbf{gemma-scope}} &
\multicolumn{3}{c}{\textbf{Base Model}} \\
& \tiny LangID & \tiny ROUGEL & \tiny LASE & \tiny LangID & \tiny ROUGEL & \tiny LASE & \tiny LangID & \tiny ROUGEL & \tiny LASE & \tiny LangID & \tiny ROUGEL & \tiny LASE \\
\hline
6  & 0.0  & 0.52 & 0.0  & 0.0  & 0.53 & 0.0  & \bestone{98.94} & \comptwo{4.14} & 13.13 & - & -  & - \\
14 & \compone{48.33} & \besttwo{4.17} & \compthree{16.55} & \compone{42.92} & \comptwo{4.02} & \compthree{15.75} & 57.73 & 3.78 & \bestthree{18.53} & - & -  & - \\
23 & 11.81 & 1.25 & 12.38 & 10.79 & 1.15 & 11.29 & 21.39 & 1.78 & 11.48 & - & -  & - \\
32 & 9.35 & 1.27 & 11.39 & 5.46 & 1.12 & 10.03 & 17.04 & 1.47 & 15.79 & - & -  & - \\
40 & 0.0  & 0.59 & 0.0  & 0.0  & 0.56 & 0.0  & 0.0  & 0.6  & 0.0  & - & -  & - \\
\hline
Prompt (No steering) & - & - & -  & - & - & -  & - & -  & - & \textbf{36.48} & \textbf{3.47} & \textbf{22.87} \\
\hline
\end{tabular}
\caption{\textbf{Gemma-2-9B CrossSumm, cross-lingual steering ($\texttt{tgt}_i \neq \texttt{steer}_j$).}
Table entries report \textbf{LangID / ROUGE-L / LaSE} (column order).
\textbf{No steering prompt results} (baseline; \textbf{LangID / ROUGE-L / LaSE}): \textbf{36.48}, \textbf{3.47}, \textbf{22.87}.
\emph{Prompt} scores are computed against the \textbf{prompt language} ($\texttt{tgt}_i$), whereas \emph{steering} scores are computed against the \textbf{steering-vector language} ($\texttt{steer}_j$) and averaged over all mismatched pairs $(i,j)$, averaging across target prompt languages for each steering language; therefore the prompt baseline and steering results are not directly comparable, but the baseline usefully characterizes the model's unsteered default behavior.
\textbf{Strong shading} marks the best value \emph{overall in the table} (per metric), while \emph{light shading} marks the best value \emph{within each SAE family} (per metric).
Highlighted cells concentrate around the best layer, and the best overall results are often achieved by \textsc{MULTI21-SAEs} at that layer.}

\end{table*}

\begin{table*}[ht]
\centering
\small
\begin{tabular}{r|ccc|ccc|ccc|ccc}
\hline
\textbf{layer} &
\multicolumn{3}{c|}{\textbf{MULTI21-SAES}} &
\multicolumn{3}{c|}{\textbf{EN-SAES}} &
\multicolumn{3}{c|}{\textbf{gemma-scope}} &
\multicolumn{3}{c}{\textbf{Base Model}} \\
& \tiny LangID & \tiny SpBLEU & \tiny COMET & \tiny LangID & \tiny SpBLEU & \tiny COMET & \tiny LangID & \tiny SpBLEU & \tiny COMET & \tiny LangID & \tiny SpBLEU & \tiny COMET \\
\hline
6  & 0.02 & 1.13 & 4.07  & 0.02 & 1.13 & 4.07  & \bestone{74.39} & 8.03  & 49.61 & - & - & - \\
14 & \compone{54.38} & \besttwo{24.80} & \bestthree{73.55} & 52.19 & \besttwo{24.90} & \compthree{73.17} & 45.04 & \comptwo{15.65} & \compthree{61.79} & - & - & - \\
23 & 24.33 & 19.73 & 58.23 & 21.73 & 18.90 & 55.26 & 25.26 & 12.49 & 44.24 & - & - & - \\
32 & 17.12 & 15.87 & 47.36 & 13.19 & 15.84 & 46.67 & 30.43 & \comptwo{16.12} & 52.00 & - & - & - \\
40 & 0.04 & 2.70 & 9.28  & 0.03 & 2.46 & 8.25  & 0.05 & 4.74  & 13.34 & - & - & - \\
\hline
Prompt (No steering) & - & - & -  & - & - & -  & - & -  & - & \textbf{75.51} & \textbf{31.31} & \textbf{85.12} \\
\hline
\end{tabular}
\caption{\textbf{Gemma-2-9B FLORES, cross-lingual steering ($\texttt{tgt}_i \neq \texttt{steer}_j$).}
Table entries report \textbf{LangID / SpBLEU / COMET} (column order).
\textbf{No steering prompt results} (baseline; \textbf{LangID / SpBLEU / COMET}): \textbf{75.51}, \textbf{31.31}, \textbf{85.12}.
\emph{Prompt} scores are computed against the \textbf{prompt language} ($\texttt{tgt}_i$), whereas \emph{steering} scores are computed against the \textbf{steering-vector language} ($\texttt{steer}_j$) and averaged over all mismatched pairs $(i,j)$, averaging across target prompt languages for each steering language; therefore the prompt baseline and steering results are not directly comparable, but the baseline usefully characterizes the model's unsteered default behavior.
\textbf{Strong shading} marks the best value \emph{overall in the table} (per metric), while \emph{light shading} marks the best value \emph{within each SAE family} (per metric).
Highlighted cells concentrate around the best layer, and the best overall results are often achieved by \textsc{MULTI21-SAEs} at that layer.}
\end{table*}

\clearpage

\newpage

\begin{table*}[ht]
\section{Raw Results ( \texttt{tgt\_$i$} $\neq$ \texttt{steer\_$j$}) for LLaMA-3.1-8B}
\centering
\small
\begin{tabular}{r|ccc|ccc|ccc|ccc}
\hline
\textbf{layer} &
\multicolumn{3}{c|}{\textbf{MULTI21-SAES}} &
\multicolumn{3}{c|}{\textbf{EN-SAES}} &
\multicolumn{3}{c|}{\textbf{llama-scope}} &
\multicolumn{3}{c}{\textbf{Base Model}} \\
& \tiny LangID & \tiny ROUGEL & \tiny LASE & \tiny LangID & \tiny ROUGEL & \tiny LASE & \tiny LangID & \tiny ROUGEL & \tiny LASE & \tiny LangID & \tiny ROUGEL & \tiny LASE \\
\hline
6  & \compone{78.70} & 3.26 & 15.79 & \bestone{79.86} & 2.92 & 16.17 & 0.00 & 0.18 & 0.00 & - & -  & - \\
13 & 66.25 & \comptwo{3.90} & 24.89 & 59.31 & 3.65 & \compthree{27.54} & 0.00 & 0.29 & 0.00 & - & -  & - \\
15 & 30.46 & 2.12 & \bestthree{30.47} & 44.49 & \besttwo{4.64} & \compthree{27.18} & 0.00 & 0.26 & 0.00 & - & -  & - \\
19 & 3.80  & 1.84 & 17.61 & 6.76  & 2.45 & 9.89  & 0.00 & 0.01 & 0.00 & - & -  & - \\
25 & 0.00  & 1.50 & 0.00  & 0.00  & 1.49 & 0.00  & 0.00 & 0.18 & 0.00 & - & -  & - \\
\hline
Prompt (No steering) & - & - & -  & - & - & -  & - & -  & - & \textbf{16.85} & \textbf{2.87} & \textbf{32.92} \\
\hline
\end{tabular}
\caption{\textbf{LLaMA-3.1-8B CrossSumm, cross-lingual steering ($\texttt{tgt}_i \neq \texttt{steer}_j$).}
Table entries report \textbf{LangID / ROUGE-L / LaSE} (column order).
\textbf{No steering prompt results} (baseline; \textbf{LangID / ROUGE-L / LaSE}): \textbf{16.85}, \textbf{2.87}, \textbf{32.92}.
\emph{Prompt} scores are computed against the \textbf{prompt language} ($\texttt{tgt}_i$), whereas \emph{steering} scores are computed against the \textbf{steering-vector language} ($\texttt{steer}_j$) and averaged over all mismatched pairs $(i,j)$, averaging across target prompt languages for each steering language; therefore the prompt baseline and steering results are not directly comparable, but the baseline usefully characterizes the model's unsteered default behavior.
\textbf{Strong shading} marks the best value \emph{overall in the table} (per metric), while \emph{light shading} marks the best value \emph{within each SAE family} (per metric).
Highlighted cells concentrate around the best layer, and the best overall results are often achieved by \textsc{MULTI21-SAEs} at that layer.}
\end{table*}

\begin{table*}[ht]
\centering
\small
\begin{tabular}{r|ccc|ccc|ccc|ccc}
\hline
\textbf{layer} &
\multicolumn{3}{c|}{\textbf{MULTI21-SAES}} &
\multicolumn{3}{c|}{\textbf{EN-SAES}} &
\multicolumn{3}{c|}{\textbf{llama-scope}} &
\multicolumn{3}{c}{\textbf{Base Model}} \\
& \tiny LangID & \tiny SpBLEU & \tiny COMET & \tiny LangID & \tiny SpBLEU & \tiny COMET & \tiny LangID & \tiny SpBLEU & \tiny COMET & \tiny LangID & \tiny SpBLEU & \tiny COMET \\
\hline
6  & 54.22 & 4.42  & 39.77 & \bestone{68.77} & 3.38  & 36.55 & 2.36 & 0.02 & 6.33 & - & -  & - \\
13 & \compone{58.24} & 16.11 & 66.27 & 62.14 & 12.20 & 60.44 & 0.47 & 0.02 & 9.65 & - & -  & - \\
15 & 56.97 & \comptwo{22.53} & \bestthree{73.25} & 60.92 & 21.02 & \compthree{71.57} & 0.10 & 0.00 & 2.72 & - & -  & - \\
19 & 24.14 & \besttwo{23.35} & 68.65 & 21.32 & \comptwo{22.44} & 62.75 & 0.12 & 0.01 & 1.86 & - & -  & - \\
25 & 0.09  & 3.77  & 11.25 & 0.64  & 12.12 & 31.88 & 0.09 & 0.01 & 2.13 & - & -  & - \\
\hline
Prompt (No steering) & - & - & -  & - & - & -  & - & -  & - & \textbf{91.06} & \textbf{31.22} & \textbf{83.58} \\
\hline
\end{tabular}
\caption{\textbf{LLaMA-3.1-8B FLORES, cross-lingual steering ($\texttt{tgt}_i \neq \texttt{steer}_j$).}
Table entries report \textbf{LangID / SpBLEU / COMET} (column order).
\textbf{No steering prompt results} (baseline; \textbf{LangID / SpBLEU / COMET}): \textbf{91.06}, \textbf{31.22}, \textbf{83.58}.
\emph{Prompt} scores are computed against the \textbf{prompt language} ($\texttt{tgt}_i$), whereas \emph{steering} scores are computed against the \textbf{steering-vector language} ($\texttt{steer}_j$) and averaged over all mismatched pairs $(i,j)$, averaging across target prompt languages for each steering language; therefore the prompt baseline and steering results are not directly comparable, but the baseline usefully characterizes the model's unsteered default behavior.
\textbf{Strong shading} marks the best value \emph{overall in the table} (per metric), while \emph{light shading} marks the best value \emph{within each SAE family} (per metric).
Highlighted cells concentrate around the best layer, and the best overall results are often achieved by \textsc{MULTI21-SAEs} at that layer.}
\end{table*}

\clearpage
\newpage

\begin{figure*}[ht]
\section{Per-Language Results ( \texttt{tgt\_$i$} $=$ \texttt{steer\_$j$}) for Gemma-2-9B}
\label{app:per_lang_same_res}
    \centering
    \includegraphics[width=0.8\linewidth]{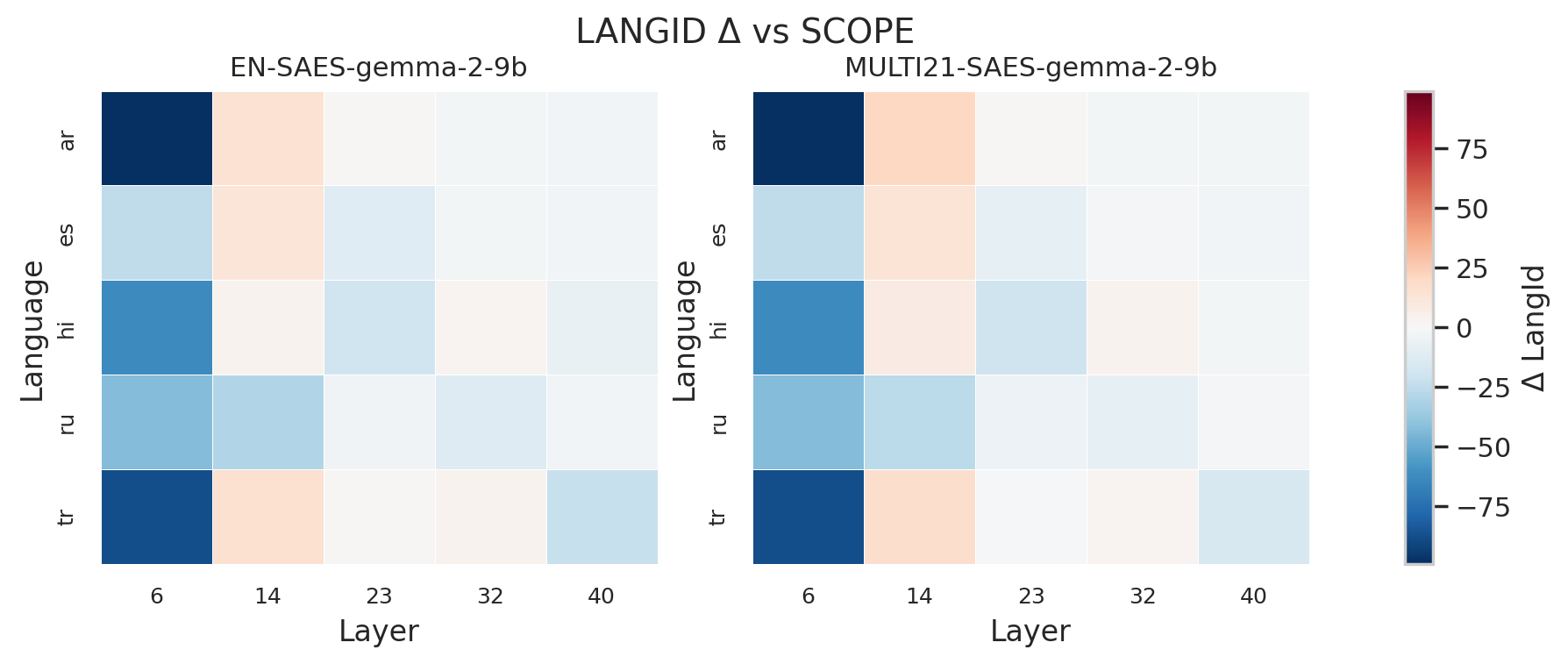}
    \includegraphics[width=0.8\linewidth]{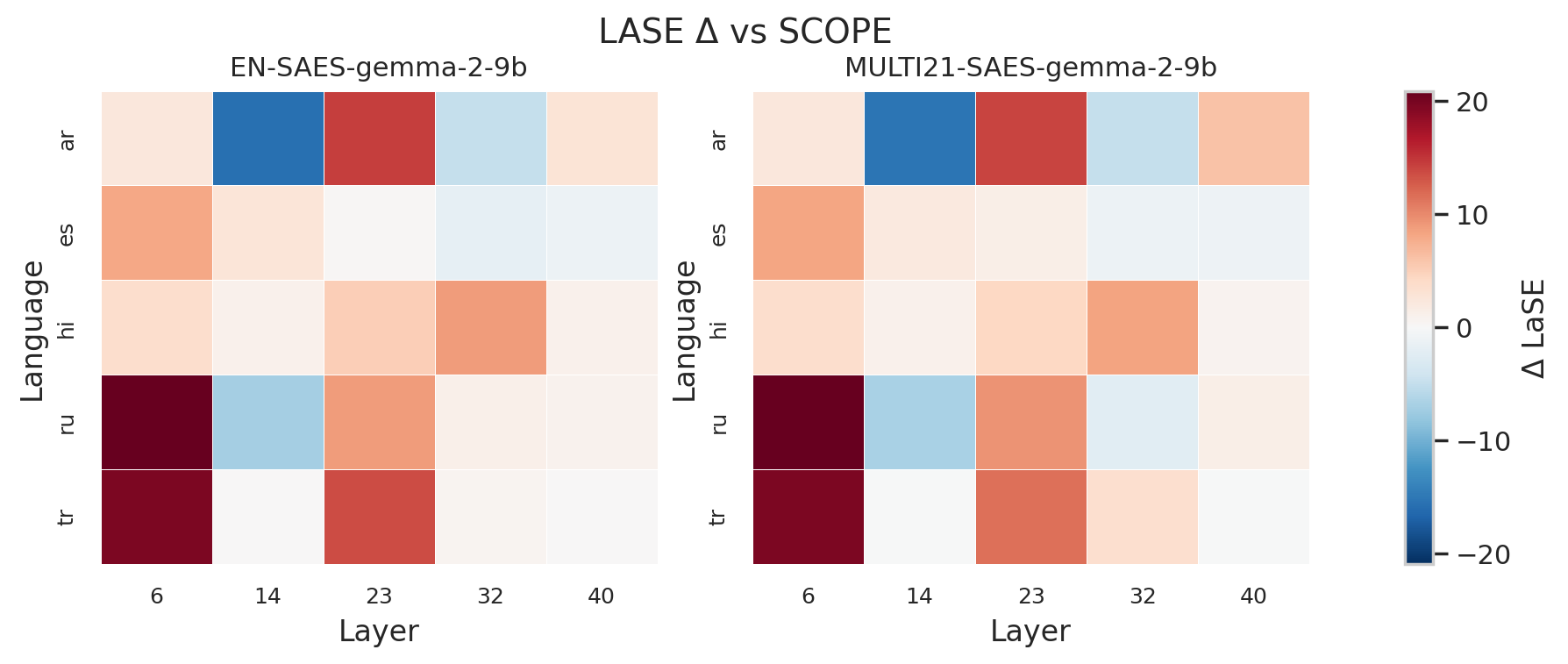}
    \includegraphics[width=0.8\linewidth]{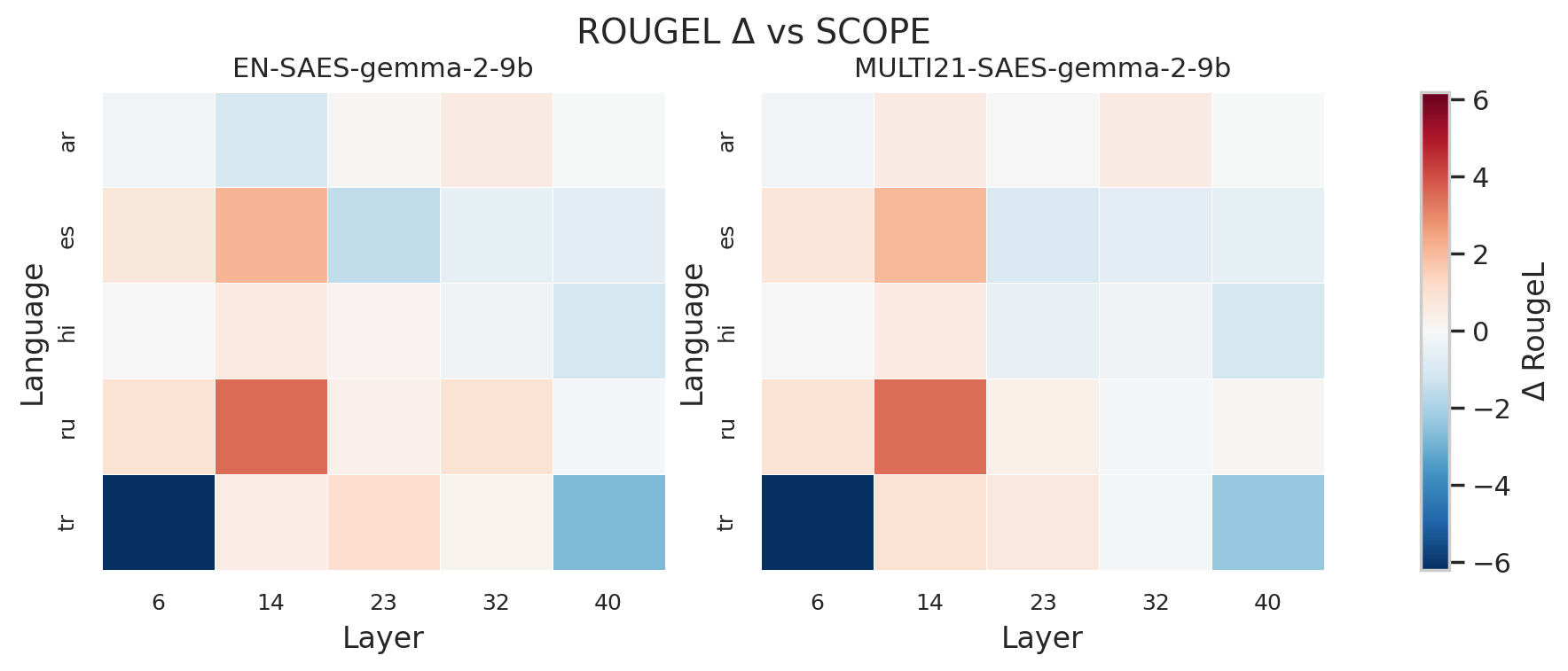}

    \caption{
    Per-language, per-layer performance deltas for \textbf{Gemma-2-9B} on the \textsc{CrossSum} task when the steering language matches the target language (\texttt{tgt\_$i$} = \texttt{steer\_$j$}). 
    Each heatmap shows the change relative to the SCOPE baseline (excluded), with rows corresponding to target languages, columns to transformer layers, and separate panels for each SAE variant. 
    Positive values indicate improvements over the baseline.
    }

    \label{fig:gemma_cross_sum_all_same}
\end{figure*}

\begin{figure*}[ht]

    \centering

    \includegraphics[width=0.8\linewidth]{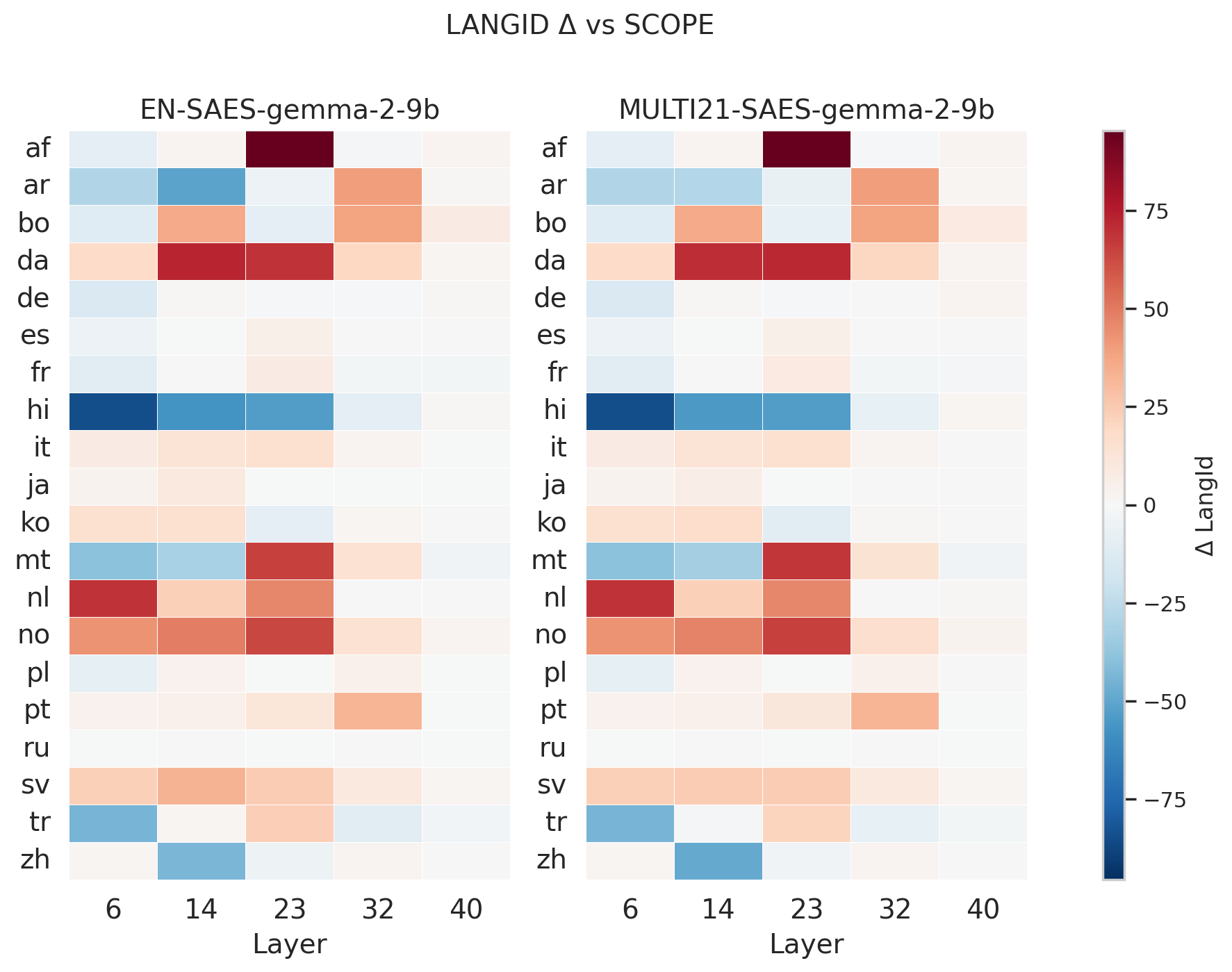}
    \includegraphics[width=0.8\linewidth]{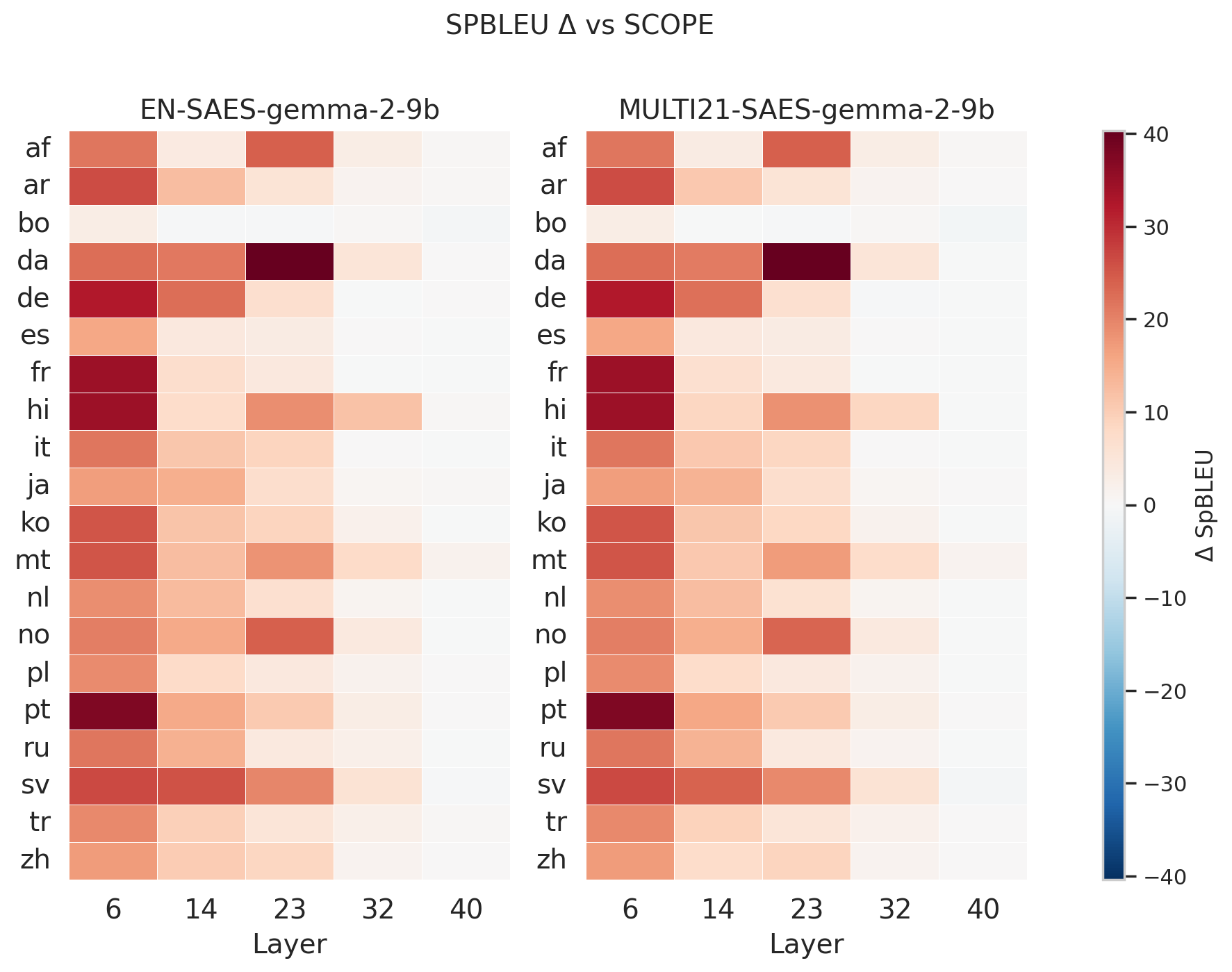}

    \caption{
Per-language, per-layer deltas for \textbf{Gemma-2-9B} on \textsc{FLORES} under matched steering and target languages (\texttt{tgt\_$i$} = \texttt{steer\_$j$}). 
The heatmaps show the impact of SAE variants on language identification and translation quality across model depth.
}

    \label{fig:gemma_flores_2_same}
\end{figure*}

\begin{figure*}[ht]

    \centering

    \includegraphics[width=0.8\linewidth]{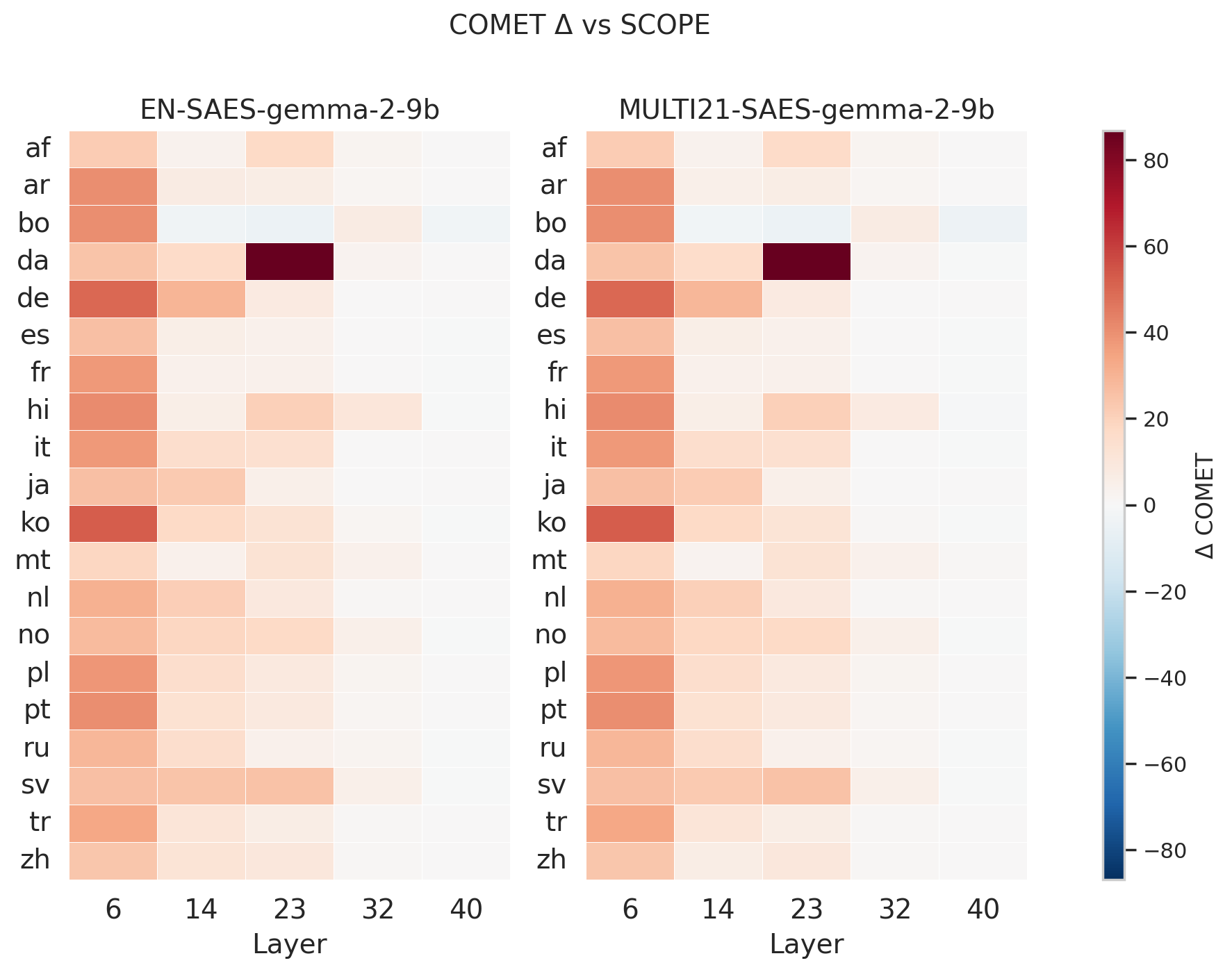}

    \caption{
Per-language, per-layer COMET score deltas for \textbf{Gemma-2-9B} on \textsc{FLORES} with matched steering and target languages (\texttt{tgt\_$i$} = \texttt{steer\_$j$}). 
This figure emphasizes how semantic translation quality varies across languages, layers, and SAE configurations relative to the SCOPE baseline.
}

    \label{fig:gemma_flores_1_same}
\end{figure*}

\clearpage

\newpage

\begin{figure*}[ht]
\section{Per-Language Results ( \texttt{tgt\_$i$} $\neq$ \texttt{steer\_$j$}) for Gemma-2-9B}
\label{app:per_lang_diff_res}
    \centering
    \includegraphics[width=0.8\linewidth]{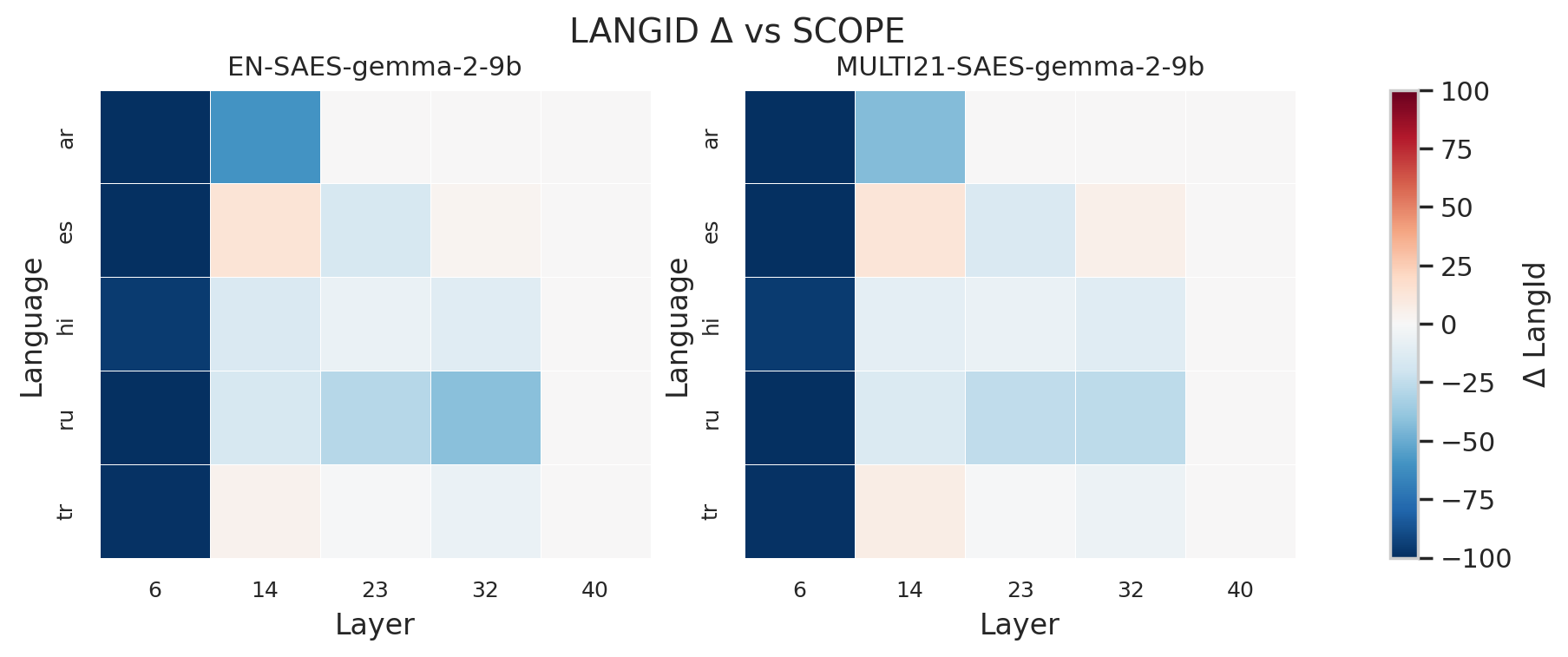}
    \includegraphics[width=0.8\linewidth]{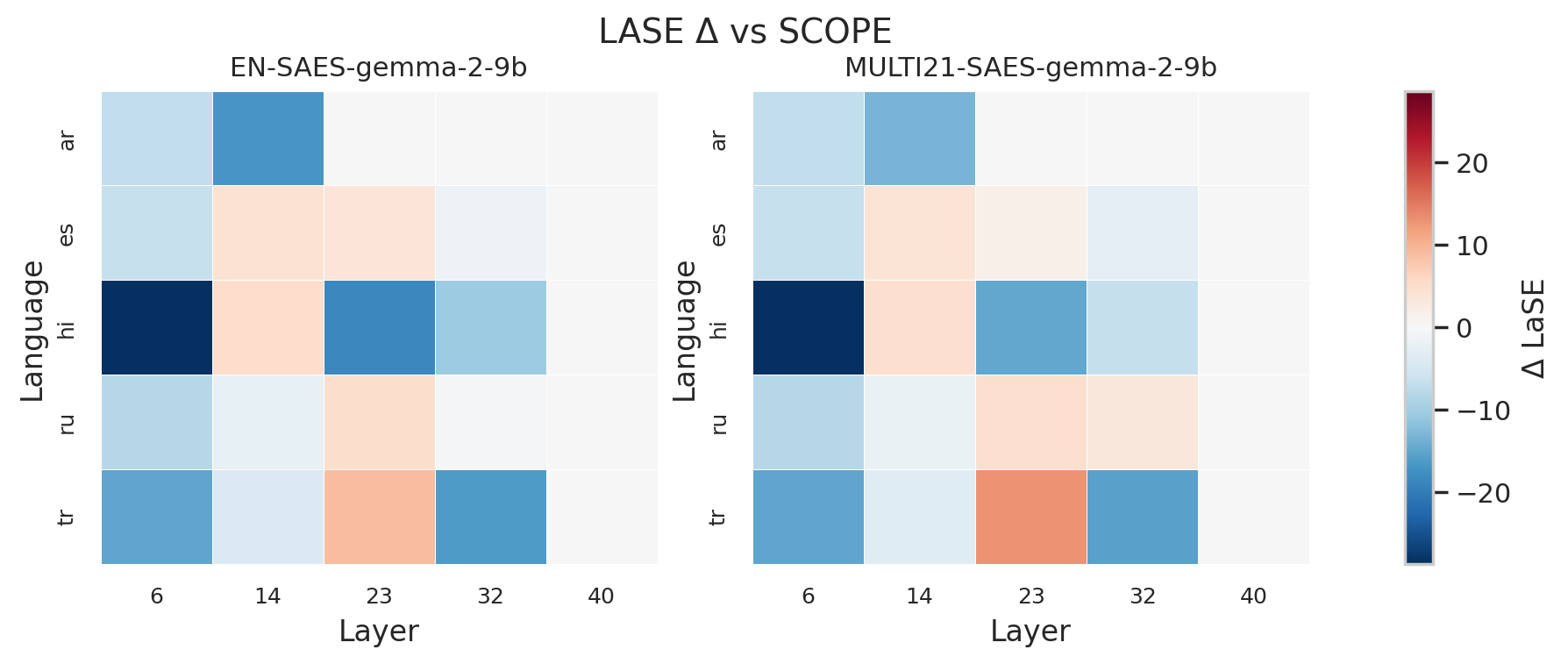}
    \includegraphics[width=0.8\linewidth]{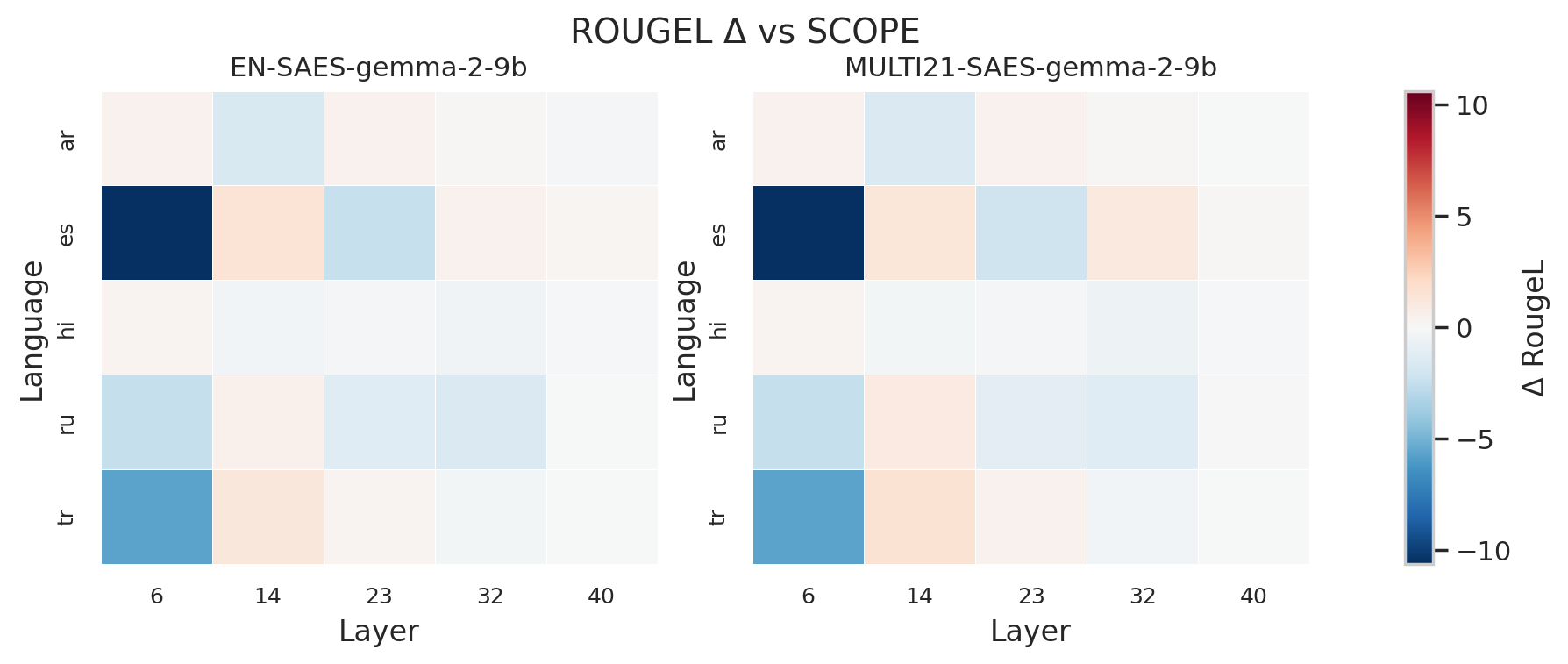}

    \caption{
Per-language, per-layer performance deltas for \textbf{Gemma-2-9B} on the \textsc{CrossSum} task under cross-lingual steering (\texttt{tgt\_$i$} $\neq$ \texttt{steer\_$j$}). 
Each heatmap shows how steering in a different language affects summarization quality and language identification across layers and SAE variants, relative to the SCOPE baseline.
}

    \label{fig:gemma_cross_sum_all_diff}
\end{figure*}

\begin{figure*}[ht]

    \centering

    \includegraphics[width=0.8\linewidth]{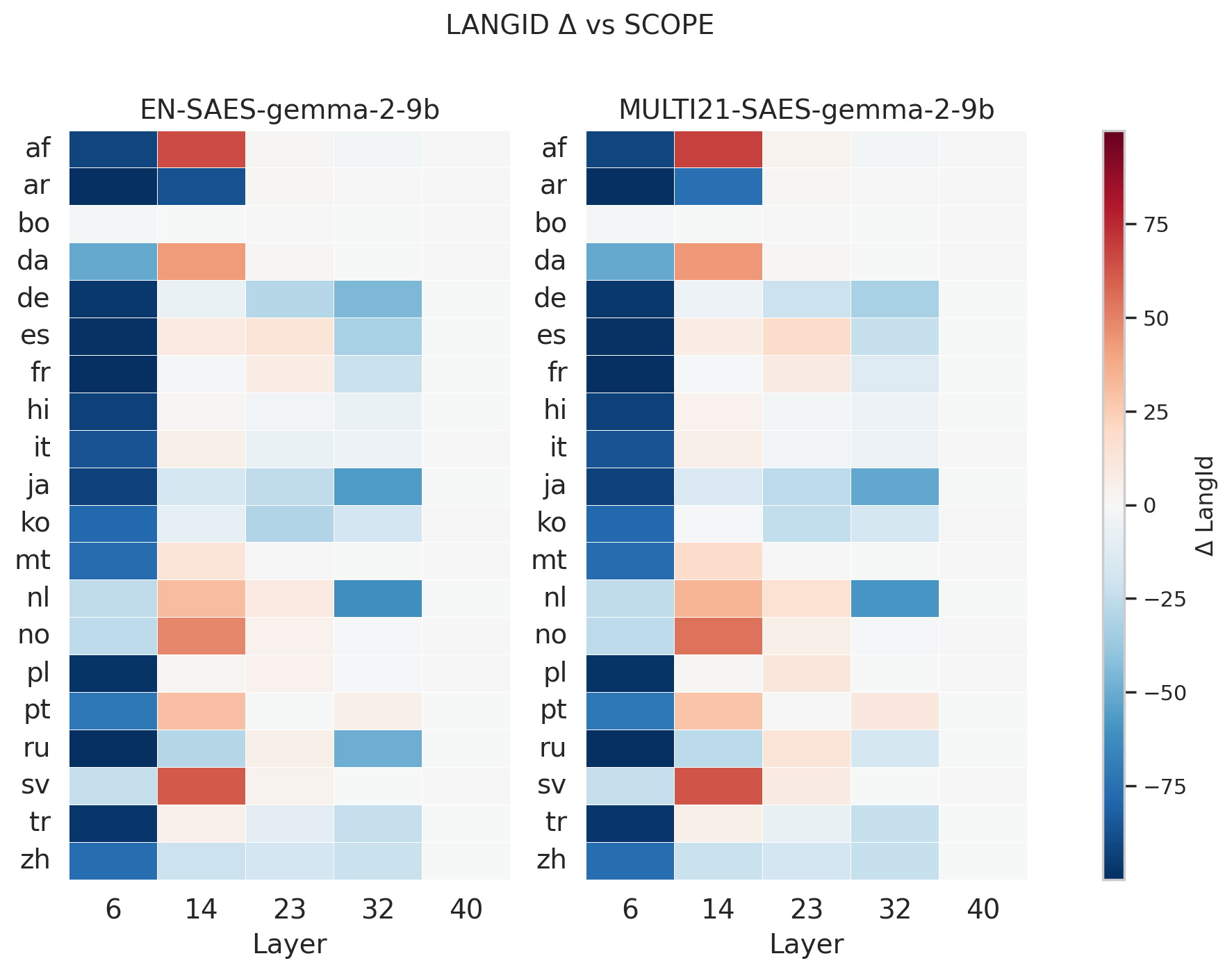}
    \includegraphics[width=0.8\linewidth]{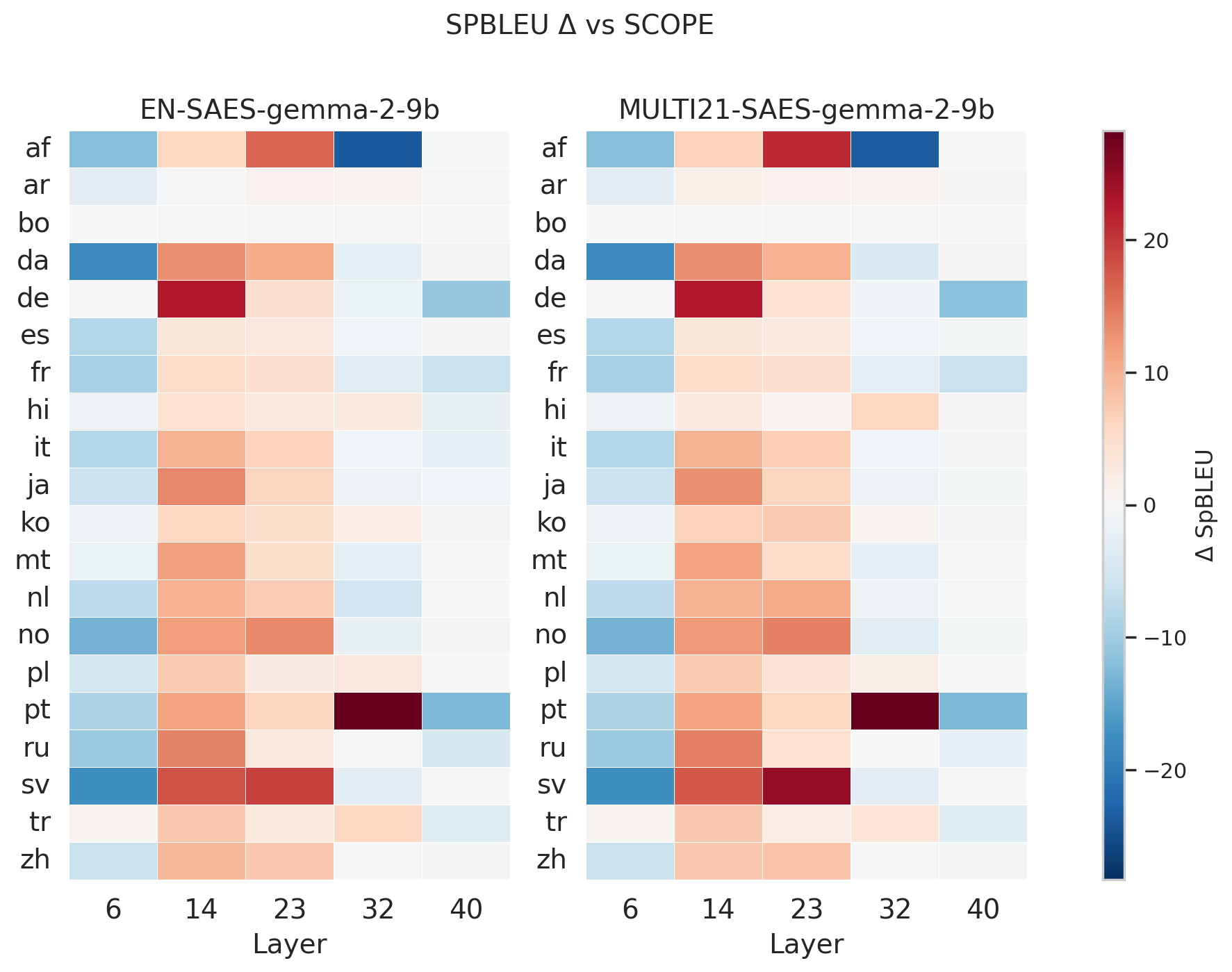}

    \caption{
Per-language, per-layer performance deltas for \textbf{Gemma-2-9B} on \textsc{FLORES} with cross-lingual steering (\texttt{tgt\_$i$} $\neq$ \texttt{steer\_$j$}). 
The figure highlights the degradation or transfer effects induced by mismatched steering languages across model depth.
}

    \label{fig:gemma_flores_2_diff}
\end{figure*}

\begin{figure*}[ht]

    \centering

    \includegraphics[width=0.8\linewidth]{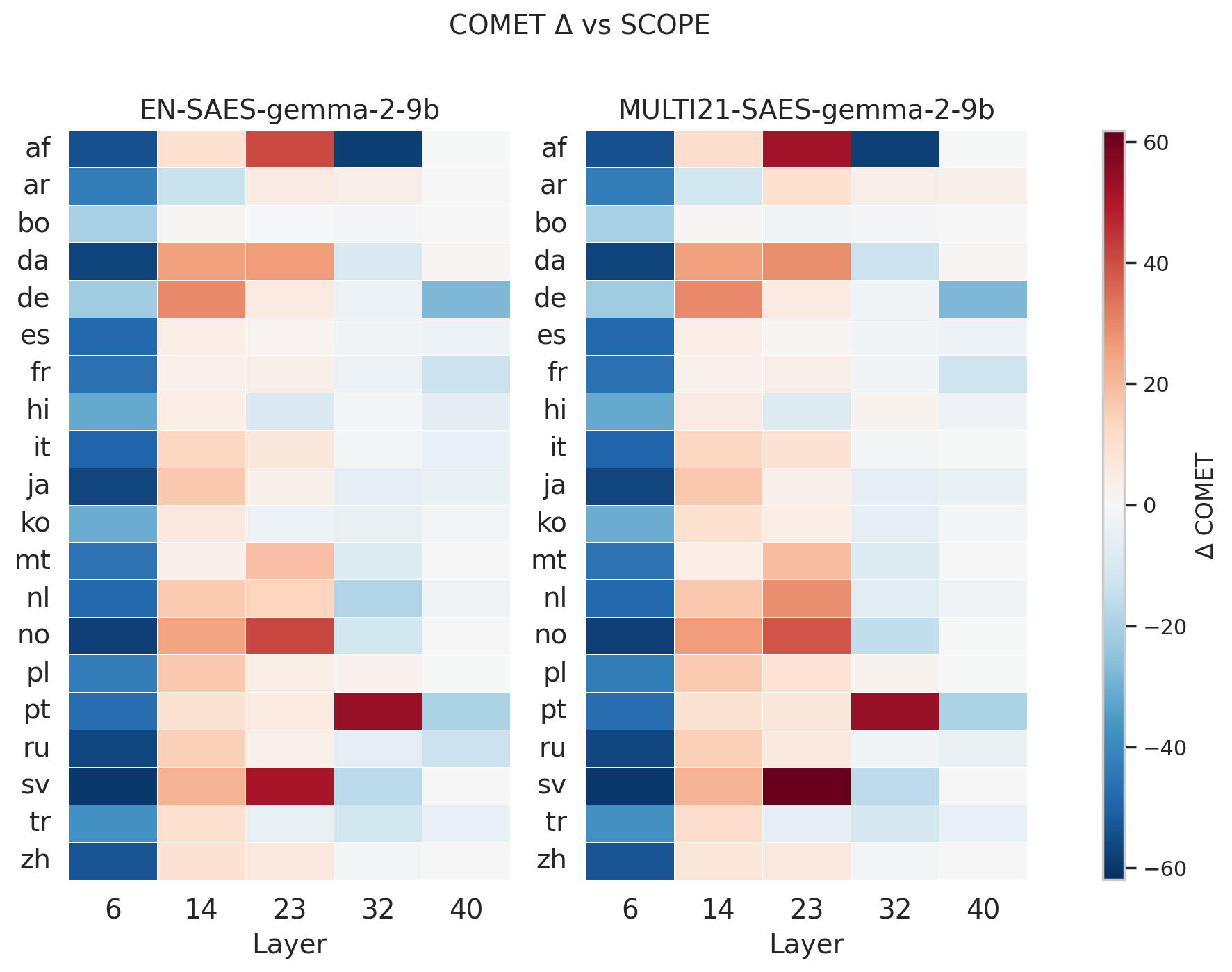}

    \caption{
Per-language, per-layer COMET score deltas for \textbf{Gemma-2-9B} on \textsc{FLORES} under cross-lingual steering (\texttt{tgt\_$i$} $\neq$ \texttt{steer\_$j$}). 
This visualization captures how semantic translation quality responds to cross-lingual steering at different layers and SAE variants, relative to the SCOPE baseline.
}

    \label{fig:gemma_flores_1_diff}
\end{figure*}

\clearpage
\newpage

\begin{figure*}[ht]
\section{Per-Language Results ( \texttt{tgt\_$i$} $=$ \texttt{steer\_$j$}) for LLaMA-3.1-8B}
\label{app:per_lang_same_res}
    \centering
    \includegraphics[width=0.8\linewidth]{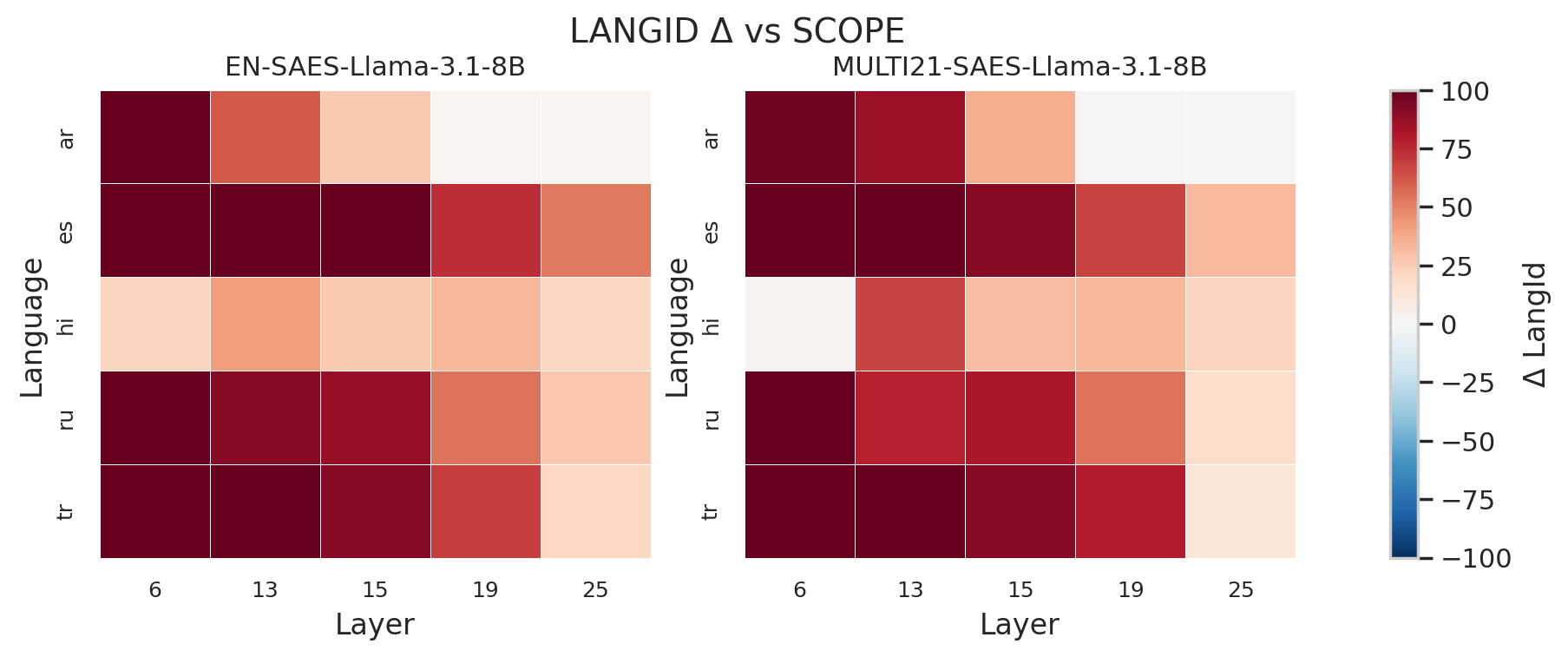}
    \includegraphics[width=0.8\linewidth]{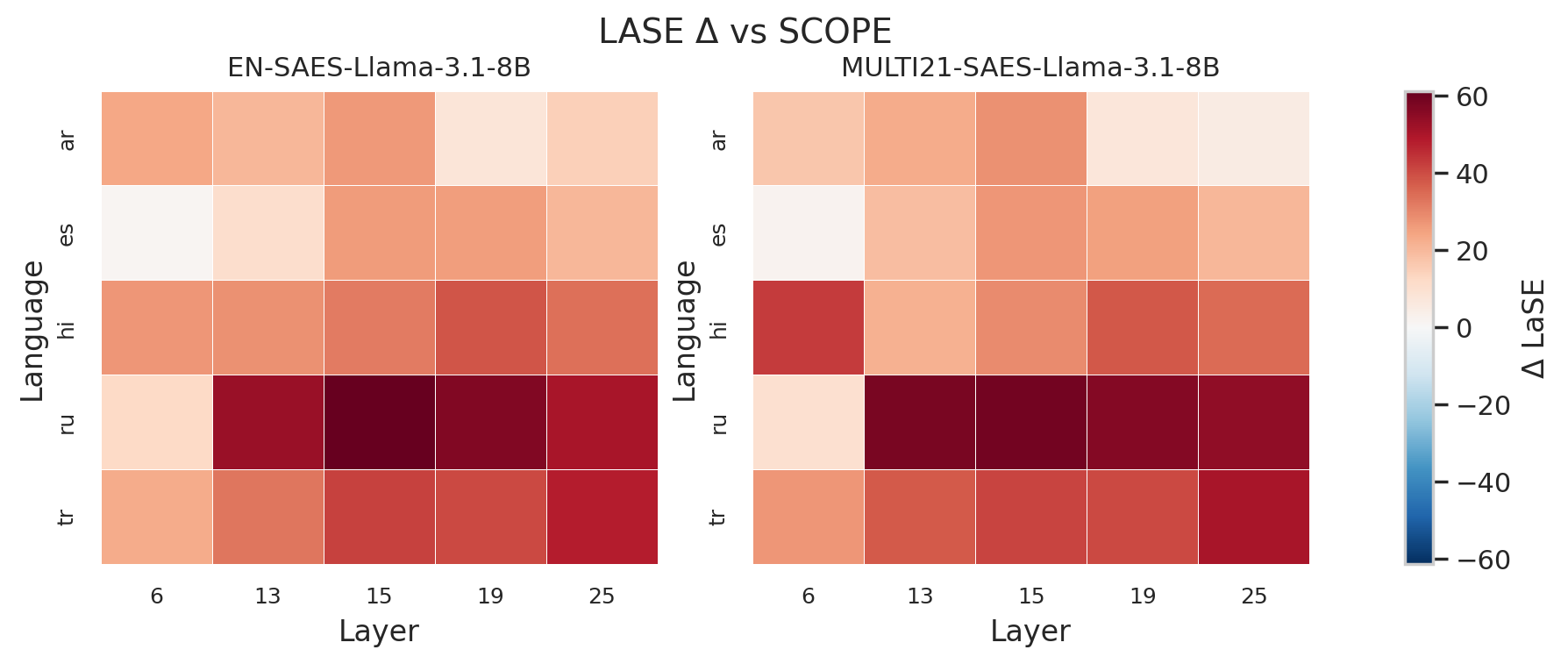}
    \includegraphics[width=0.8\linewidth]{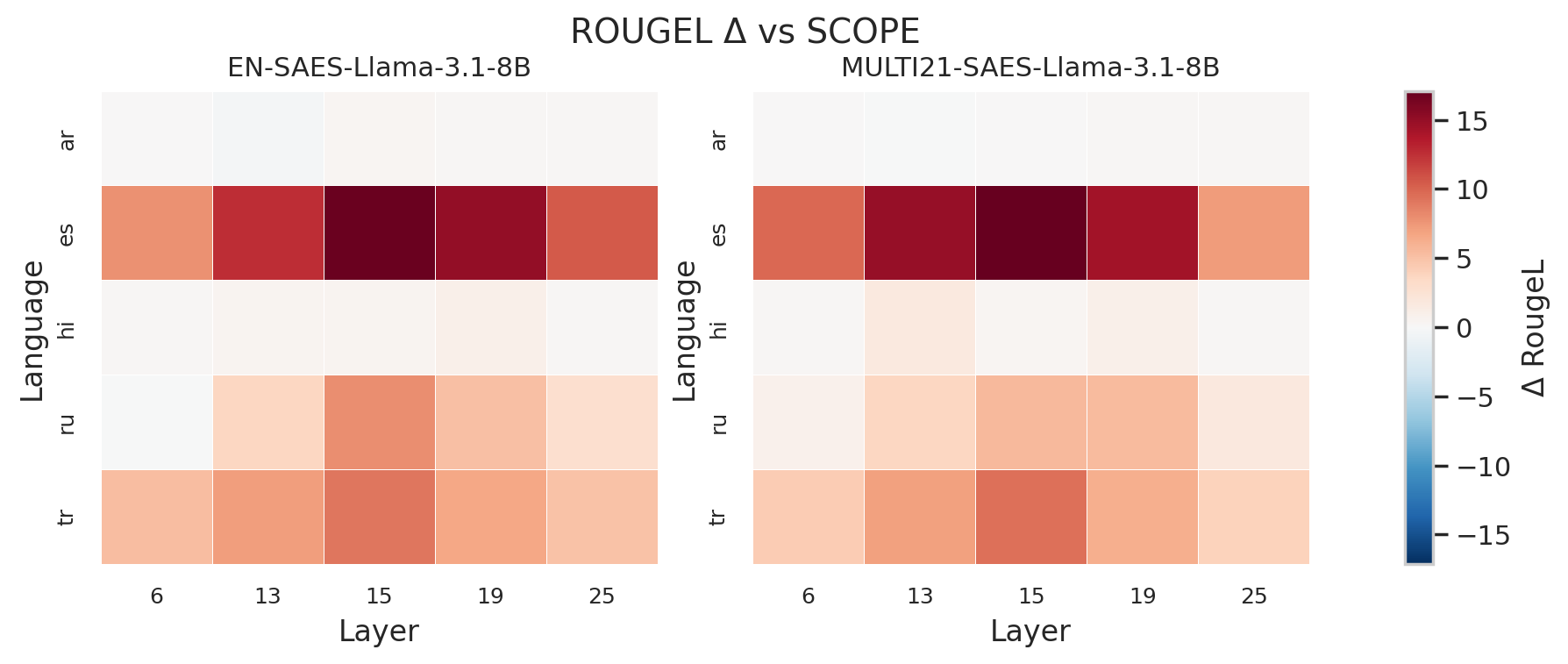}
    
    \caption{
    Per-language, per-layer performance deltas for \textbf{LLaMA-3.1-8B} on the \textsc{CrossSum} task when the steering language matches the target language (\texttt{tgt\_$i$} = \texttt{steer\_$j$}). 
    Each heatmap shows the change relative to the SCOPE baseline (excluded), with rows corresponding to target languages, columns to transformer layers, and separate panels for each SAE variant. 
    Positive values indicate improvements over the baseline.
    }
    
    \label{fig:llama_cross_sum_all_same}
\end{figure*}

\begin{figure*}[ht]
    \centering

    \includegraphics[width=0.8\linewidth]{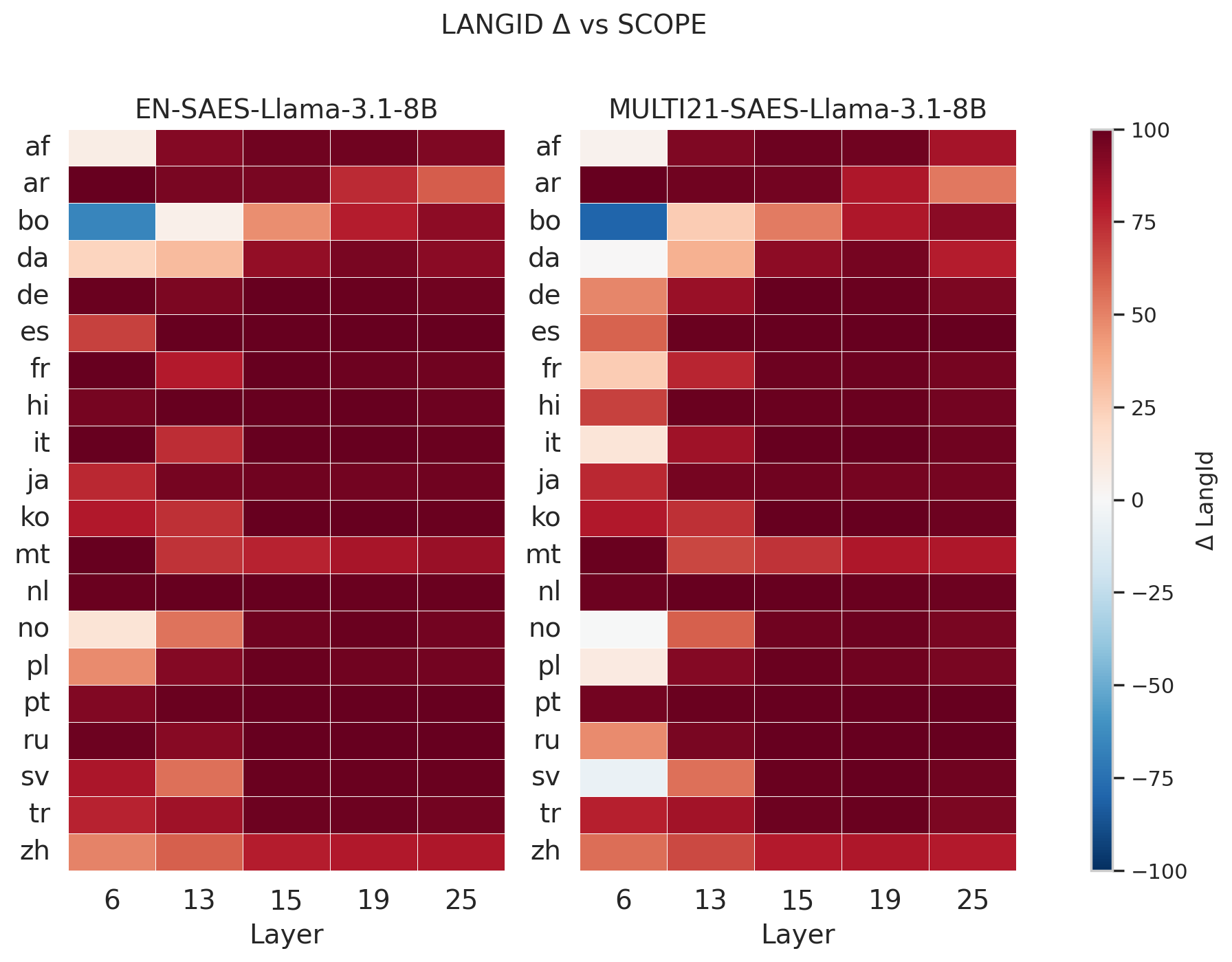}
    \includegraphics[width=0.8\linewidth]{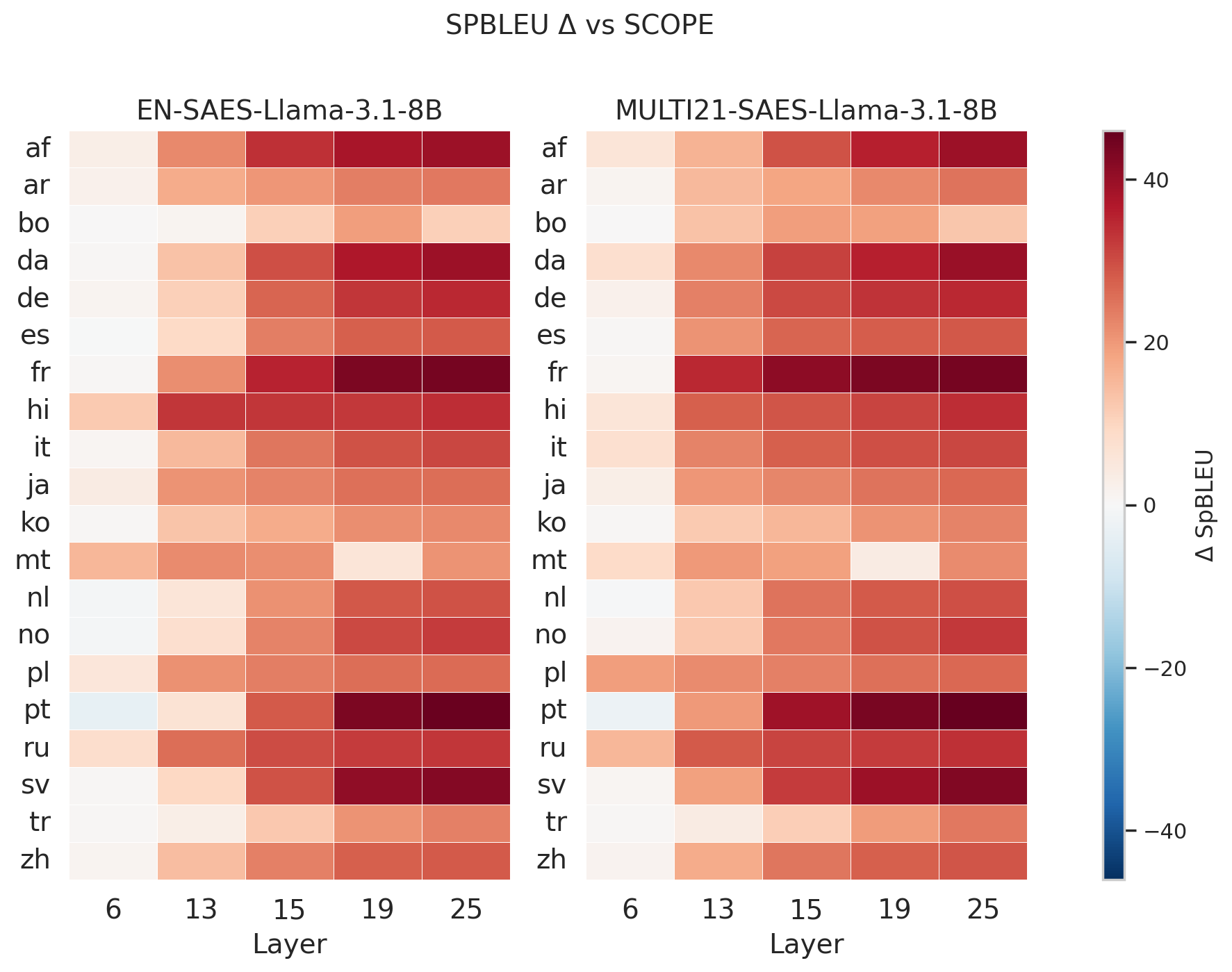}

    \caption{
    Per-language, per-layer performance deltas for \textbf{LLaMA-3.1-8B} on the \textsc{FLORES} benchmark when the steering language matches the target language (\texttt{tgt\_$i$} = \texttt{steer\_$j$}). 
    Results are shown for language identification (LangID) and translation quality (SpBLEU), aggregated per SAE variant and measured relative to the SCOPE baseline.
    }

    \label{fig:llama_flores_2_same}
\end{figure*}

\begin{figure*}[ht]
    \centering

    \includegraphics[width=0.8\linewidth]{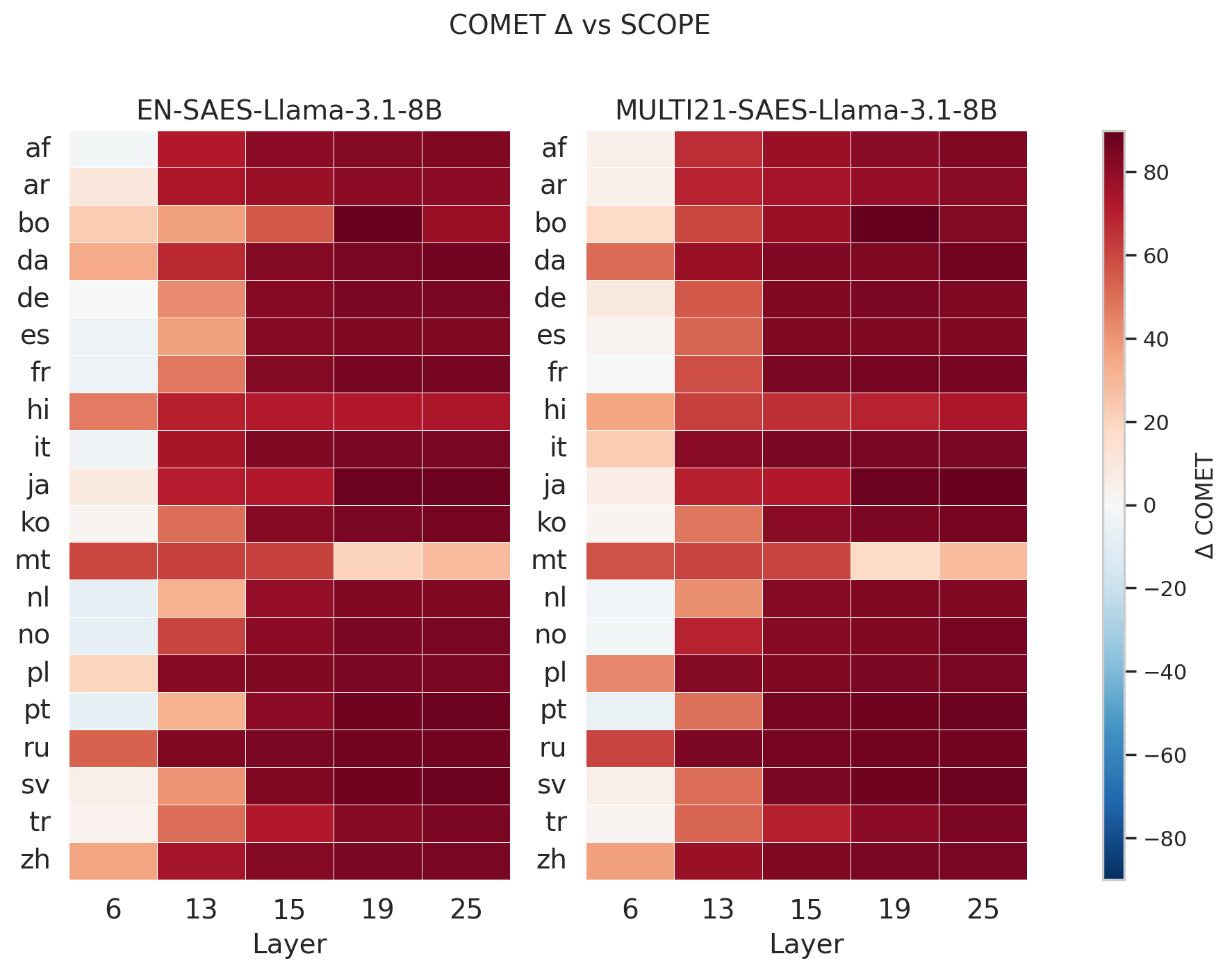}

    \caption{
    Per-language, per-layer COMET score deltas for \textbf{LLaMA-3.1-8B} on \textsc{FLORES} under matched steering and target languages (\texttt{tgt\_$i$} = \texttt{steer\_$j$}). 
    The heatmap highlights how SAE interventions affect semantic translation quality across languages and model depth, relative to the SCOPE baseline.
    }

    \label{fig:llama_flores_1_same}
\end{figure*}

\clearpage

\newpage

\begin{figure*}[ht]
\section{Per-Language Results ( \texttt{tgt\_$i$} $\neq$ \texttt{steer\_$j$}) for LLaMA-3.1-8B}
\label{app:per_lang_diff_res}
    \centering
    \includegraphics[width=0.8\linewidth]{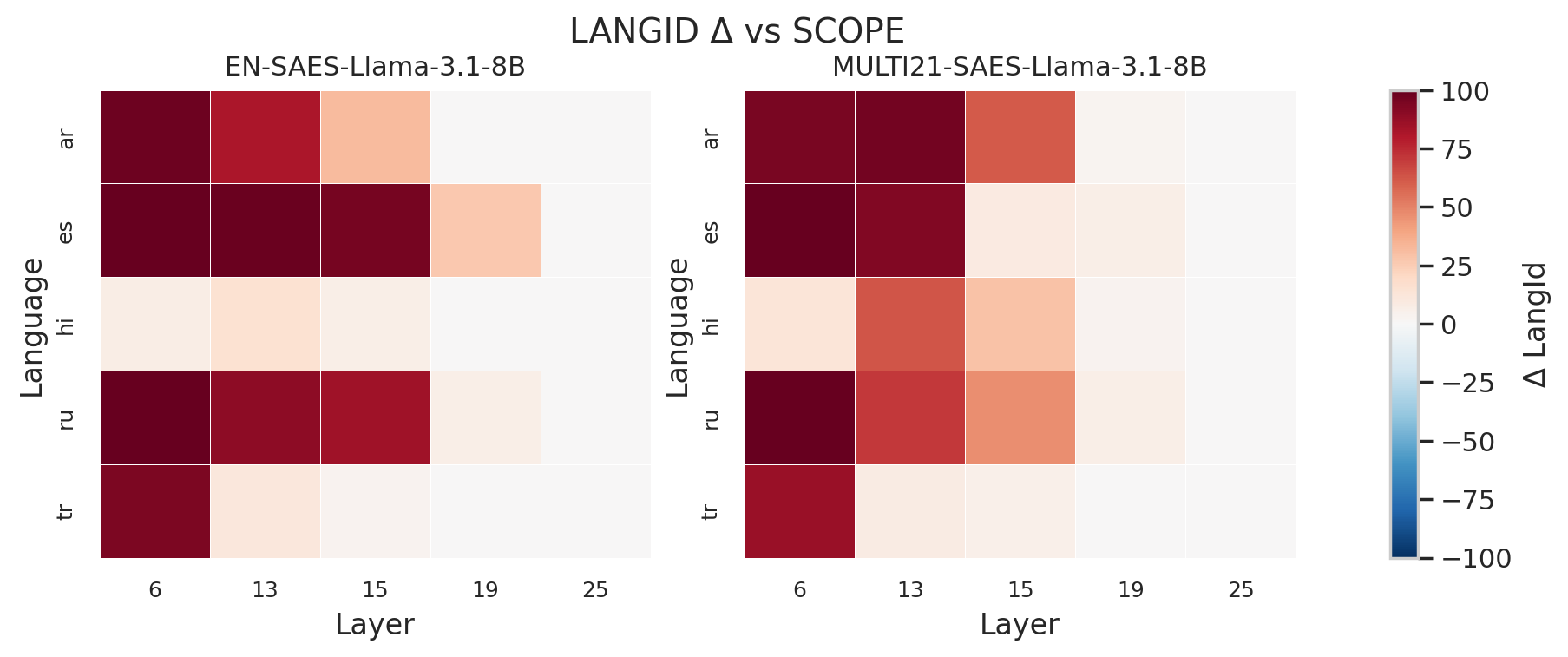}
    \includegraphics[width=0.8\linewidth]{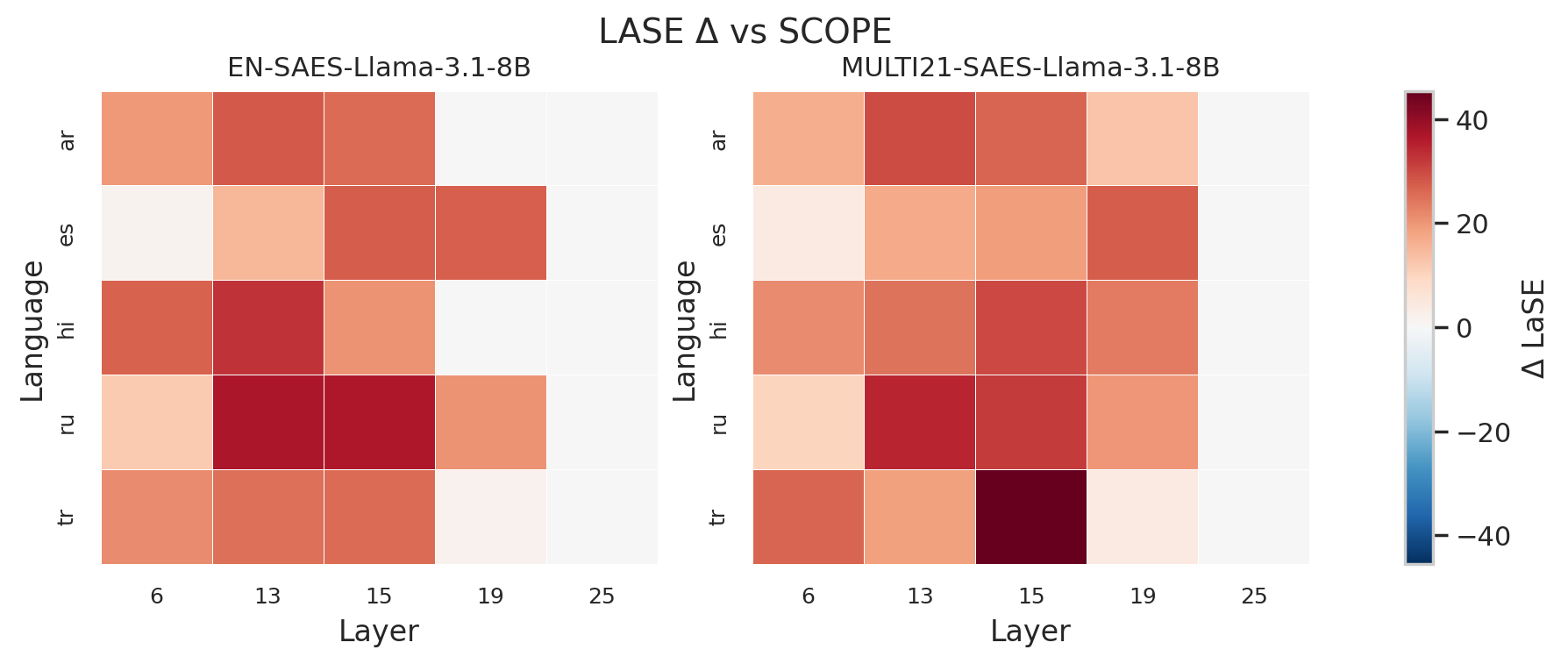}
    \includegraphics[width=0.8\linewidth]{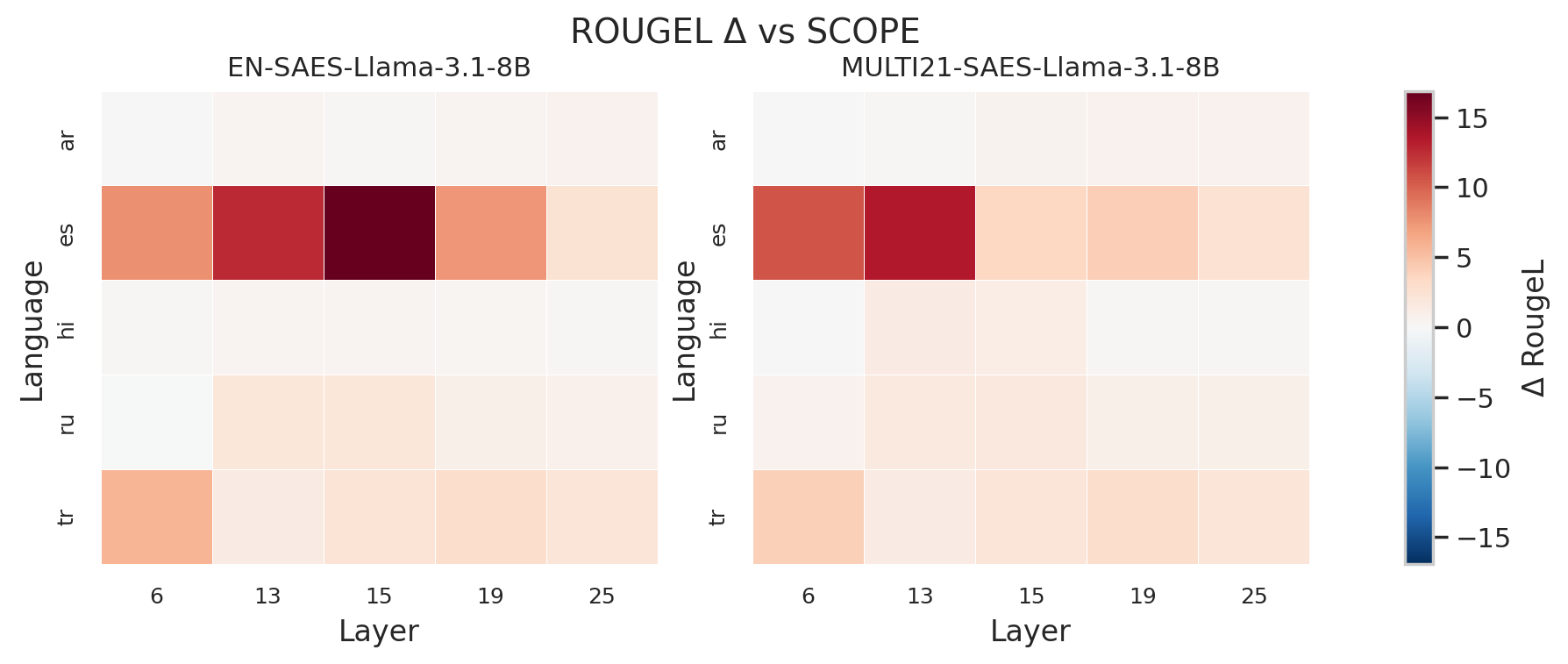}

    \caption{
Per-language, per-layer performance deltas for \textbf{LLaMA-3.1-8B} on the \textsc{CrossSum} task under cross-lingual steering (\texttt{tgt\_$i$} $\neq$ \texttt{steer\_$j$}). 
Each panel corresponds to a different SAE variant, showing how mismatched steering languages impact summarization quality and language identification across layers, relative to the SCOPE baseline.
}

    \label{fig:llama_cross_sum_all_diff}
\end{figure*}

\begin{figure*}[ht]

    \centering

    \includegraphics[width=0.8\linewidth]{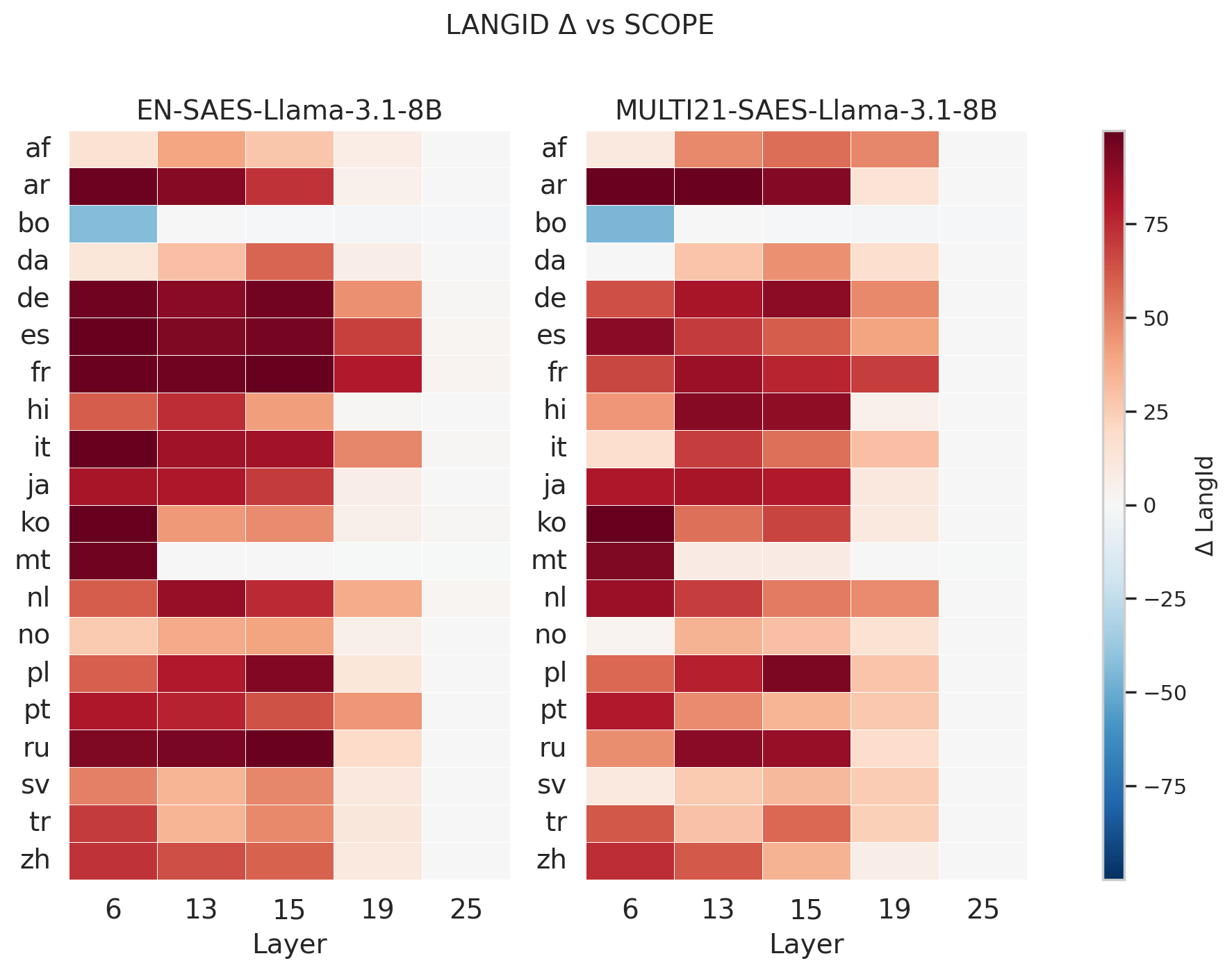}
    \includegraphics[width=0.8\linewidth]{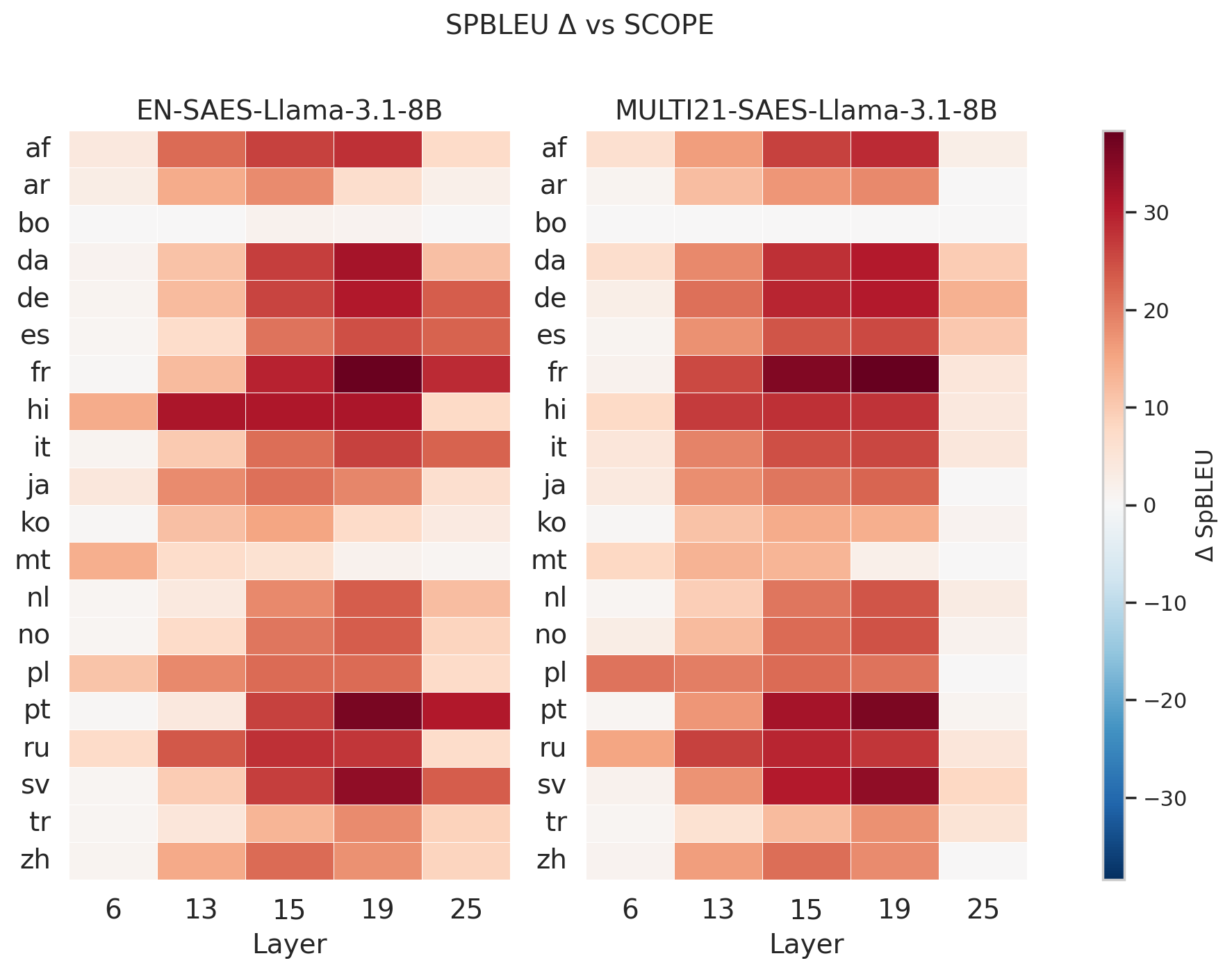}

    \caption{
Per-language, per-layer performance deltas for \textbf{LLaMA-3.1-8B} on \textsc{FLORES} with cross-lingual steering (\texttt{tgt\_$i$} $\neq$ \texttt{steer\_$j$}). 
The figure illustrates how steering in a different language affects language identification accuracy and translation quality across layers and SAE variants.
}

    \label{fig:llama_flores_2_diff}
\end{figure*}

\begin{figure*}[ht]
    \centering

    \includegraphics[width=0.8\linewidth]{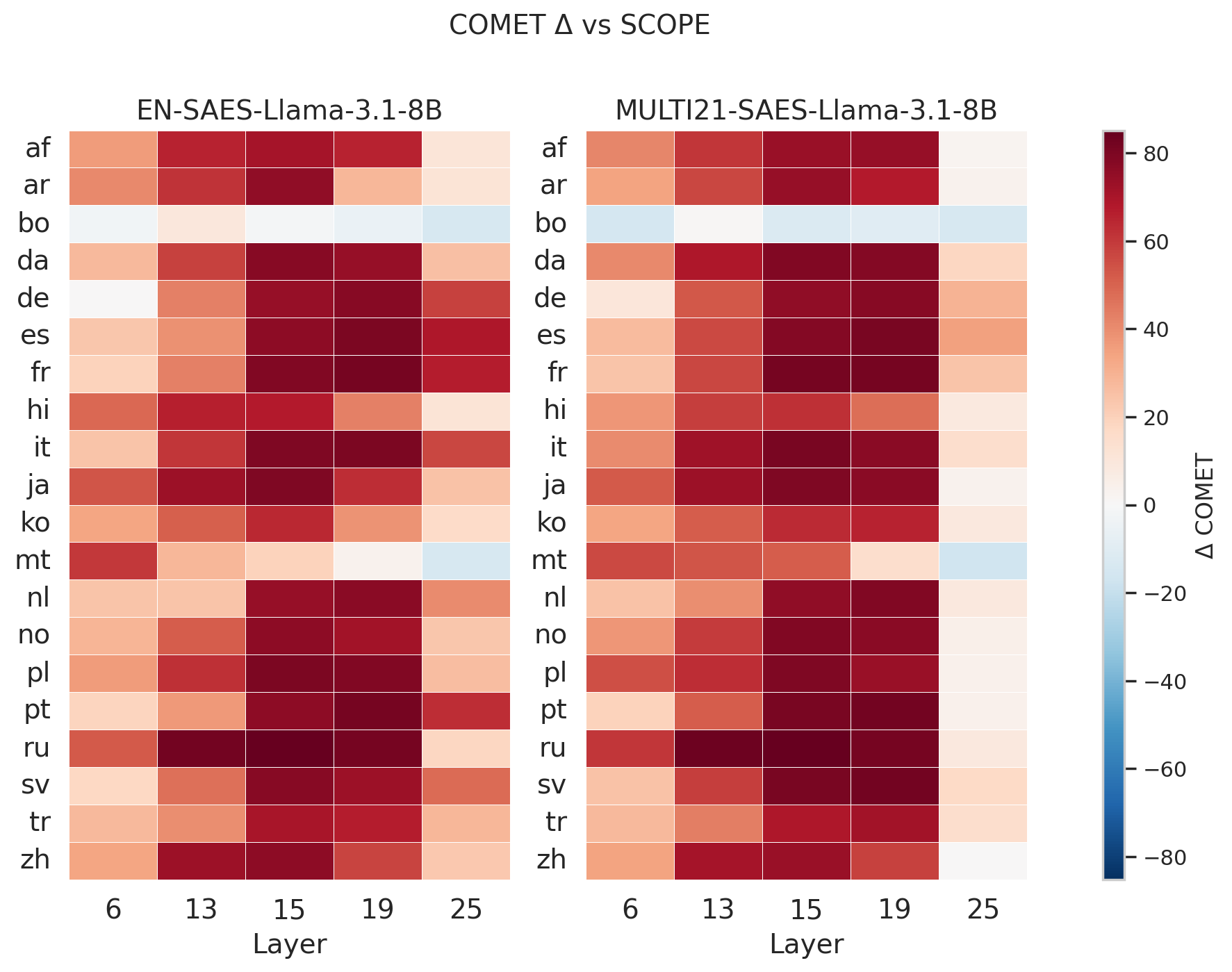}
    \caption{
Per-language, per-layer COMET score deltas for \textbf{LLaMA-3.1-8B} on \textsc{FLORES} under cross-lingual steering (\texttt{tgt\_$i$} $\neq$ \texttt{steer\_$j$}). 
Results highlight the sensitivity of semantic translation quality to steering language mismatches at different depths of the model.
}

    \label{fig:llama_flores_1_diff}
\end{figure*}

\end{document}